\documentclass[lettersize,journal]{IEEEtran}
\usepackage{amsmath,amsfonts}
\usepackage{algorithmic}
\usepackage{algorithm}
\usepackage{array}
\usepackage{booktabs}
\usepackage{ragged2e}
\usepackage[caption=false,font=normalsize,labelfont=sf,textfont=sf]{subfig}
\usepackage{textcomp}
\usepackage{threeparttable}
\usepackage{multirow}
\usepackage{stfloats}
\usepackage{url}
\usepackage{verbatim}
\usepackage{graphicx}
\usepackage{cite}
\usepackage{xcolor}
\begin{document}
\onecolumn
\begingroup
\fontsize{20pt}{20pt}\selectfont
IEEE Copyright Notice\\
\endgroup
\\
© 2023 IEEE. Personal use of this material is permitted. Permission from IEEE must be obtained for all other uses, in any current or future media, including reprinting/republishing this material for advertising or promotional purposes, creating new collective works, for resale or redistribution to servers or lists, or reuse of any copyrighted component of this work in other works.
\twocolumn

\title{Modeling Lead-vehicle Kinematics\\For Rear-end Crash Scenario Generation}

\author{Jian Wu, Carol Flannagan, Ulrich Sander, and Jonas Bärgman
\thanks{This work was supported by the FFI program sponsored by Vinnova, the Swedish governmental agency for innovation, as part of the project Improved quantitative driver behavior models and safety assessment methods for ADAS and AD (QUADRIS: nr. 2020-05156).}
\thanks{Jian Wu is with the Volvo Cars Safety Center, 41878 Gothenburg, Sweden, and the Department of Mechanics and Maritime Sciences, Chalmers University of Technology, 41756 Gothenburg, Sweden. (e-mail: jian.wu.2@volvocars.com)}
\thanks{Carol Flannagan is with the University of Michigan Transportation Research Institute (UMTRI), Ann Arbor, Michigan 48109, USA, and the Department of Mechanics and Maritime Sciences, Chalmers University, 41756 Gothenburg, Sweden. (email: cacf@umich.edu)}
\thanks{Ulrich Sander is with the Volvo Cars Safety Center, 41878 Gothenburg, Sweden. (e-mail: ulrich.sander@volvocars.com)}
\thanks{Jonas Bärgman is at the Department of Mechanics and Maritime Sciences, Chalmers University of Technology, 41756 Gothenburg, Sweden. (e-mail: jonas.bargman@chalmers.se)}
}

\markboth{IEEE Transactions on Intelligent Transportation Systems,~Vol.~14, No.~8, August~2021}%
{Shell \MakeLowercase{\textit{et al.}}: A Sample Article Using IEEEtran.cls for IEEE Journals}


\maketitle

\begin{abstract}
\textcolor{black}{The use of virtual safety assessment as the primary method for evaluating vehicle safety technologies has emphasized the importance of crash scenario generation. One of the most common crash types is the rear-end crash, which involves a lead vehicle and a following vehicle. Most studies have focused on the following vehicle, assuming that the lead vehicle maintains a constant acceleration/deceleration before the crash. However, there is no evidence for this premise in the literature. This study aims to address this knowledge gap by thoroughly analyzing and modeling the lead vehicle's behavior as a first step in generating rear-end crash scenarios. Accordingly, the study employed a piecewise linear model to parameterize the speed profiles of lead vehicles, utilizing two rear-end pre-crash/near-crash datasets. These datasets were merged and categorized into multiple sub-datasets; for each one, a multivariate distribution was constructed to represent the corresponding parameters. Subsequently, a synthetic dataset was generated using these distribution models and validated by comparison with the original combined dataset. The results highlight diverse lead-vehicle speed patterns, indicating that a more accurate model, such as the proposed piecewise linear model, is required instead of the conventional constant acceleration/deceleration model. Crashes generated with the proposed models accurately match crash data across the full severity range, surpassing existing lead-vehicle kinematics models in both severity range and accuracy. By providing more realistic speed profiles for the lead vehicle, the model developed in the study contributes to creating realistic rear-end crash scenarios and reconstructing real-life crashes.}
\end{abstract}


\section{Introduction}
\IEEEPARstart{V}{irtual} safety assessment has emerged as the primary approach for evaluating the safety of Advanced Driver Assistance Systems (ADAS) and Autonomous Driving (AD) systems due to its cost-effectiveness and efficiency compared to traditional field testing \cite{georgi2009new, riedmaier2020survey, riedmaier2020model, yves2015comprehensive}. The two main approaches to such assessment are traffic-simulation-based \cite{shah2018airsim, li2019aads, feng2021intelligent} and in-depth-crash-data-based (referred to as IDC-based) \cite{savino2016robust, bargman2017counterfactual, leledakis2021method}.

\textcolor{black}{The traffic-simulation-based approach simulates daily driving in order to create crash events in a virtual naturalistic driving environment \cite{shah2018airsim, li2019aads, feng2021intelligent}. Typically, traffic simulation models are built using naturalistic driving data (NDD), which includes few crashes, and those captured are typically of low severity. For safety evaluation, simulations are often conducted over an extended period (measured in millions of simulation hours) using the subject vehicle (i.e., the vehicle for which the system is assessed) both with and without the specific ADAS or AD systems. The number of crashes experienced in each situation is subsequently compared.}

\textcolor{black}{This approach has three primary challenges. First, it is very inefficient; due to the high dimensionality of the environment and the rareness of safety-critical events, demonstrating the safety performance of autonomous vehicles requires hundreds of millions of miles \cite{feng2021intelligent}. To tackle this problem, Feng et al. \cite{feng2021intelligent} proposed a solution known as the naturalistic and adversarial driving environment (NADE), which introduces sparse but adversarial modifications in order to reduce the number of virtual test miles needed while maintaining unbiased evaluations. However, even with the NADE technique, a substantial number of test miles is still necessary. Second, utilizing NDD as the initial condition for generating crash scenarios may lead to stark differences in crash characteristics compared to real-world crashes, both at the individual level and in terms of their overall distribution. Olleja et al. \cite{olleja2022can} compared crash generation methods using normal driving data and near-crash incidents with crashes obtained from in-depth crash databases. The results showed substantial disparities: normal driving data failed to reflect the crash outcomes and criticality observed in crashes from in-depth crash databases. Third and finally, crashes generated by the traffic-simulation-based approach rely heavily on accurate models of road-user behaviors that can produce realistic crashes (representative of real-world scenarios). However, validation of the details of the generated crashes is infrequent.}

\textcolor{black}{In contrast to the traffic-simulation-based approach, the IDC-based approach uses in-depth crash data containing reconstructed (and sometimes, although much more rarely, recorded) information such as vehicle kinematics to generate virtual crashes, either directly (by constructing digital twins for individual crashes) or indirectly (by sampling from distributions of relevant crash characteristics). A simulation with the ADAS or AD technology \cite{yves2015comprehensive, wang2020many, seacrist2020efficacy, savino2016robust, bargman2017counterfactual} is then run for each generated crash to answer the question, "What would happen if vehicle X were equipped with technology Y?"}  

\textcolor{black}{The IDC-based approach, however, also has its challenges. First, the safety assessments of ADAS and AD typically require more real-world crash instances than what is currently accessible. Furthermore, the limited availability of real-world crashes with in-depth information hampers the representation of the diverse range of crashes within specific scenarios. Consequently, "synthetic crashes", which can be viewed as variations of the original crashes, must be generated to fill in the gaps between real crashes \cite{leledakis2021method}. Second, the selection criteria used in traditional in-depth crash databases inherently introduce a bias toward severe crashes. Relying solely on these databases to create synthetic crashes \cite{leledakis2021method, gambi2019generating, wang2022autonomous} skews the crash generation models, potentially distorting the overall analysis. A third issue relates to how crashes are generated and their representativeness in the real world. Using reconstructions to generate crashes can be problematic as it involves making assumptions about individual road users, such as their braking profiles, without relying on detailed pre-crash recordings. While the crash outcome, such as the change in speed during the crash, may be reasonably accurate, the pre-crash kinematics are influenced by the decisions made during reconstruction and by the reconstruction software itself, leading to models based on assumptions and software rather than on explicit descriptions of the pre-crash kinematics of real crashes. In one such model, Gambi et al. \cite{gambi2019generating} proposed a model to efficiently generate crash scenarios by extracting crash information from police reports using Natural Language Processing (NLP) techniques. However, this model mainly relies on information from police reports and considers basic kinematics, overlooking driver behavior, leaving uncertainty about how accurately crashes generated from this "simplistic kinematics" method reflect the safety benefits of the assessed systems. A fourth issue is encountered when generating crashes at the tails of distributions. For example, Wang et al. \cite{wang2022autonomous} demonstrated that using independent component analysis (ICA) followed by kernel density estimation (KDE) to generate synthetic crashes can introduce biases, particularly near boundaries and in distributions with long tails \cite{zambom2013review}. Overall, methodological choices significantly impact the accuracy and generalizability of the crash scenarios produced for both scenario generation approaches.}

\textcolor{black}{To address the limitations of both approaches, a novel method combining data from naturalistic driving and recorded pre-crash kinematics from in-depth crash databases is proposed. This combined dataset covers the full severity range (from low to high severity levels) and aids in developing crash-generation models applicable to both traffic-simulation-based and IDC-based approaches. This study focuses specifically on rear-end crash scenario generation as an initial step in demonstrating the proposed method.}

A rear-end crash, in which the front of one vehicle collides with the rear of another, is a common crash type. In the United States, for example, rear-end crashes accounted for 27.8 percent of all car crashes in 2020 \cite{accidentreport2020}. Hence, studying rear-end crash scenario generation is essential. Moreover, the rear-end crash is relatively simple since it refers mainly to longitudinal maneuvers, and only two vehicles (the lead and the following vehicles) are involved. Consequently, synthetic rear-end crashes can be created based on models of the two involved vehicles, in which the lead vehicle is independent of the following vehicle, and the following vehicle responds to the presence and actions of the lead vehicle.

\textcolor{black}{To address these issues, we propose a novel approach to rear-end scenario generation, which combines the lead-vehicle kinematics model and the following vehicle behavior model to obtain representative rear-end crash scenarios across the full severity range. The work presented in this paper addresses only the first step of this crash scenario generation approach. Many other studies have analyzed the following vehicle's behavior during rear-end emergencies (crashes and near-crashes) by means of a driver response model \cite{brown2001human, lee2002collision, markkula2012review, li2014effectiveness, bargman2015does, markkula2016farewell, wang2016drivers, svard2021computational}. For example, Markkula et al. \cite{markkula2016farewell} used a piecewise linear model and used driver glance behaviors to model the following vehicles' speed profiles in naturalistic rear-end emergencies. They found that braking typically started less than a second after the kinematic urgency reached certain threshold levels; faster reactions occurred at higher urgencies.}

\textcolor{black}{However, there has been a notable lack of research on the lead vehicle's behavior in such situations despite its significant influence on the following vehicle. In crash reconstruction and rear-end emergency studies, it is commonly assumed that the lead vehicle maintains a constant acceleration or deceleration before the crash \cite{lee2002collision, li2014effectiveness, wang2016drivers}, even though there is inadequate evidence to support this assumption. To address this knowledge gap, the objective of this study is to develop a model of the lead-vehicle kinematics in rear-end crashes across the full severity range as the first step in generating rear-end crash scenarios. Future work will use models of the following vehicle’s behavior together with the lead-vehicle kinematics model from this work to generate rear-end crash scenarios.}

In this study, a piecewise linear model was employed to parameterize the speed profiles of the lead vehicles in two established datasets of rear-end pre-crash/near-crash incidents. These datasets consist of recorded vehicle kinematics prior to the actual crash or near-crash events. The two datasets were combined to create a comprehensive dataset that spans the full range of severity, which was then categorized into multiple sub-datasets based primarily on lead-vehicle speed change patterns. Next, a multivariate distribution model of the parameters was built for each sub-dataset. Finally, a synthetic dataset was generated by sampling the synthetic lead vehicles' speed profiles created by the distribution models in proportion to the sample size of each sub-dataset. The synthetic dataset was then validated by comparing the parameter distributions and kinematics of the generated crashes with the original combined dataset.

\section{Datasets} \label{section:datasets}
This study focuses on passenger vehicle rear-end crashes/near-crashes, and the data come from two sources: the Crash Investigation Sampling System (CISS) and the Second Strategic Highway Research Program (SHRP2) Naturalistic Driving Study (NDS).

CISS is a nationally representative complex probability sample of general passenger vehicle crashes in the United States in which at least one light vehicle was towed away \cite{zhang2019crash, subramanian2020crash}. The data collection started in 2017 and is still ongoing. The data are from in-depth crash investigations that include inspection of damaged vehicles and crash sites, as well as estimation of crash kinematics. In addition, Event Data Recorder (EDR) data were extracted and included in the sample whenever possible.

In the SHRP2 NDS, over 3,300 passenger vehicles were instrumented with a data acquisition system (DAS) that collected four video views (driver’s face, driver’s hands, forward roadway, and rear roadway) and information from vehicle networks and sensors \cite{hankey2016description}. Naturalistic driving data from six sites around the United States were collected through the participant vehicles between 2010 and 2013. Unlike the CISS dataset, which only contains crashes, the SHRP2 dataset contains both crashes and near-crashes. A dozen trigger algorithms were executed on collected trip files followed by manual annotation to identify crashes and near-crashes (defined as any circumstances that require a rapid evasive maneuver by the subject vehicle or any other vehicle, pedestrian, cyclist, or animal to avoid a crash \cite{hankey2016description}). It is worth mentioning that there is no overlap between the CISS and SHRP2 datasets since the data were collected at different times.

\subsection{Data Selection}
Rear-end pre-crash/near-crash data for incidents in which the subject vehicle that collected the data was the lead vehicle (struck vehicle) were extracted from both datasets. The subject vehicle's speed $v$ (unit: $m/s$) was the only signal we used; the signal was directly estimated from the wheel speed or in-vehicle inertial sensor.

In the CISS dataset, only rear-end crashes in which the struck vehicle was equipped with an event data recorder containing recorded data from the pre-crash phase were extracted. Among these cases, the ones with a data frequency greater than 5 Hz were selected for this study. Of the selected 52 CISS crashes, three have a frequency of 5 Hz, while the rest have a frequency of 10 Hz.

In the SHRP2 dataset, incidents (crashes and near-crashes) labeled "Rear-end, struck" were selected. The frequency of all the SHRP2 data used is 10 Hz.

More details about data selection, including the code for extracting cases, are provided in Section I of the supplement.

\begin{table}[!t]
\centering
\caption{All extracted events\label{tab:allEvents}}
\begin{threeparttable}
\begin{tabular}{ccccc}
\hline
Group & Notation & Source & Severity level$^a$ & Sample size$^b$\\
\hline
1 & CISS$_{sc}$ & CISS & Severe & 49/52\\
2 & SHRP2$_{sc}$ & SHRP2 & Severe & 20/24\\
3 & SHRP2$_{nsc}$ & SHRP2 & Non-severe & 63/106\\
4 & SHRP2$_{nc}$ & SHRP2 & None & 171/272\\
\hline
\end{tabular}
\begin{tablenotes}
\RaggedRight
\item $^a$ The severity level here does not correspond with the Abbreviated Injury Scale (AIS) \cite{gennarelli2006ais}. A crash is indexed as severe if it fulfills the 'SHRP2 severity level I' definition \cite{shrp2document}; otherwise, non-severe. Furthermore, the severity level for a near-crash incident is designated as 'None'.
\item $^b$ Valid sample size/raw sample size.
\end{tablenotes}
\end{threeparttable}
\end{table}

\subsection{Data Groups}
\textcolor{black}{In the study, the crashes extracted from the SHRP2 dataset were originally labeled according to severity level: I (most severe), II (police-reportable), and III (minor). Note that level IV (low-risk tire strikes) is not included. A crash that involves an airbag deployment, injury to the driver, pedal cyclist or pedestrian, vehicle rollover, high Delta V, or requires vehicle towing is classified as severity level I. A level II crash is any police-reportable crash that does not meet the level I crash requirements. All other crashes that involve physical contact with minimal damage are considered level III crashes. \cite{shrp2document} In this study, a crash is indexed as 'Severe' if it fulfills the SHRP2 severity level I definition and 'Non-severe' otherwise. In addition, the severity level of any near-crash is designated as 'None'.}

The data were separated into four groups according to source and severity level, as Table \ref{tab:allEvents} shows. Group 1, CISS$_{sc}$, comprises the extracted CISS crashes. They fulfill the SHRP2 severity level I definition and are, therefore, considered severe crashes. Extracted SHRP2 severity level I crashes belong to Group 2, SHRP2$_{sc}$. The SHRP2 crashes at severity levels II and III make up Group 3, SHRP2$_{nsc}$ (non-severe crashes). Group 4, SHRP2$_{nc}$, consists of SHRP2 near-crashes. The raw sample sizes of Groups 1-4 are 52, 24, 106, and 272, respectively. However, not all samples are valid; the conditions for selecting valid samples are introduced in the following subsection.

\subsection{Event Data Extraction}
Time zero, for a crash, is the impact moment. For a near-crash, it is the moment when the following vehicle reaches the minimum distance to the subject vehicle. This moment was annotated manually according to the video.

CISS data typically contains five seconds before the impact, while SHRP2 data contains a longer duration. To make all events equivalent, -5 s was set as the start-point of all events. The closest data point to the impact moment was excluded to avoid a possible sharp acceleration pulse near impact. Because the lowest data frequency is 5 Hz, raw data were extracted only up to -0.3 s before impact ($t = -5$ s to $t = -0.3$ s) for each crash. For a near-crash, the extracted duration is from $t = -5$ s to $t = 0$ s.

The extracted events fulfilling the following conditions were considered valid and selected for further analysis:
\begin{itemize}
    \item The total sample duration should be no less than three seconds (due to missing or invalid data).
    \item The fitted accelerations should range between -1 $g$ and 1 $g$, where $g$ is the gravitational acceleration. (This study simplifies the lead-vehicle speed profile as a sequence of straight lines. The fitted accelerations are the slopes of those lines. More details on fitted acceleration can be found in Section \ref{section:methodology}.)
\end{itemize}
There were 49, 20, 63, and 171 valid samples for Groups 1-4, respectively (see Table \ref{tab:allEvents}).

\section{Methodology} \label{section:methodology}
\begin{figure*}[!t]
    \centering
    \includegraphics[width=0.8\textwidth]{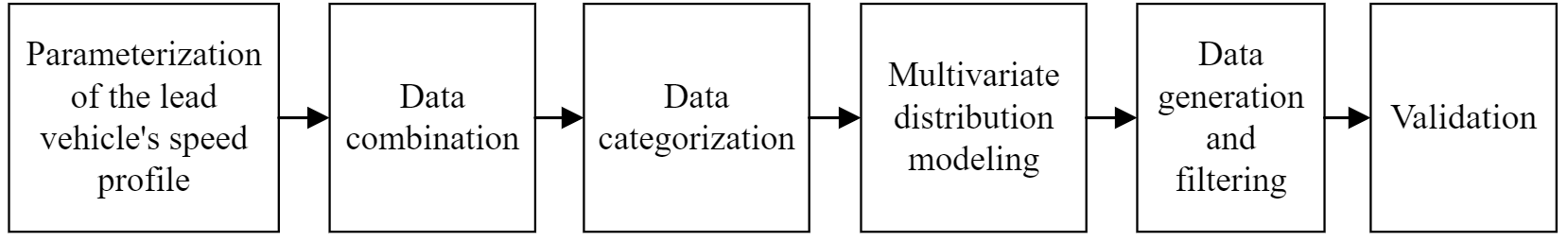}
    \caption{Flowchart of six steps of data analysis performed in the study.}
    \label{fig:methodology}
\end{figure*}
The following six steps (see Fig. \ref{fig:methodology}) were performed successively: 
\begin{enumerate}
    \item Parameterization of the lead-vehicle speed profiles
    \item Data combination
    \item Data categorization
    \item Multivariate distribution modeling
    \item Data generation and filtering
    \item Validation
\end{enumerate}
The methods used in each step will be presented in this section, with the exception of the data categorization step, which is based on the results from other steps and introduced in Section \ref{section:results}.

\subsection{Parameterization of the Lead-vehicle Speed Profiles}
For each case, the lead-vehicle speed profile is fitted to a piecewise linear model, similar to that used by Markkula et al. \cite{markkula2016farewell}. However, the applications are different. Their work modeled the following vehicle's braking behavior (acceleration, not speed) as only one braking phase with constant deceleration. The lead vehicle's braking behavior can have multiple phases, however, making the piecewise linear model here more complicated.

\subsubsection{Piecewise linear model}
In this model, the lead-vehicle speed profile is simplified as several consecutive straight lines. The connection points of these lines are named breakpoints. The model contained the following steps:
\begin{list}{}{}
    \item {\bf Step 0:} Start.
    \item {\bf Step 1:} Set sample weight to emphasize the different importance levels of different samples. The closer to time zero, the more relevant the sample is to the crash/near-crash, and thus the more important and the greater the weight. In this case, we can prioritize capturing the speed changes closer to time zero and avoid overfitting the early samples. The weight of sample $i$ is defined as
    \begin{equation} \label{eq:fitwgt}
        w_{i} = (0.1 - t_{i})^{-0.5},
    \end{equation}
     where $w_{i}$ and $t_{i}$ are the weight and time of sample $i$ respectively. (More details about setting the sample weights can be found in Section II-A1 of the supplement.)
    \item {\bf Step 2:} Fit the lead-vehicle speed profile using weighted piecewise linear regressions with the number of breakpoints $n_b$ from zero to the pre-configured maximum number of breakpoints $n_{b,max}$. The weighted piecewise linear regression is based on the "piecewise-regression" package in python \cite{pilgrim2021piecewise}, which fits a curve as several consecutive straight lines. The option to consider sample weights has been added to the package locally.
    \item {\bf Step 3:} Select the best regression. Compute the loss $L$ according to (\ref{eq:lossfunc}) for each regression, and choose the regression with the minimum loss.
    \begin{equation} \label{eq:lossfunc}
        L = (\epsilon + \lambda \cdot \frac{\text{max}(v)}{\Delta v + \epsilon}) \cdot n_b - R^2,
    \end{equation}
    where $\epsilon$ is a small positive value to avoid a zero denominator or a zero penalty for the number of breakpoints when max($v$) is zero, $\lambda\ (>0)$ is a pre-configured penalty coefficient, $\Delta v(=\text{max}(v)-\text{min}(v))$ is the maximum speed change, and $R^2$ is the R-squared of the regression (indicating the fitting accuracy). The loss function penalizes an excessive number of breakpoints to avoid overfitting, especially when max($v$) is large and $\Delta v$ is small. (The justification for the loss function can be found in Section II-A2 of the supplement.)
    \item {\bf Step 4:} Modify the fitting results to avoid any negative estimated speed $\hat{v}$ (due to estimation error) in the modeling duration, -5 to 0 s (the sampled duration might be shorter than 5 seconds, yet the model can predict the speed for the missing part). There are two sub-steps:
    \begin{enumerate}
        \item Add one breakpoint in the start or end segment where $\hat{v}$ is 0 m/s if there is any negative $\hat{v}$ at the start-point or end-point of the modeling duration. Then set $\hat{v}$ to 0 m/s from the newly added breakpoint to the start-point or end-point.
        \item Change $\hat{v}$ at that breakpoint to 0 m/s if there is any negative estimated speed value at any breakpoint. Then connect the modified breakpoint with other points.
    \end{enumerate}
    Given that $\hat{v}$ is non-negative at the start-point, end-point, and all breakpoints, $\hat{v}$ during the whole modeling duration should be non-negative.
    \item {\bf Step 5:} End.
\end{list}

This piecewise linear model aims to fit the lead-vehicle speed profile with the simplest regression model possible. Consequently, we set $n_{b, max} = 3$, $\lambda = 0.006$, and $\epsilon=1 \times 10^{-6} m/s$. $n_{b, max}$ was set as the maximum number of breakpoints annotated among a small sub-dataset that was randomly sampled, while $\lambda$ was set as the elbow of the curve, representing the total number of breakpoints for all events plotted against $\lambda$. (More details of the selection of pre-configured parameters are provided in Section II-A3 of the supplement.)

\subsubsection{Parameters}
\begin{figure}[!t]
    \centering
    \includegraphics[width=0.35\textwidth]{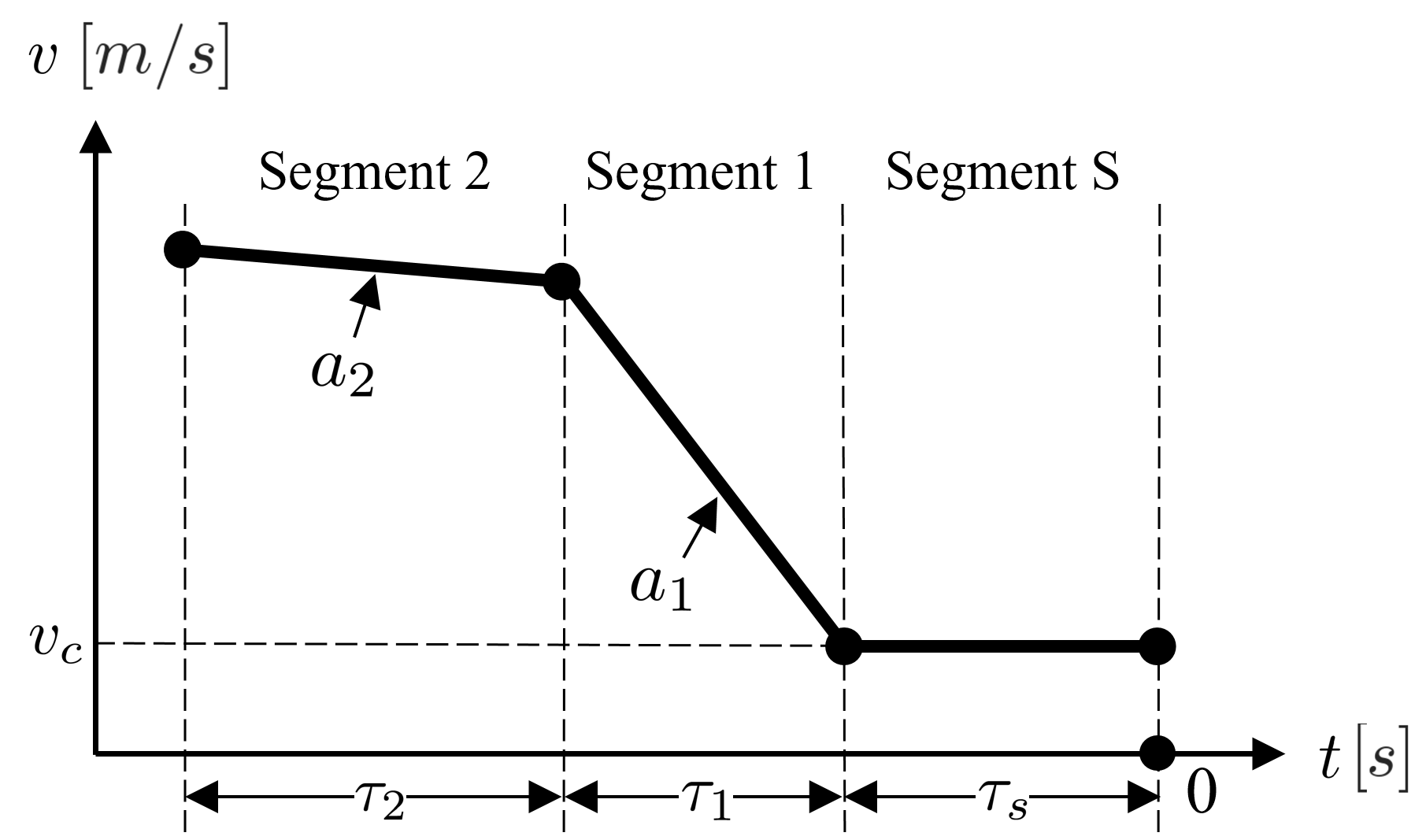}
    \caption{Three selected segments.}
    \label{fig:segments}
\end{figure}

The piecewise linear model consists of a maximum of four consecutive lines with three breakpoints. Given the priorities of simplification and sample weights, it is unnecessary to include the segments relatively far from time zero when there are more than three segments. If the lead vehicle reaches a steady speed at the last segment (S), at most, three segments closest to time zero are selected; otherwise, at most, two segments closest to time zero are selected (i.e., segment S is removed). Fig. \ref{fig:segments} shows an example of three selected segments. The following are explanations of each segment (backward in time from time zero), including two descriptive parameters, in the context of lead-vehicle pre-crash kinematics:
\begin{itemize}
    \item Segment S: The lead vehicle maintains a steady speed in this segment. $\tau_s$ is the segment duration, and $v_c$ is the lead vehicle's estimated speed at time zero.
    \item Segment 1: The lead vehicle keeps a non-zero constant acceleration in this segment. $\tau_1$ is the segment duration, and $a_1$ is the constant acceleration.
    \item Segment 2: The lead vehicle keeps a constant acceleration in this segment. $\tau_2$ is the segment duration, and $a_2$ is the constant acceleration.
\end{itemize}

The six-parameter vector $[v_c, a_1, a_2, \tau_s, \tau_1, \tau_2]$ is used to represent an event. It is important to note that not every event contains all three segments. Segment S and Segment 1 can exist independently, while Segment 2 can only exist when Segment 1 exists. There were five possible combinations, including their proportions: Segment S (8.9\%); Segment S \& 1 (8.3\%); Segment S \& 1 \& 2 (22.8\%); Segment 1 (6.9\%); and Segment 1 \& 2 (53.1\%). The parameters of any non-existent segments are defined according to the following rules:
\begin{enumerate}
    \item If Segment S is non-existent, $\tau_s=0$ s.
    \item If Segment 1 is non-existent, $\tau_1=0$ s and $a_1=0$ m/s$^2$.
    \item If Segment 2 is non-existent, $\tau_2=0$ s and $a_2=a_1$.
\end{enumerate}

\subsection{Data Combination} \label{section:datacombination}
The CISS and SHRP2 datasets can be interpreted as two perspectives of crashes in the United States. By weighting the CISS and SHRP2 crashes appropriately, we can combine these two perspectives to achieve a combined dataset that describes the full range of passenger vehicle crashes (from non-severe to severe), leveraging each dataset's individual strength. For the sake of simplicity, the terms ‘crash,’ ‘near-crash,’ and 'incident' (without any further specification) refer to the lead vehicle's behavior (speed profile) for each respective type.

Assuming that severe crashes in both the CISS and SHRP2 datasets (CISS$_{sc}$ and SHRP2$_{sc}$) come from the same distribution is crucial before combining the two datasets. The rationale for this assumption is the similarity between the definitions of CISS crashes (police-reported towed vehicle crashes) and severe SHRP2 crashes (SHRP2 severity level I crashes). Moreover, non-parametric tests were conducted to determine if CISS$_{sc}$ and SHRP2$_{sc}$ are significantly different. The results do not show any significance. (More details of the comparison can be found in Section II-B of the supplement.)

The two datasets were combined in the following three steps: 1) pre-processing crash data, 2) reweighting crash data, and 3) adding selected near-crashes as variations of crashes. Steps 1-2 combined crashes in the CISS and SHRP2 datasets into one dataset, the combined crash dataset. Step 3 added selected SHRP2 near-crashes to the combined crash dataset, and the new dataset, including the selected near-crashes, is called the combined incident dataset. It contains the parameterized (not raw) incident and their sample weights and is published online and available to the public \cite{Wu_QUADRIS_project_pre-crash_near-crash}. 

\subsubsection{Pre-processing crash data}
Generally, sample weights are used so that the weighted data represent the frequency of occurrence. Before combining the crashes in the two datasets, we pre-processed the crash data so that the sample weights of the two datasets were compatible. Preprocessing tunes the existing sample weights (for the CISS dataset) or sets sample weights (for the SHRP2 dataset) so that the sum of sample weights equals the valid sample size for each respective dataset.

According to \cite{zhang2019crash}, CISS crashes were sampled using a probability sampling method with sampling features such as stratification, clustering, and unequal selection probabilities. Thus, the CISS sample is not a simple random sample; each CISS crash was assigned a sample weight to produce an unbiased estimation. These original sample weights range expansively, from 21.8 to 3833.6. However, extreme variation in sampling weights can result in excessively large sampling variances when the data and the selection probabilities are not positively correlated \cite{potter2015methods}. To reduce sampling variance, a weight-trimming approach proposed in \cite{van2014weight} was applied. (More details of this approach are provided in Section II-C of the supplement.) The next is to scale the trimmed weights so that the sum of the scaled weights equals the valid sample size.
\begin{equation} \label{eq:reweight}
    w_{1,i} = n_{1,vld} \cdot \frac{w_{1,i,t}}{\sum_{i} {w_{1,i,t}}},
\end{equation}
where $w_{1,i}$ is the weight of crash $i$ in CISS$_{sc}$ after scaling, $w_{1,i,t}$ is the weight of crash $i$ in CISS$_{sc}$ after trimming, and $n_{1,vld}$ is the valid sample size of CISS$_{sc}$.

SHRP2 crashes (SHRP2$_{sc}$ and SHRP2$_{nsc}$), since they occurred during the data collection period in the SHRP2 project, were not sampled but directly collected. Consequently, there are no sample weights in the SHRP2 dataset, yet all raw crashes can be seen with the same weight. The crashes in Group $i\ (i = 2,3)$ are assigned with the same weight $w_i$, which is computed as
\begin{equation}
    w_i = (n_{2,vld} + n_{3,vld}) \cdot \frac{n_i}{n_2 + n_3} \cdot \frac{1}{n_{i,vld}}.
\end{equation}
(As shown in Table \ref{tab:allEvents}, among the $n_{2} = 24$ and $n_{3} = 106$ samples in SHRP2$_{sc}$ and SHRP2$_{nsc}$, respectively, there were $n_{2, vld} = 20$ and $n_{3,vld} = 63$ valid samples, respectively.) The rationales for this sample weighting design are 1) the valid samples are selected to represent the raw samples, 2) the weighted SHRP2 data should have the same proportions of severe and non-severe crashes as the raw SHRP2 data, and 3) the sum of the weights equals the valid sample size ($n_{2, vld}+n_{3,vld}$).

\subsubsection{Reweighting crash data}
This step combined the CISS and SHRP2 crashes into a combined crash dataset. The combined dataset includes more information than either the CISS or SHRP2 dataset alone, but it also should retain the raw distributions. The lead vehicle's estimated speed at time zero, $v_c$, is the most important and representative of the six parameters. The objective of retaining the raw distributions of the original datasets can be described as two sub-objectives. For the combined crash dataset:
\begin{enumerate}
    \item Keep the same distribution of $v_c$ for severe crashes as the CISS dataset.
    \item Keep the same proportions of severe and non-severe crashes as the SHRP2 dataset.
\end{enumerate}

\begin{figure}[!t]
    \centering
    \includegraphics[width=0.4\textwidth]{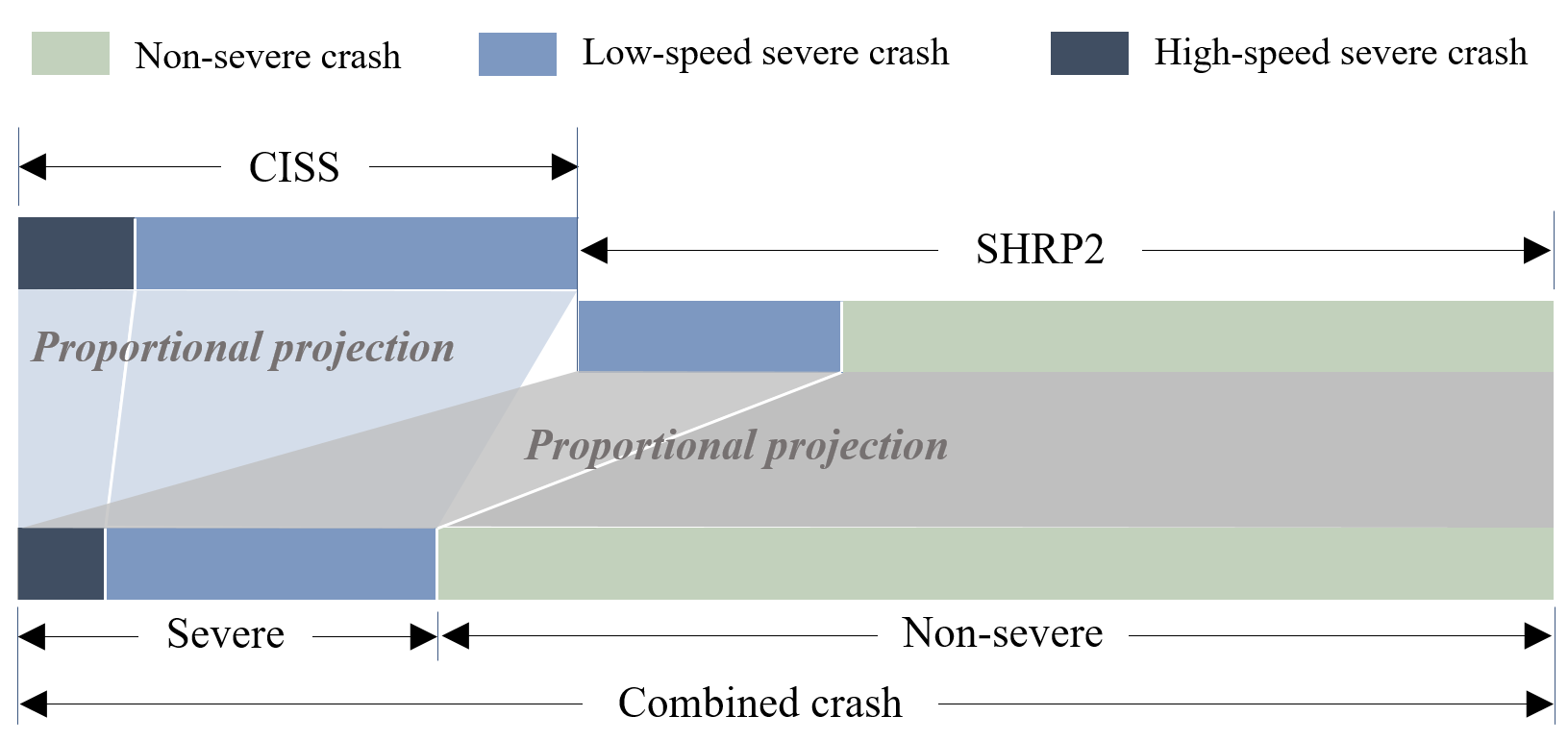}
    \caption{Combination of crashes in the CISS and SHRP2 datasets. The dataset's icon length corresponds with its valid sample size. The combined crash dataset keeps 1) the same proportions of low-speed and high-speed severe crashes as the CISS dataset and 2) the same proportions of severe and non-severe crashes as the SHRP2 dataset.}
    \label{fig:datacombination}
\end{figure}

Compared with SHRP2$_{sc}$, CISS$_{sc}$ has a wider range of $v_c$: SHRP2$_{sc}$ [0, 7.9] m/s, CISS$_{sc}$ [0, 30.4] m/s. Further, the crashes in CISS$_{sc}$ and SHRP2$_{sc}$ can be divided into two types:
\begin{itemize}
    \item Low-speed: Crashes where $v_c$ is lower than the maximum $v_c$ in SHRP2$_{sc}$.
    \item High-speed: Crashes where $v_c$ is higher than the maximum $v_c$ in SHRP2$_{sc}$.
\end{itemize}
Per definition, SHRP2$_{sc}$ contains only low-speed severe crashes, while CISS$_{sc}$ contains both low-speed and high-speed severe crashes. Thus, the first sub-objective is equivalent to keeping the same proportions of low-speed and high-speed severe crashes given crashes in the CISS$_{sc}$ and SHRP2$_{sc}$ are from the same distribution. Fig. \ref{fig:datacombination} shows the two sub-objectives of combining crashes in the CISS and SHRP2 datasets.

Consequently, this step reweighted the CISS and SHRP2 crashes and combined them as Fig. \ref{fig:datacombination} shows. The proportion of non-severe crashes in SHRP2 crashes, $\eta_{ns}$, and the proportion of high-speed severe crashes in CISS, $\eta_{hss}$, are computed according to (\ref{eq:proportionofns}) and (\ref{eq:proportionofhss}) respectively.  
\begin{equation} \label{eq:proportionofns}
    \eta_{ns} = \frac{n_{3}}{n_2 + n_3},
\end{equation}
\begin{equation} \label{eq:proportionofhss}
    \eta_{hss} = \frac{W_{hss}} {n_{1,vld}},
\end{equation}
where $W_{hss}$ is the total weight of high-speed severe crashes before combination; it is computed as
\begin{equation}
    W_{hss} = \sum_{i | v_{c,1,i} > max(v_{c,2})} {w_{1,i}},
\end{equation}
where $v_{c,1,i}$ is the estimated lead vehicle's speed at time zero for crash $i$ in Group 1, CISS$_{sc}$, and $v_{c,2}$ is the estimated lead vehicle's speed values at time zero for crashes in Group 2, SHRP2$_{sc}$.

As (\ref{eq:sumofn}) shows, the sample size of the combined crash dataset, $n_{cmb}$, is the sum of the valid samples of CISS$_{sc}$, SHRP2$_{sc}$ and SHRP2$_{nsc}$.
\begin{equation} \label{eq:sumofn}
    n_{cmb} = \sum_{i = 1} ^ {3} n_{i,vld}.
\end{equation}

Because the combined crash dataset should retain the same $\eta_{ns}$ and $\eta_{hss}$, the equivalent sample sizes can be computed as
\begin{equation}
    n^{\prime}_{ns} = n_{cmb} \cdot \eta_{ns},
\end{equation}
\begin{equation}
    n^{\prime}_{hss} = n_{cmb} \cdot (1 - \eta_{ns}) \cdot \eta_{hss},
\end{equation}
\begin{equation}
    n^{\prime}_{lss} = n_{cmb} \cdot (1 - \eta_{ns}) \cdot (1 - \eta_{hss}), 
\end{equation}
where $n^{\prime}_{ns}$ is the sample size of non-severe crashes (SHRP2$_{nsc}$) after combination, and $n^{\prime}_{hss}$ and $n^{\prime}_{lss}$ are the sample sizes of high-speed and low-speed severe crashes after combination. The weights can then be deduced according to (\ref{eq:g3weight}-\ref{eq:g1weight}).
\begin{equation} \label{eq:g3weight}
    w^{\prime}_{3} = \frac{n^{\prime}_{ns}} {n_{3,vld}},
\end{equation}
\begin{equation} \label{eq:g2weight}
    w^{\prime}_{2} = n^{\prime}_{lss} \cdot \frac{ w_2} {W_{lss}},
\end{equation}
\begin{equation} \label{eq:g1weight}
    w^{\prime}_{1,i} =
    \begin{cases}
        {n^{\prime}_{lss} \cdot \frac{w_{1,i}} {W_{lss}},} & {\text{if}\ v_{c,1,i} \leq max(v_{c,2});} \\
        {n^{\prime}_{hss} \cdot \frac{w_{1,i}} {W_{hss}},} & {\text{if}\ v_{c,1,i} > max(v_{c,2}).}
    \end{cases},
\end{equation}
where $W_{lss}$, computed according to (\ref{eq:sumoflssweights}), is the total sample weight of low-speed severe crashes before combination.
\begin{equation} \label{eq:sumoflssweights}
    W_{lss} = w_2 \cdot n_{2,vld} + \sum_{i | v_{c,1,i} \leq max(v_{c,2})} {w_{1,i}}.
\end{equation}

\subsubsection{Adding selected near-crashes as variations of crashes} \label{section:addnc}
After the last step, the combined crash dataset was acquired. However, the sample size of the combined crash dataset, namely the sum of the valid samples from CISS$_{sc}$, SHRP2$_{sc}$, and SHRP2$_{nsc}$, is only 132. As this is a low number for modeling distributions of the six parameters, $[v_c, a_1, a_2, \tau_s, \tau_1, \tau_2]$, we increased the number of samples by adding near-crashes similar in terms of the lead vehicle's behavior to any crash (more details in Section \ref{section:addnc}). Sample weights must be adjusted so that the valid sample size (sum of sample weights) remains the same and the added near-crashes do not change the distributions in the combined crash dataset. More details of the rationale and consequences of this action will be discussed in Section \ref{section:discussion}.

In the SHRP2 dataset, not all near-crashes were captured. The SHRP2 near-crashes were automatically captured by designed trigger specifications and then validated by manual annotation. In other words, those trigger specifications contribute to a certain sampling bias of near-crashes in the SHRP2 dataset. For the rear-end near-crashes, the primary trigger specification is the "Longitudinal Deceleration," which requires the level of longitudinal acceleration to be less than or equal to -0.65 g, and the threshold is exceeded for at least one timestamp \cite{hankey2016description}. Consequently, near-crashes in which the lead vehicle does not brake harshly enough to reach that threshold are under-represented because they are not captured. Therefore, to avoid introducing bias, we cannot simply add all SHRP2 near-crashes directly to the combined crash dataset to form an even more comprehensive dataset covering all incident severities from near-crashes to severe crashes. However, lead-vehicle behaviors near-crashes in some near-crashes are similar to those observed in crashes; these near-crashes can be added to the combined crash dataset as variations of crashes. The implementation of the merging of crashes with a subset of the near-crashes was done with the following steps.
\begin{enumerate}
    \item For near-crash $i$, defined by the parameter space $[v^i_c, a^i_1, a^i_2, \tau^i_s, \tau^i_1, \tau^i_2]$, find the most similar crash across the combined crash dataset using the Euclidean distance computed based on the standardized six parameters (z-score). The crash most similar to near-crash $i$ (of all the crashes in the combined crash dataset) is defined as the one with the minimum Euclidean distance, $d_{i,min}$, and is called the 'most similar crash' of near-crash $i$.
    \item Near-crashes with a minimum Euclidean distance less than a set threshold $d_{thd}$ ($d_{i,min} \leq d_{thd}$) are considered similar enough to crashes to be selected for addition to the combined crash dataset.
    \item For crash $j$, if there are in total $n^j_{nc}$ ($n^j_{nc} \geq 0$) near-crashes whose most similar crash is crash $j$, the weights of crash $j$ and those near-crashes in the combined incident dataset are set as $\frac{w^j}{1+n^j_{nc}}$, where $w^j$ is the weight of crash $j$ in the combined crash dataset.
\end{enumerate}

The threshold $d_{thd}$ was set as 0.78, and it was according to the analysis of the similarity between crashes in the combined crash dataset. As in the first step, we computed the minimum Euclidean distances for the lead-vehicle behaviors in crashes from the combined crash dataset. Then $d_{thd}$ was set based on the cumulative distribution function (CDF) of the mentioned distances considering sample weights in the combined crash dataset. (More details of the choice of $d_{thd}$ can be found in Section II-D of the supplement.)

In the first two steps, to select near-crashes similar to crashes in the combined crash dataset, the Euclidean distance was used to measure the similarity between incidents. The last step was to adjust the sample weights to retain the raw distributions in the combined crash dataset.

\subsection{Multivariate Distribution Modeling}
The combined incident dataset was categorized into several sub-datasets (more details on this process in Section \ref{section:datacate}). For each sub-dataset, a multivariate distribution model was built to generate synthetic lead-vehicle speed profiles. A synthetic incident dataset was then built by sampling the generated speed profiles in proportion to the sample size of each sub-dataset.

The modeling principle here is to make things as simple as possible because a large amount of data is required to create a complicated model, while the actual amount of data available is very limited. Several simplifications were made in the modeling process and are discussed in Section \ref{section:discussion}.

Two terms used in the modeling procedure are defined. A \textit{point-mass mixture distribution parameter} contains a point-mass (a particular value with more observations than a continuous distribution can describe), which requires a mixture distribution model to describe its distribution. It is generally difficult to model the relationship between this parameter type and some other type.

A correlation coefficient is a numerical measure ranging from -1 to 1 that measures the strength and direction of a linear relationship between two quantitative variables \cite{easterling2010passion}. A large significant absolute coefficient indicates a strong linear relationship between the measured variables. The sign indicates whether the relationship is positive or negative. A \textit{significant and non-weak correlation} is defined as a correlation whose coefficient has a p-value less than 0.05 (significant) and an absolute value greater than or equal to 0.3 (non-weak).

\subsubsection{Procedure}
\begin{figure}[!t]
\centering
\includegraphics[width=0.45\textwidth]{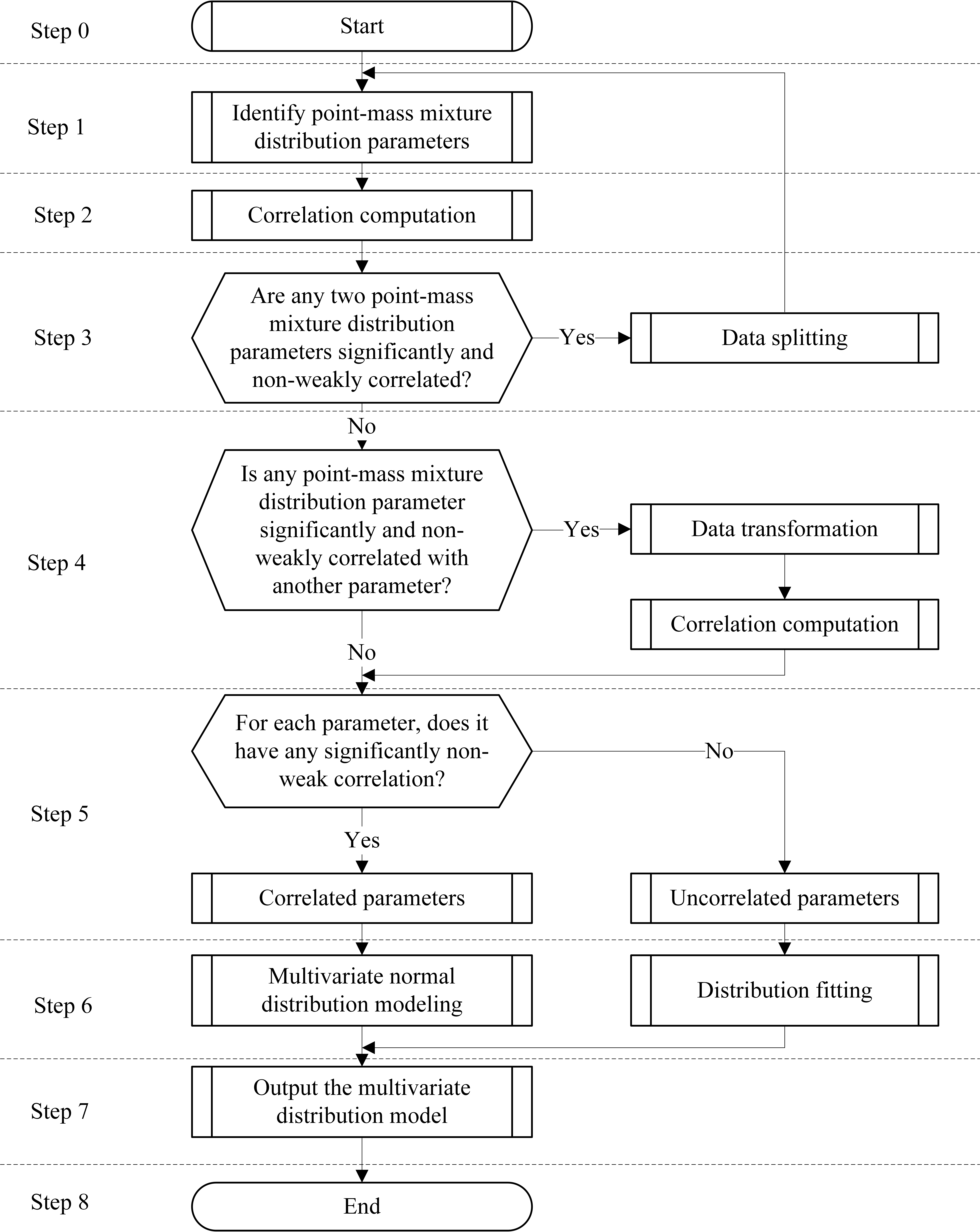}
\caption{The procedure of the multivariate distribution modeling.}
\label{fig:procedure}
\end{figure}

The steps in the procedure for multivariate distribution modeling (shown in Fig. \ref{fig:procedure}) are listed below. More details of some steps are provided in the next subsections.
\begin{list}{}{}
    \item {\bf Step 0:} Start.
    \item {\bf Step 1:} Identify point-mass mixture distribution parameters in the input data.
    \item {\bf Step 2:} Correlation computation. The correlation coefficients between every two parameters are computed. We use the "weights" package in R \cite{weightsrpackage} to compute the weighted (Pearson) correlation coefficients between all two parameter combinations, considering the sample weights.
    \item {\bf Step 3:} Check if any two point-mass mixture distribution parameters are significantly and non-weakly correlated. If so, split the data into two sub-datasets based on whether the parameter equals its point-mass value for either point-mass mixture distribution parameter; then, for each sub-dataset, go through the modeling from Step 1; otherwise, go to Step 4.
    \item {\bf Step 4:} Check if any point-mass mixture distribution parameter is significantly and non-weakly correlated with another parameter. If not, go to Step 5. Otherwise, perform data transformation followed by a correlation computation of the transformed data, then go to Step 5. The data transformation aims to decorrelate the point-mass mixture distribution parameter from any other parameter so that it can be modeled independently.
    \item {\bf Step 5:} Classify each parameter as correlated or uncorrelated. A parameter is categorized as a correlated parameter if it exhibits a significant and strong correlation with any other parameter and as uncorrelated if no such correlation is observed.
    \item {\bf Step 6:} Model the distribution. For correlated parameters, a multivariate normal distribution model is used (Multivariate normal distribution modeling in Fig. \ref{fig:procedure}). The parameters are fitted to their own distributions separately (Distribution fitting in Fig. \ref{fig:procedure}).
    \item {\bf Step 7:} Output the multivariate distribution model for the input data.
    \item {\bf Step 8:} End.
\end{list}

\subsubsection{Data splitting} \label{section:datasplit}
Step 3 performed data splitting if two point-mass mixture distribution parameters have a significant and non-weak correlation. In this way, we built two sub-models for the two sub-datasets instead of building a complicated model for two correlated point-mass mixture distribution parameters. This simplification was performed because it is difficult to model the correlation between two point-mass mixture distribution parameters. In this case, the decorrelation method utilized in Step 4 is ineffective when dealing with the correlation between a regular parameter and a point-mass mixture distribution parameter. Because the point-mass of the point-mass mixture distribution parameter used as the independent variable cannot be produced in the generated data. However, future work should aim to incorporate this modeling aspect.

\subsubsection{Data transformation}
The purpose was to ensure that no parameter is correlated with any point-mass mixture distribution parameter after transformation. For a parameter, $x_i$, that is significantly and non-weakly correlated with any point-mass mixture distribution parameter, the transformed parameter $x^{\prime}_i$ was computed according to
\begin{equation}
    x^{\prime}_{i} = x_{i} - f(X_{pm}),
\end{equation}
where $f(X_{pm})$ is the estimated linear regression model, in which $x_i$ and the vector of all point-mass mixture distribution parameters $X_{pm}$ are the explanatory and dependent variables respectively. Consequently, $x^{\prime}_{i}$ is not correlated with any point-mass mixture distribution parameter.

In contrast, for a parameter that is neither significantly nor non-weakly correlated with any point-mass mixture distribution parameter, it is unchanged after transformation.

\subsubsection{Distribution fitting}
The data for a parameter, which does not include a point mass, was fitted into a set of distributions, including normal, skew-normal, exponential-normal, and gamma distributions, using the maximum likelihood estimation (MLE). Akaike information criterion (AIC) was used to select the best-fitting distribution, the one with the lowest AIC value. AIC is an estimator of prediction error and, thereby, the relative quality of statistical models for a given set of data \cite{stoica2004model}.

For a point-mass mixture distribution parameter $x_j$, a mixture distribution model combining a binomial distribution and a continuous distribution was used. The mixture distribution model, in this case, is a hurdle model \cite{cameron2013regression}, in that all the point-mass values come only from the binomial distribution, while all other values come from the continuous distribution. Therefore, the two sub-distributions were fitted separately using MLE:
\begin{enumerate}
    \item The binomial distribution's estimated success possibility is the point-mass value proportion.
    \item For the continuous distribution model, select successively from the gamma, generalized gamma, and exponential distribution models using AIC.
\end{enumerate}

\subsubsection{Multivariate normal distribution modeling}
This process modeled all $n$ correlated parameters as a multivariate normal distribution according to the following steps.
\begin{enumerate}
    \item Fit the distribution fitting for all correlated parameters with the same method used in the sub-step, distribution fitting.
    \item Use the quantile transformation (also known as quantile mapping) \cite{panofsky1958some} for each parameter to transform the data to the standard normal distribution $\mathcal{N}(0, 1)$.
    \item Compute the covariance matrix $\boldsymbol{\Sigma} \in \mathbf{R}^{n \times n}$ of the normalized parameters and set $\boldsymbol{\Sigma}_{ii}=1\  (i=1,...,n)$.
    \item Build the multivariate normal distribution $\mathcal{N}(\mathbf{M}, \boldsymbol{\Sigma})$, where $\mathbf{M} = (0,...,0)^T$.
\end{enumerate}
The $\boldsymbol{\Sigma}_{ii}$ and $\mathbf{M}$ were set directly because the normalized parameters all belong to $\mathcal{N}(0, 1)$.

\subsection{Data Generation and Filtering} \label{section:datagen}
Synthetic lead-vehicle speed profiles represented by the six parameters were generated in two steps based on each developed multivariate distribution model, which might contain a multivariate normal distribution (for correlated parameters) and several fitted distribution models (for uncorrelated parameters).
\begin{enumerate}
    \item Data generation from the sub-model(s): For correlated parameters, the normalized data is generated from the multivariate normal distribution model. Then, we perform a quantile transformation (an inverse version of the quantile transformation done in the sub-step, multivariate normal distribution modeling) of the normalized data. For every uncorrelated parameter, the data is generated solely from its fitted distribution model.
    \item Inverse transformation: For each parameter, perform the inverse transformation of any transformation conducted in the data transformation sub-step during the modeling.
\end{enumerate}

Three types of constraints were set to remove any invalid generated speed profiles.
\begin{itemize}
    \item Range constraints: Each parameter has its own range limit, such as $\tau_1 \geq 0$ s and $v_c \geq 0$ m/s.
    \item Physical constraints: First, the lead vehicle should not reverse; its speed should be no less than 0 m/s for the whole duration. Second, the physical constraint for vehicle acceleration applied to extracted events is also applied here: $a_i \in [-g, g]\ (i=1,2)$.
    \item Categorization constraints: The whole dataset was categorized into multiple sub-datasets according to certain conditions that lead to the modeling sub-dataset, so the generated data should also fulfill these conditions. Some examples are shown in Section \ref{section:datacate}.
\end{itemize}

\subsection{Validation} \label{section:validation}
A synthetic dataset containing 10,000 synthetic lead-vehicle speed profiles was built by proportionally sampling the speed profiles generated by the distribution model of each sub-dataset. Besides descriptive statistics analysis, non-parametric tests, particularly the weighted two-sample Kolmogorov–Smirnov (KS) tests (using the "Ecume" package in R \cite{ecumerpackage}), were conducted to test whether the synthetic and raw lead-vehicle speed profiles are from different distributions. \textcolor{black}{While lack of significance in the KS test does not mean that the distributions are the same, it does mean that the sample distributions are similar enough that a conclusion of "different" cannot be made with high confidence. Since non-parametric tests generally have lower statistical power (the probability of a test correctly rejecting the null hypothesis) than parametric tests \cite{sullivan2012essentials}, and since similarity is of interest in this application, we adjusted the significance level ($\alpha$) to 0.10 rather than 0.05. Doing so increases power and reduces the probability of a Type II error (a failure to reject a null hypothesis that is actually false) \cite{dybaa2006systematic}.}

\section{Results} \label{section:results}
\subsection{Parameterization of the Lead-vehicle Speed Profile}

\begin{figure}[!t]
    \centering
    \subfloat[]{\includegraphics[width=0.22\textwidth]{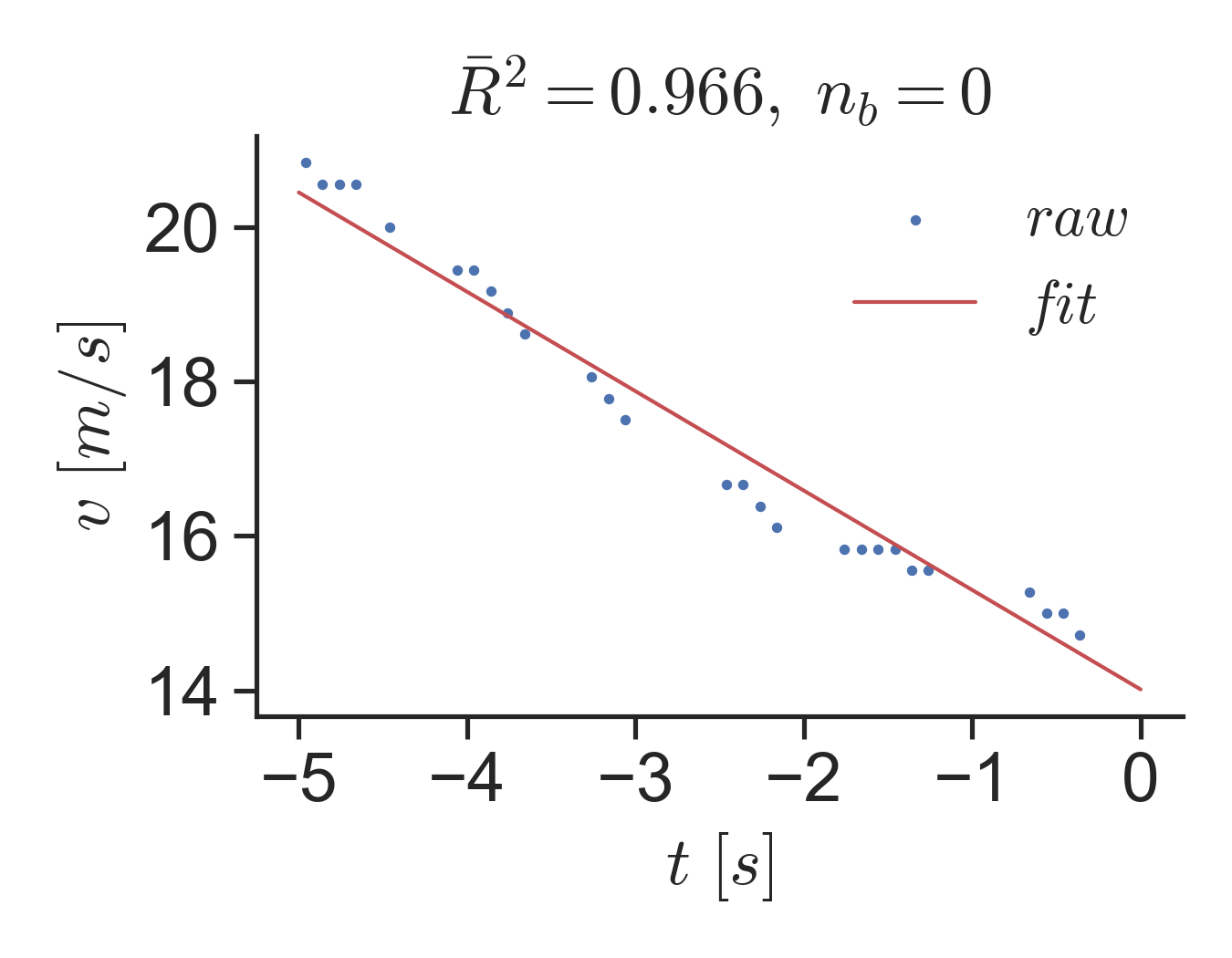}}
    \hfil
    \subfloat[]{\includegraphics[width=0.22\textwidth]{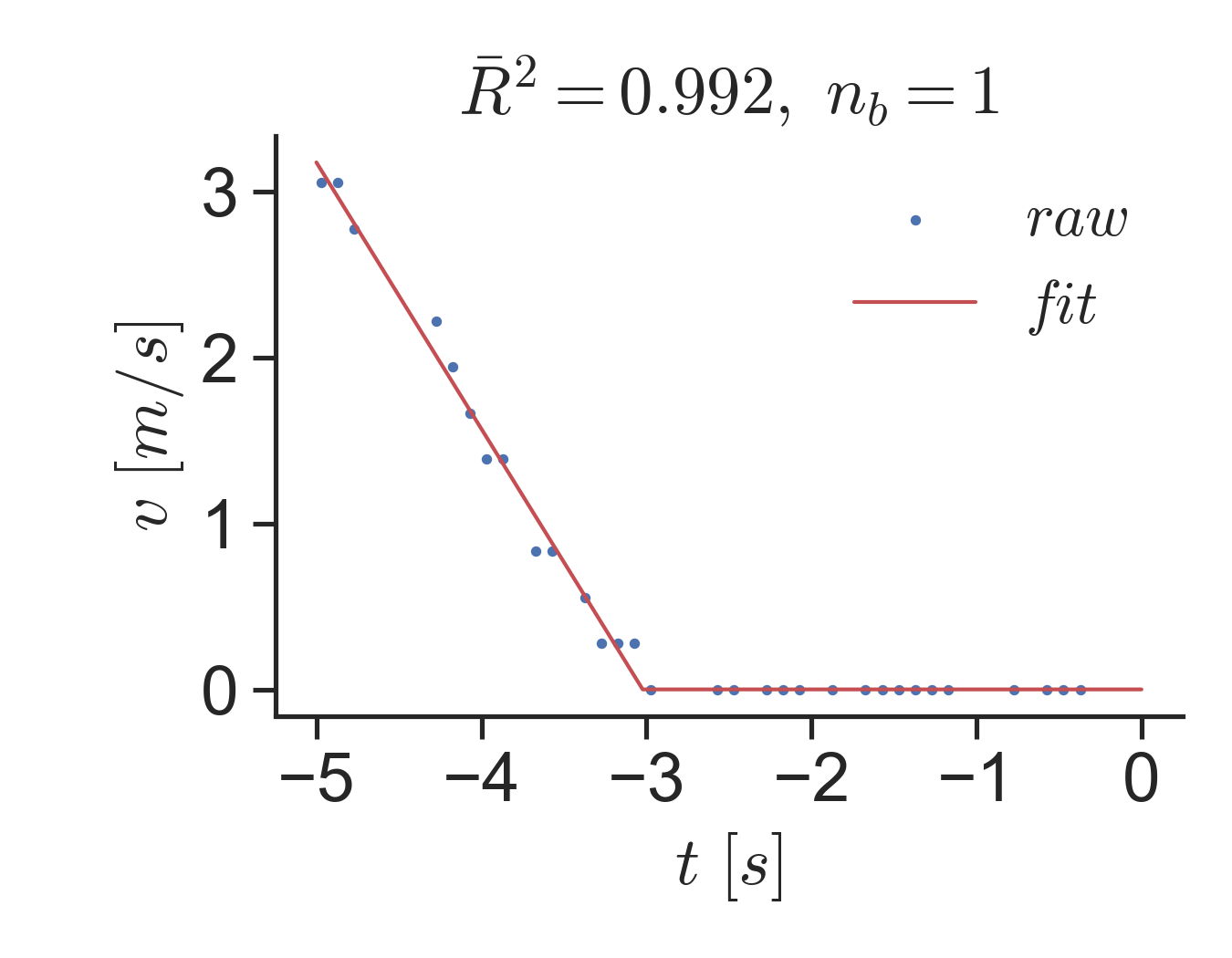}}
    \vfil
    \subfloat[]{\includegraphics[width=0.22\textwidth]{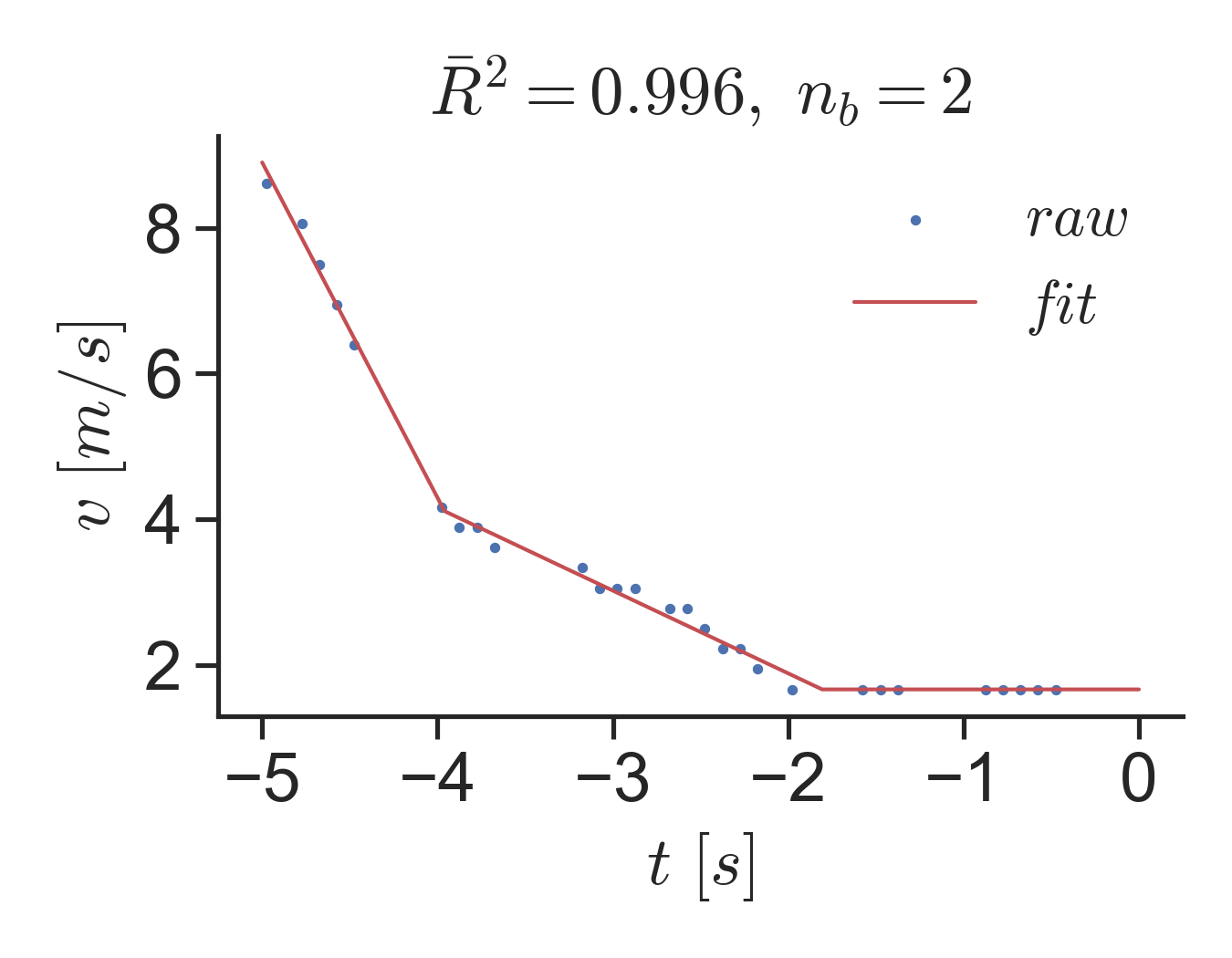}}
    \hfil
    \subfloat[]{\includegraphics[width=0.22\textwidth]{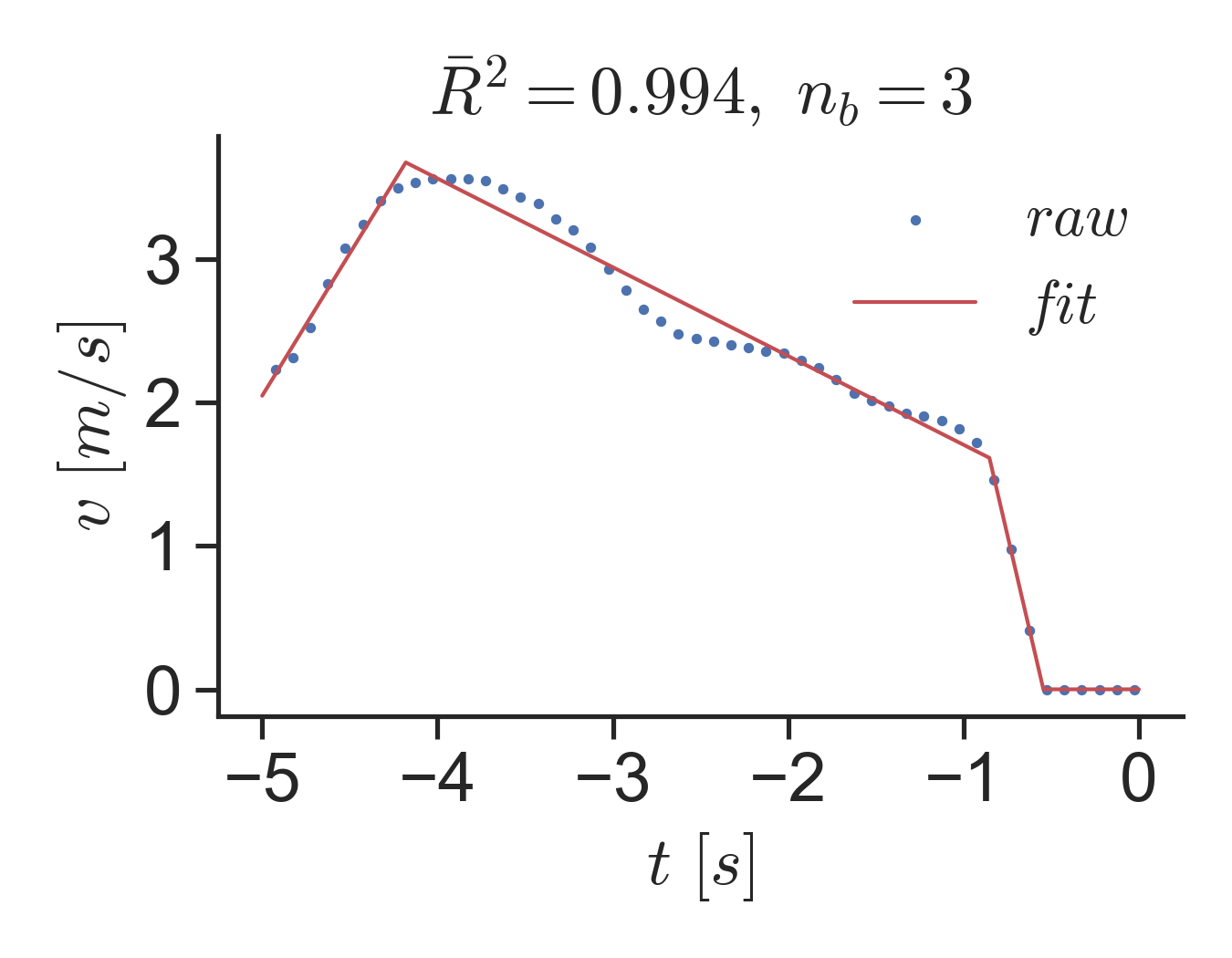}}
    \caption{Examples of fit results. (a)-(d) show the weighted piecewise linear regressions with zero to three breakpoints.}
    \label{fig:examplesoffit}
\end{figure}

All events have a decent fitness level except for those where a constant value regression is performed. 98.3{\%} (298 out of 303) of the events have an adjusted R-squared $\Bar{R}^2$ greater than 0.9. Fig. \ref{fig:examplesoffit} shows several examples of the fit results.

\subsection{Data Combination}
\begin{figure}[!t]
    \centering
    \includegraphics[width=0.35\textwidth]{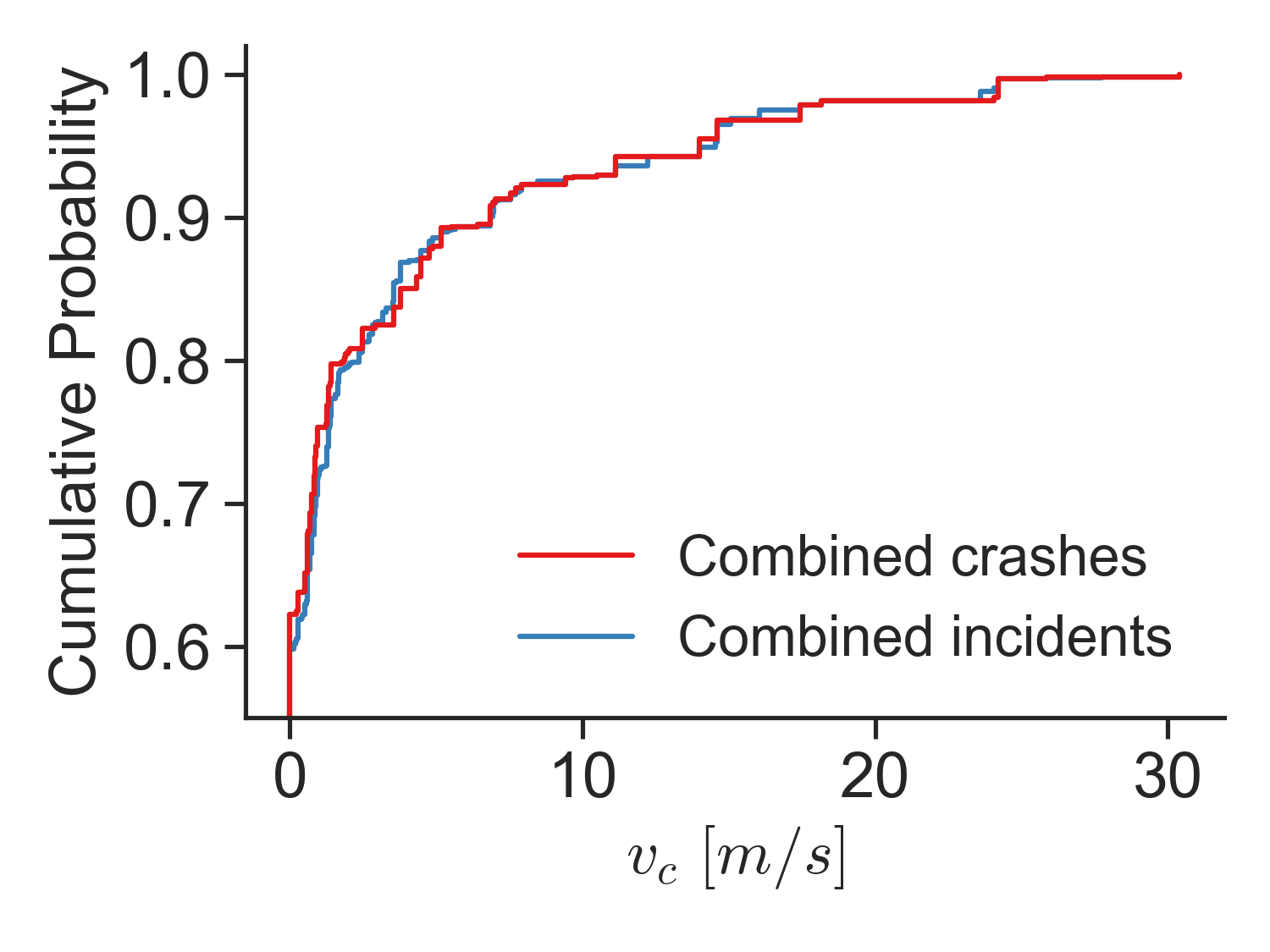}
    \caption{Weighted CDFs of $v_c$ of both the combined crash and combined incident datasets.}
    \label{fig:cdfofallvc}
\end{figure}

Eighty-two near-crashes, 62.1\% of the total number of crashes, are selected as variations of crashes and added to the combined crash dataset, resulting in a sample size of 214 incidents in the combined incident dataset. Table \ref{tab:combineddataset} shows the composition. The weighted CDFs of the six parameters are checked for both the combined crash and combined incident datasets (for instance, the weighted CDFs of $v_c$ are shown in Fig. \ref{fig:cdfofallvc}). \textcolor{black}{The difference between the two datasets regarding a single parameter's marginal distribution is negligible. However, there are noticeable variations in the joint distribution when considering multiple variables, as illustrated in Fig. \ref{fig:jointdisofa1&a2}.} (More details of the comparison can be found in Section III-A of the supplement.)

\begin{table}[!t] 
\caption{Composition of the combined incident dataset\label{tab:combineddataset}} 
\centering
\begin{tabular}{ccc}
\hline
Group & Notation & Sample size\\
\hline
1 & CISS$_{sc}$ & 49\\
2 & SHRP2$_{sc}$ & 20\\
3 & SHRP2$_{nsc}$ & 63\\
4 & SHRP2$_{nc}$ & 82\\
\hline
\end{tabular}
\end{table}

\begin{figure}[!t]
    \centering
    \includegraphics[width=0.35\textwidth]{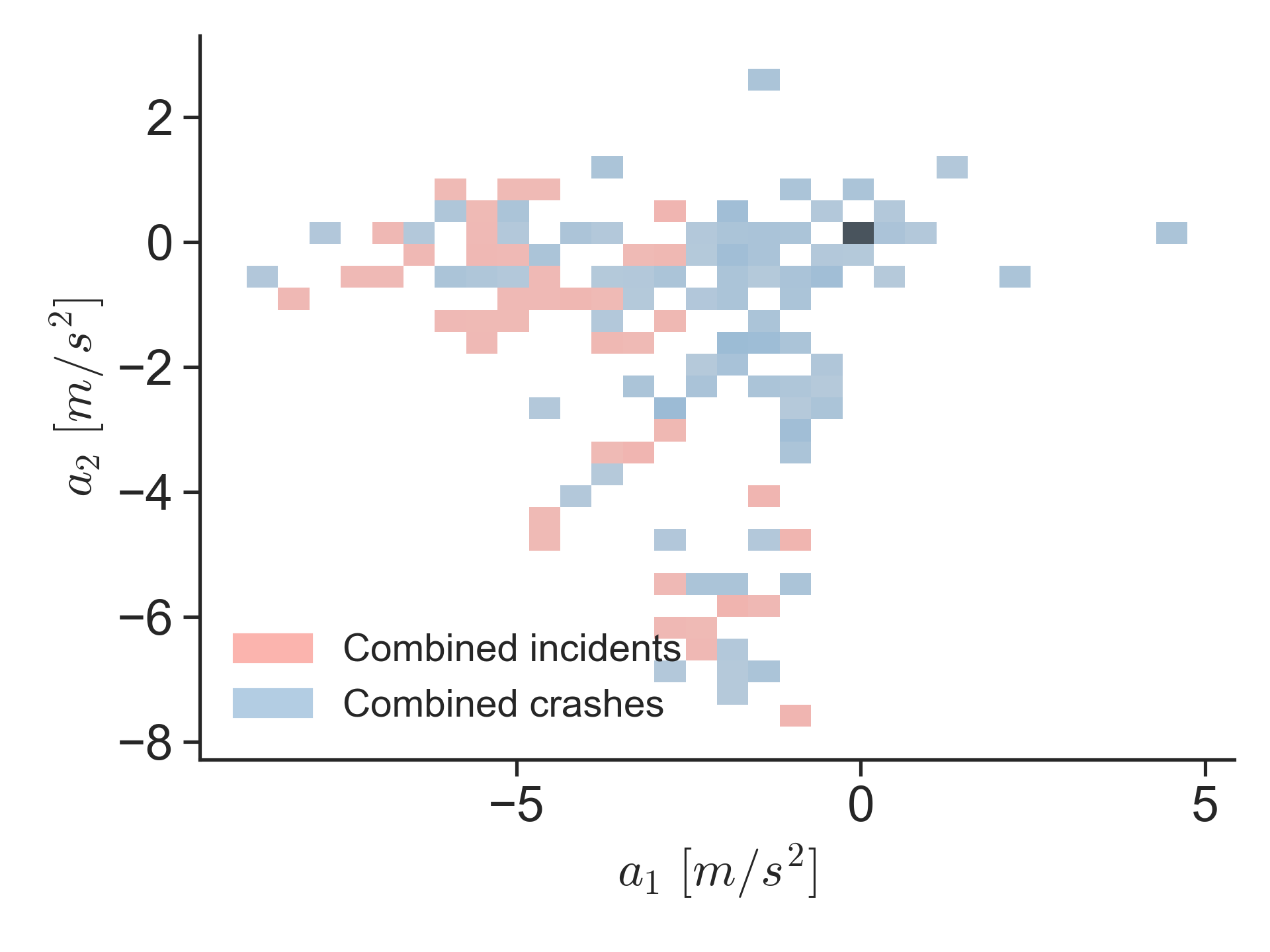}
    \caption{Joint distributions of $a_1$ and $a_2$ of both the combined crash and combined incident datasets. The combined crash dataset is a subset of the combined incident dataset.}
    \label{fig:jointdisofa1&a2}
\end{figure}

\subsection{Data Categorization} \label{section:datacate}
There can be different patterns in the combined incident dataset, and it is difficult to build a comprehensive multivariate distribution model that covers all the data. However, creating sub-datasets allows a simpler model to be applied to each one.

The relationship of each pair of the six parameters in the combined incident dataset was checked. The relationship between $a_2$ and $a_1$, indicating the lead-vehicle speed change trend, shows the most distinct patterns, including the proportions in the combined incident dataset, as follows (see Fig. \ref{fig:threepatterns}).

\subsubsection{Constant acceleration ($a_1 = a_2$, 46.2\%)}
The lead vehicle keeps a constant acceleration until a steady speed is reached or an impact occurs. In this case, Segment 2 is non-existent ($\tau_2 = 0\ s$), and $a_2$ is set to $a_1$.
\subsubsection{Increasing acceleration ($a_1 > a_2$, 20.3\%)}
The lead vehicle increases its acceleration as time goes from Segment 2 to Segment 1. For example, the lead vehicle brakes harshly, followed by gentle braking or even acceleration.
\subsubsection{Decreasing acceleration ($a_1 < a_2$, 33.5\%)}
The lead vehicle decreases its acceleration from Segment 2 to Segment 1. For example, the lead vehicle accelerates first and then starts to brake harshly.

\begin{figure}[!t]
\centering
\includegraphics[width=0.4\textwidth]{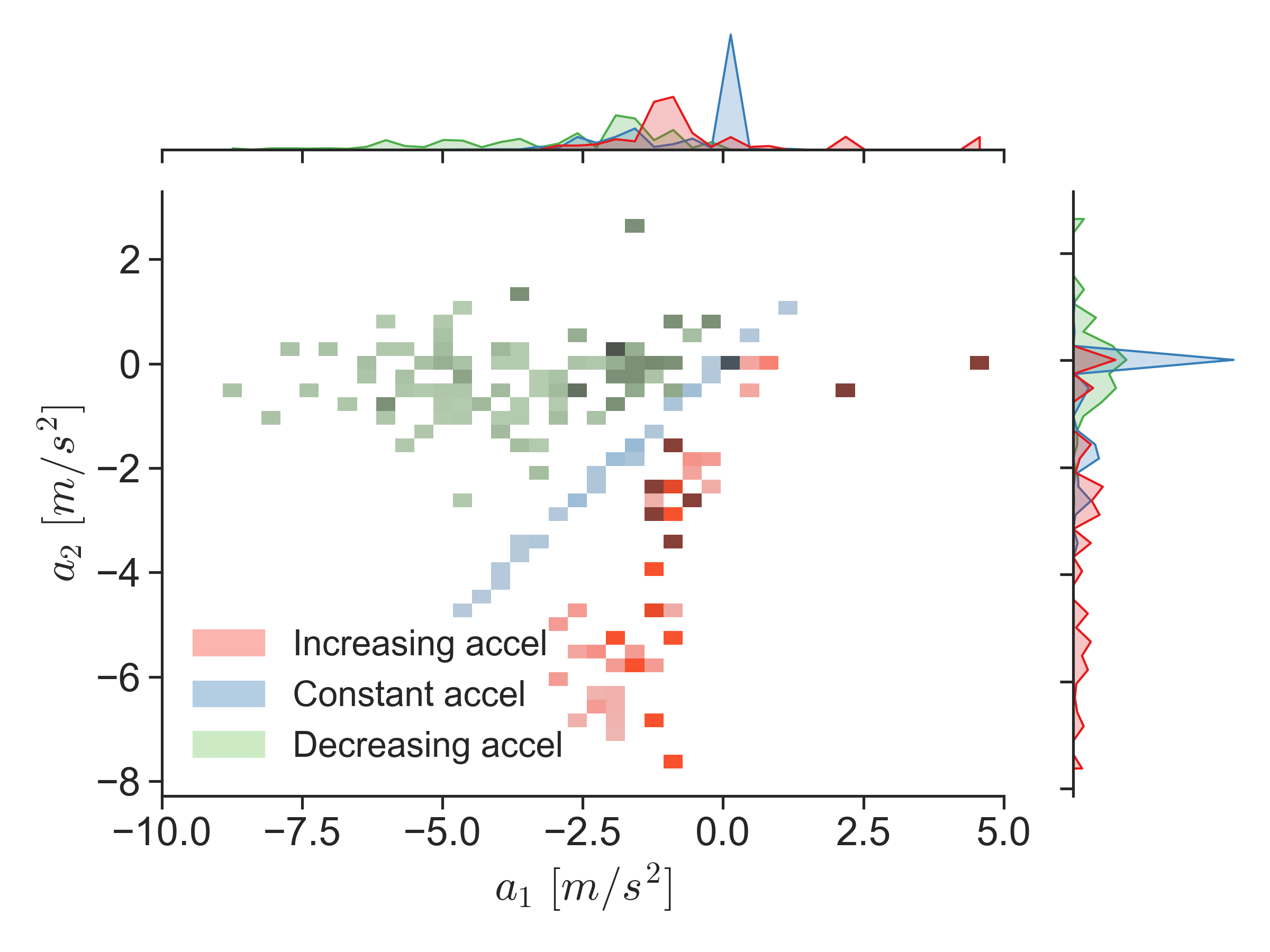}
\caption{Three patterns of the lead-vehicle speed change trend.}
\label{fig:threepatterns}
\end{figure}

However, the data of each pattern can be appropriately modeled if they are further categorized into smaller sub-datasets. In this case, there were seven sub-datasets (S1-S7), and we needed to model all of them except S1. All the sub-datasets, including the proportions in the combined incident dataset, are listed below.

According to the number of actual parameters, the data of the constant acceleration pattern were divided into the following three sub-datasets:
\begin{itemize}
    \item S1: Standstill; 25.5\%.
    \item S2: Constant acceleration; 10.5\%.
    \item S3: Constant non-zero acceleration then steady speed; 10.2\%. 
\end{itemize}

The increasing acceleration pattern data contains two sub-patterns depending on whether or not the lead vehicle is decelerating in Segment 1, as shown in Fig. \ref{fig:threepatterns}. Therefore, the data were divided into two sub-datasets:
\begin{itemize}
    \item S4: Increasing acceleration and $a_1 < 0\ m/s^2$; 15.7\%.
    \item S5: Increasing acceleration and $a_1 > 0\ m/s^2$; 4.6\%.
\end{itemize}

Finally, the decreasing acceleration pattern data contains two significantly and non-weakly correlated point-mass mixture distribution model parameters, $v_c$ and $\tau_s$. In the multivariate distribution modeling, the data of the decreasing acceleration pattern were divided into two sub-datasets based on whether $\tau_s$ equals its point-mass value, 0 s, or not:
\begin{itemize}
    \item S6: Decreasing acceleration and $\tau_s = 0\ s$; 13.3\%.
    \item S7: Decreasing acceleration and $\tau_s > 0\ s$; 20.2\%.
\end{itemize}

As mentioned in \ref{section:datagen}, along with two other constraints, categorization constraints were used to remove invalid speed profiles. For example, for sub-dataset S4, the categorization constraints were: $a_1 > a_2$ (increasing acceleration), and $a_1 < 0\ m/s^2$.

\subsection{Comparison between the Synthetic and Raw Incidents}

\begin{table}[!t] 
\caption{Comparison between the raw and synthetic incidents \label{tab:comparison}} 
\centering
\begin{threeparttable}
\begin{tabular}{cccccccc}
\hline
\multirow{3}{*}{Parameter} & \multirow{3}{*}{Unit} & \multicolumn{2}{c}{Raw} & \multicolumn{2}{c}{Synthetic} & \multirow{3}{*}{statistic} & \multirow{3}{*}{p-value}\\
& & \multicolumn{2}{c}{($n=132^a$)} & \multicolumn{2}{c}{($n=10,000$)}\\
\cmidrule(rl){3-4} \cmidrule(rl){5-6}
& & Mean & SD & Mean & SD & \\
\hline
$v_c$ & $m/s$ & 2.01 & 4.69 & 1.79 & 3.91 & 0.05 & 0.98\\
$a_1$ & $m/s^2$ & -1.37 & 1.82 & -1.57 & 1.71 & 0.10 & 0.25\\
$a_2$ & $m/s^2$ & -0.95 & 1.72 & -1.04 & 1.69 & 0.07 & 0.75\\
$\tau_s$ & $s$ & 1.73 & 2.07 & 1.71 & 2.06 & 0.03 & 0.99\\
$\tau_1$ & $s$ & 1.98 & 1.64 & 1.95 & 1.62 & 0.04 & 0.99\\
$\tau_2$ & $s$ & 1.18 & 1.30 & 1.18 & 1.28 & 0.05 & 0.99\\
\hline
\end{tabular}
\begin{tablenotes}
\RaggedRight
\item $^a$ Valid sample size = sum of sample weights.
\end{tablenotes}
\end{threeparttable}
\end{table}

\begin{figure}[!t]
    \centering
    \includegraphics[width=0.35\textwidth]{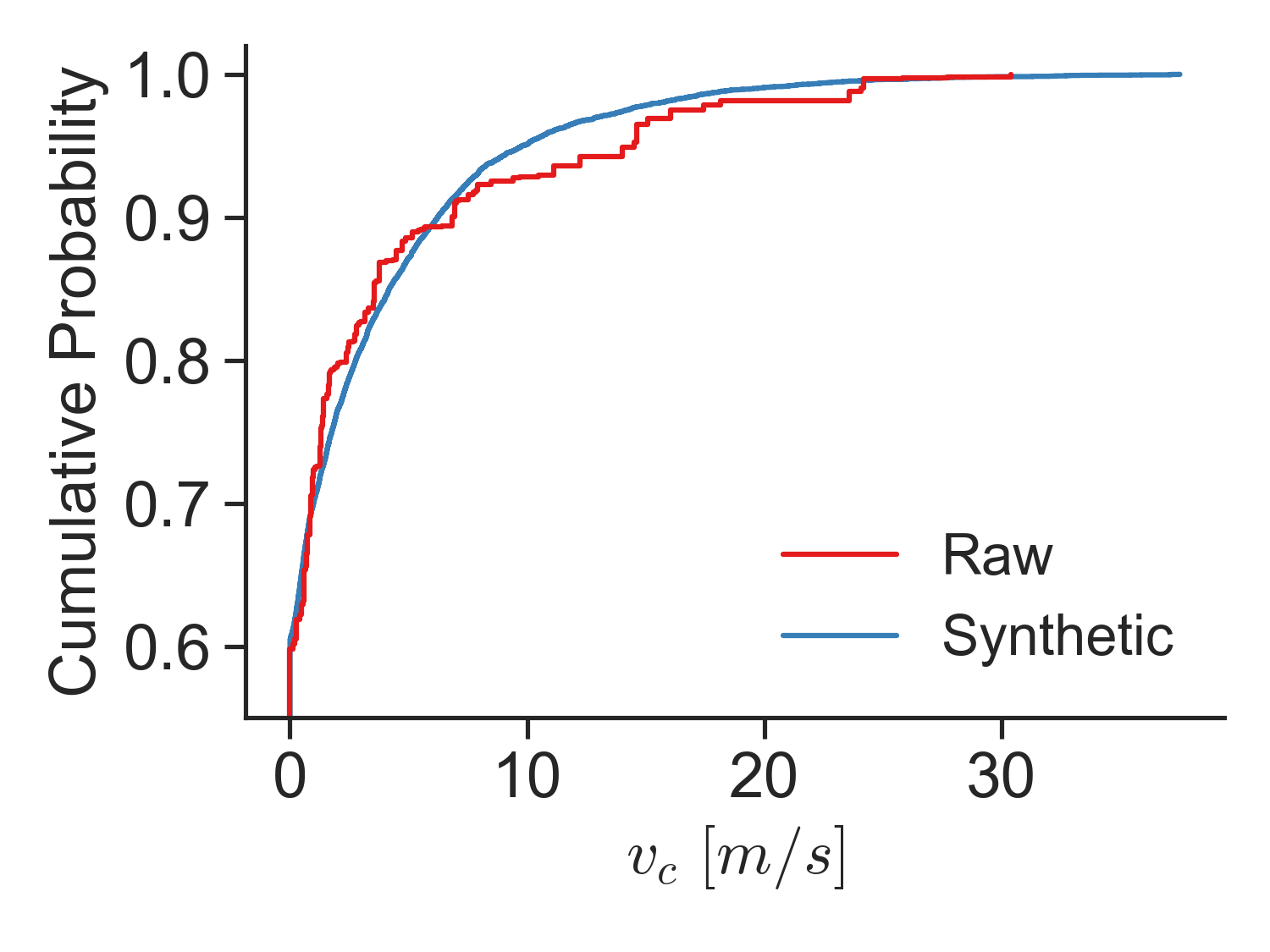}
    \caption{Weighted CDFs of $v_c$ of the raw and synthetic incidents. The raw incidents are properly weighted to be combined into one dataset. All synthetic incidents have a weight of 1.}
    \label{fig:compvc}
\end{figure}

\begin{figure}[!t]
\Centering
\includegraphics[width=0.4\textwidth]{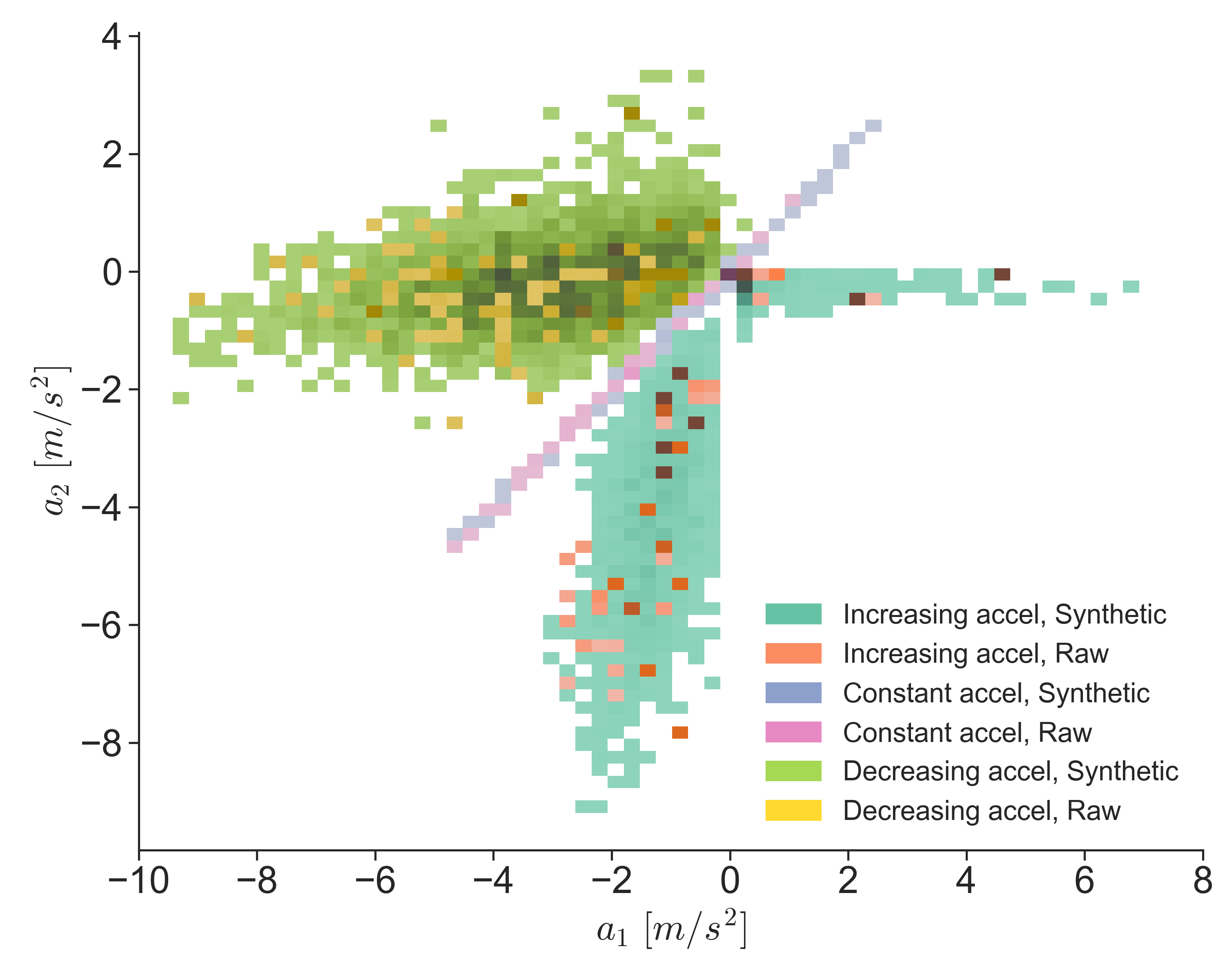}
\caption{Comparison between the raw and synthetic incidents: joint distribution of $a_1$ and $a_2$.}
\label{fig:threepatternscomparision}
\end{figure}

\begin{figure}[!t]
\Centering
\includegraphics[width=0.35\textwidth]{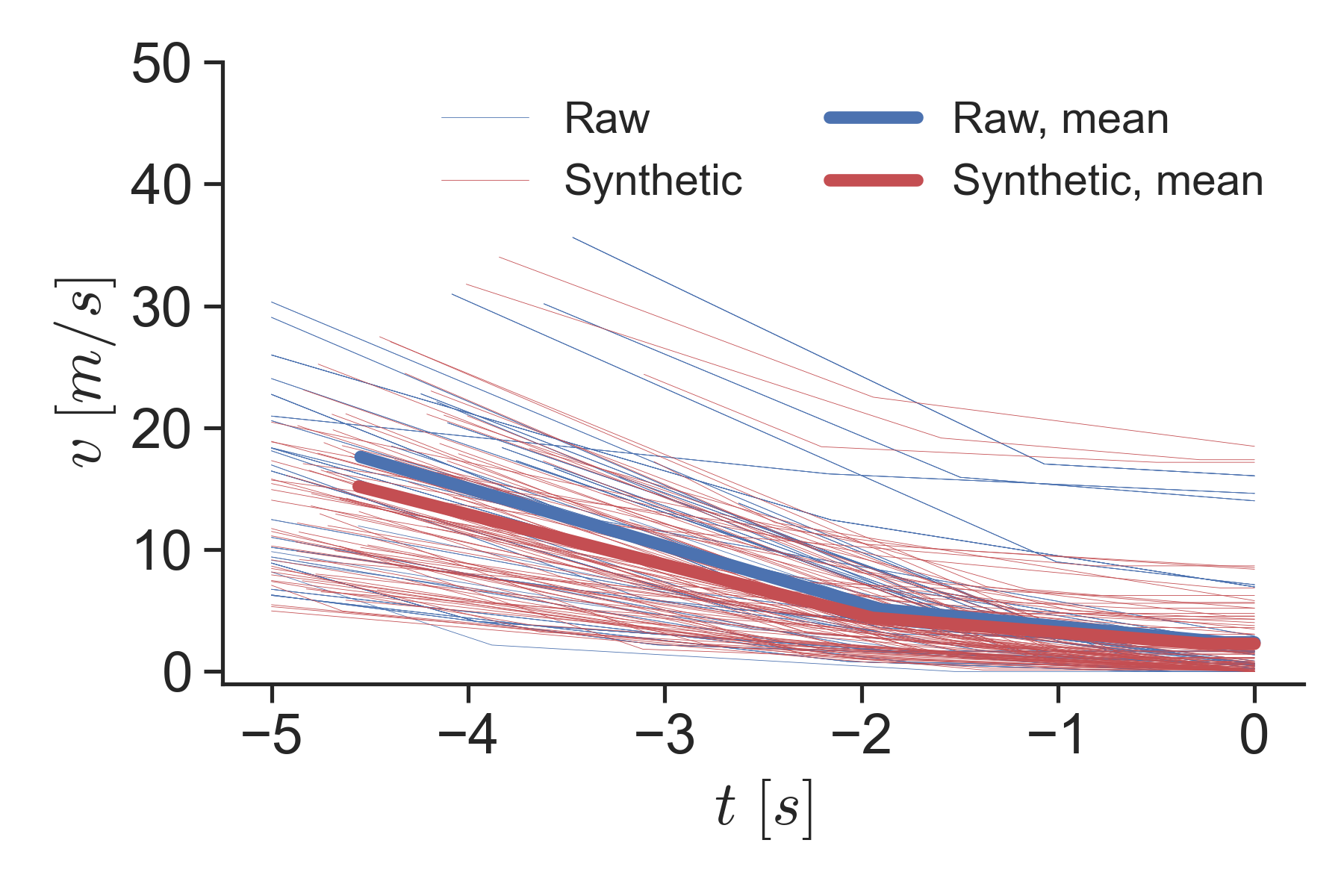}
\caption{Lead-vehicle speed profiles from the raw and synthetic incidents of sub-dataset S4 (Increasing acceleration and $a_1 < 0\ m/s^2$). The bold lines are with the weighted mean values of parameters describing the speed profiles. The thin lines are 100 randomly sampled profiles for the raw and synthetic incidents respectively.}
\label{fig:s4speedprofile}
\end{figure}

\begin{figure}[!t]
\Centering
\includegraphics[width=0.45\textwidth]{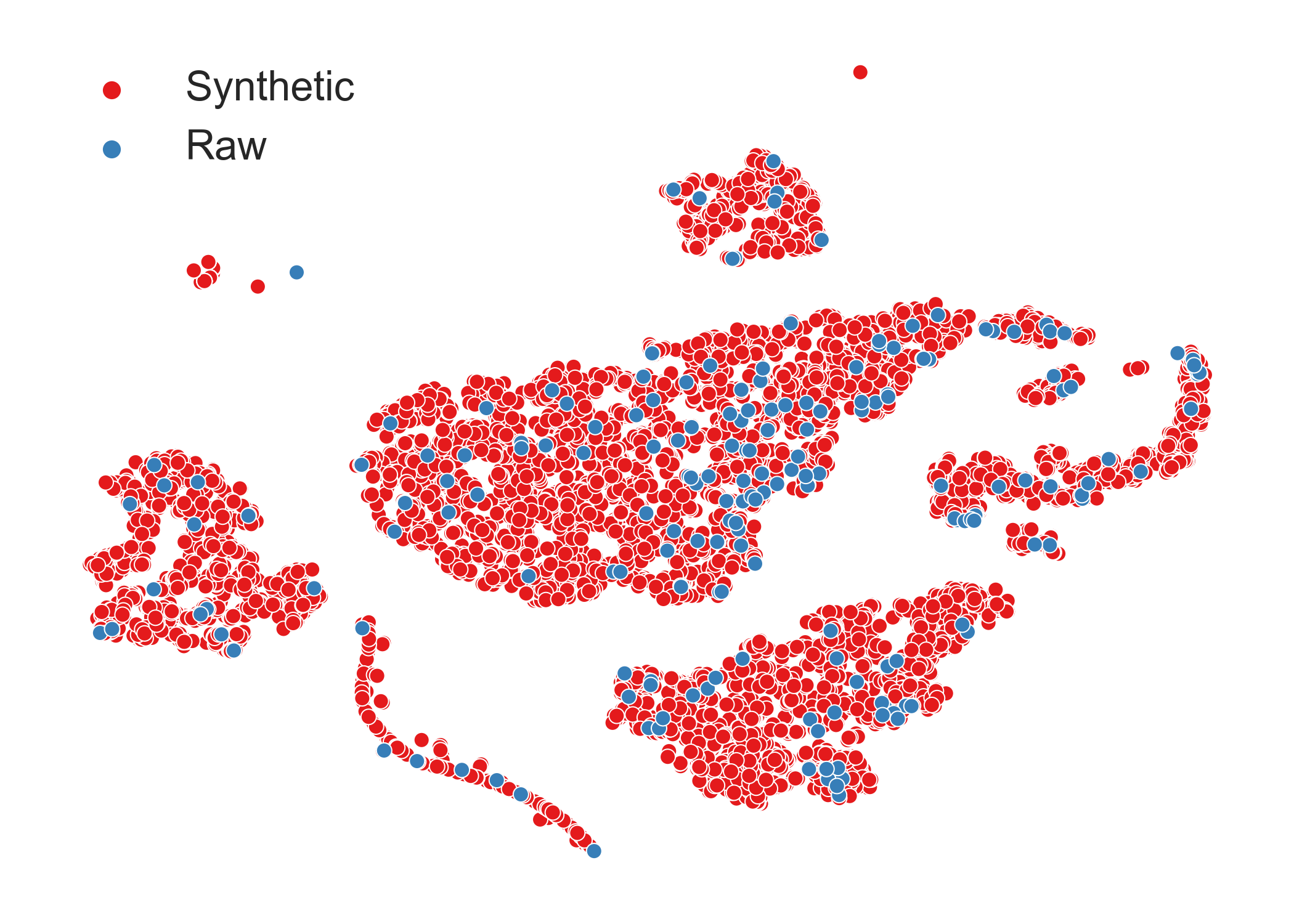}
\caption{t-SNE projection of the raw and synthetic incidents.}
\label{fig:tsne}
\end{figure}

Table \ref{tab:comparison} compares the six parameters for the synthetic incident dataset and the combined (raw) incident dataset. There are minor differences between the raw and synthetic incidents for each parameter regarding the weighted mean and standard deviation (SD). Furthermore, the two datasets were subjected to weighted Kolmogorov-Smirnov tests to assess whether there are any significant differences in each of the six parameters. The results, presented as p-values in the table, do not indicate any significant difference. However, it is important to note that the lack of significance does not necessarily imply that the datasets are from the same distribution. Despite this, the visual comparison of the well-aligned weighted cumulative distribution functions (CDFs) for each of the six parameters in the two datasets (an example of $v_c$ is illustrated in Fig. \ref{fig:compvc}) reveals substantial similarities. (More details of the weighted CDFs of each parameter can be found in Section III-B of the supplement.)

Moreover, the joint distributions of every two parameters and the raw and synthetic speed profiles of every sub-dataset were compared. For instance, Fig. \ref{fig:threepatternscomparision} shows the joint distribution of $a_1$ and $a_2$, and Fig. \ref{fig:s4speedprofile} shows the comparison results for S4.

As a final step, t-distributed stochastic neighbor embedding (t-SNE) was used to visualize the raw and synthetic incidents in two dimensions; see Fig. \ref{fig:tsne}. t-SNE is a statistical method for visualizing high-dimensional data by giving each data point a location in a two or three-dimensional map \cite{van2008visualizing}. In Fig. \ref{fig:tsne}, the blue dots (projection of raw incidents) are surrounded by the red dots (projection of synthetic incidents).

In summary, the synthetic and raw incidents are similar and well-aligned. (More details regarding the comparison can be found in Section III-B of the supplement.)

\section{Discussion and Conclusions} \label{section:discussion}
This study focuses on the lead vehicle's behavior in rear-end crashes, which is mostly independent of the following vehicle's behavior. Thus, this study models the lead-vehicle speed profile without considering its interaction with the following vehicle.

\textcolor{black}{A piecewise linear model was used to represent the lead-vehicle speed profile in the pre-crash phase, providing a more accurate digital representation of the lead-vehicle kinematics than the conventional constant acceleration/deceleration model. Two datasets (CISS and SHRP2) were combined to produce a comprehensive rear-end critical incident (crash/near-crash) dataset that captures the full severity range. Multivariate distribution models were constructed to generate synthetic lead-vehicle speed profiles that were compared with the raw speed profiles.}

\textcolor{black}{The results show that the piecewise linear model has good fitting performance. The raw and synthetic incidents display a notable alignment. Moreover, a range of different lead-vehicle speed patterns were revealed, indicating the proposed piecewise linear model's greater accuracy compared to the conventional constant acceleration/deceleration model. For example, the lead vehicle could exhibit harsh braking followed by gentle braking (as shown in Fig. \ref{fig:s4speedprofile}) or even acceleration. In addition, the lead vehicle does not necessarily brake harshly. In fact, in many cases, the lead vehicle keeps a constant speed or is at a standstill for a considerable time (up to five seconds) prior to the crash.}

\textcolor{black}{In summary, the proposed model accurately matches lead-vehicle kinematics from in-depth pre-crash/near-crash data across the full severity range, outperforming previously existing lead vehicle models in terms of both severity range and precision. Furthermore, in addition to generating simulated rear-end crash scenarios, this model has the potential to aid substantially in the reconstruction of individual real-world crashes. That is, by offering more realistic speed profiles for reconstructed crashes (considering the speed at impact and other constraints as discussed in Section \ref{section:application}), the model provides a means of generating a distribution of possible speed profiles during the reconstruction process instead of providing only a single speed profile.}

\subsection{Robustness of the Results}

\begin{table}[!t]
    \caption{Results of bootstrapping}
    \centering
    \begin{tabular}{ccc}
    \hline
    \multirow{2}{*}{Parameter} & \multicolumn{2}{c}{p-value} \\
    \cmidrule(rl){2-3}
    & 90\% samples & 80\% samples \\
    \hline
    $v_c$ & 0.98 & 0.94\\
    $a_1$ & 0.83 & 0.63\\
    $a_2$ & 0.96 & 0.93\\
    $\tau_s$ & 0.80 & 0.75\\
    $\tau_1$ & 0.96 & 0.91\\
    $\tau_2$ & 0.73 & 0.67\\
    \hline
    \end{tabular}
    \label{tab:bootstrapping}
\end{table}

\textcolor{black}{When conducting a study with a relatively small sample size, it is crucial to examine the robustness of the results. To do so, we conducted a bootstrapping study to test the multivariate distribution modeling method. This process is outlined below:
\begin{enumerate}
    \item Create 200 new datasets by randomly sampling 100 times (without replacement) from all samples with a sample size reduction of 10\% and 20\%, respectively.
    \item For every new dataset, go through the multivariate distribution modeling process and generate a new synthetic dataset with a sample size of 1,000.
    \item Perform two-sample KS tests for each of the six parameters to compare each of the bootstrapped synthetic datasets with the original synthetic dataset generated using all samples.
    \item Compute the p-value of bootstrapping, which represents the proportion of bootstrap samples for which the KS test is not significant ($p > 0.1$) \cite{efron1994introduction}.
\end{enumerate}
The results do not show any significance (all p-values are larger than 0.1), as indicated in Table \ref{tab:bootstrapping}. Hence, the robustness of the proposed modeling method was demonstrated.}

\subsection{Sampling Bias of Driver Age in the SHRP2 Dataset}
In addition to the sampling bias in near-crashes mentioned in Section \ref{section:datacombination}, there is also a driver-age sampling bias in the SHRP2 dataset. Both young and old drivers are over-represented \cite{flannagan2019replacement}. We investigated the possible impact of this bias and came to the conclusion that it can be ignored in this study. (More details are provided in Section IV of the supplement.)

\subsection{Adding Selected Near-crashes as Variations of Crashes}
In this study, near-crashes similar to a crash in the combined crash dataset were selected and added as variations of crashes with sample weighting adjustment. There are two rationales for doing so.

First, the lead-vehicle speed profiles in a near-crash and a crash can be similar. Given the same lead-vehicle behavior, a rear-end near-crash incident can easily turn into a crash if the following vehicle’s driver reacts more slowly or brakes less harshly.

Second, the sample weighting adjustment practically mitigates the risk that the added near-crashes will change the raw distributions. With the sample weighting adjustment mentioned in Section \ref{section:methodology}, the raw parameter distributions of the combined crash dataset are retained in the new, combined incident dataset. At the same time, with the added samples, there are more observed values. Thus, a more reliable distribution modeling can be achieved.

It is also worth mentioning the alternative to weigh the six parameters (based on prior knowledge) when computing the Euclidean distance between two events. For instance, $v_c$ could have a larger weight than others because it directly relates to the impact result. Future work should address such a weighting method.

\subsection{Limitations}
\textcolor{black}{In addition to the issue with correlated point-mass mixture distribution parameters (as discussed in Section \ref{section:datasplit}) and the reduced statistical power of the non-parametric tests used (as outlined in Section \ref{section:validation}), the following limitations are noteworthy:
\begin{enumerate}
    \item Only the kinematics of the lead vehicle are considered in this work. Numerous variables can influence the occurrence of a crash, including road structure, traffic signals, and weather conditions. Future research should address these considerations when more comprehensive data are available. Nonetheless, a precise description of lead-vehicle kinematics by utilizing diverse data sources and considering crashes occurring in various situations is instrumental in capturing an important part of the overall variability observed in real-world rear-end crashes. 
    \item The modeled lead vehicle's acceleration is not consistently smooth. This could be attributed to the fact that the speed of the lead vehicle is modeled using a piecewise linear model, resulting in a sudden change in acceleration as it moves from one segment to another. Future work should aim to smooth the acceleration profile, potentially by introducing jerk during transitions.
    \item The method of multivariate distribution modeling only considers the linear correlation between two parameters and disregards any potential nonlinear relationship between them, as well as weak or non-significant correlations. In addition, the correlated parameters are assumed to follow a multivariate normal distribution, which effectively models the parameters as linearly related. These simplifications are made to keep the model tractable and avoid over-interpreting the relationships between parameters, as it is not feasible to create a complex multivariate model with a small dataset without a substantial risk of overfitting. These simplifications may, however, reduce the accuracy of the model. Unfortunately, it is not possible to investigate the consequences with the available data, but future work should address this issue.
\end{enumerate}}

\subsection{Application} \label{section:application}
\subsubsection{Data combination method}
The proposed data combination method combines rear-end crashes from two datasets and includes selected rear-end near-crashes from the SHRP2 dataset as variations of crashes. This method is generic and can be adapted to other situations, such as combining multiple crash datasets of other crash scenarios. It is also important to mention that near-crashes are used as substitutes for crashes because of their strong connection and similarities. When applying this method, we need to ensure the data to be added can be used as substitutes.

\subsubsection{Multivariate distribution modeling method}
The multivariate distribution modeling method proposed in this research can be easily adapted to other situations where building a distribution model from a relatively small dataset is needed and an understanding of the underlying distribution is available. For instance, this method can be used to analyze other crash scenarios.

\subsubsection{Synthetic data}
The synthetic data generated in this study can be useful in both rear-end crash reconstructions and safety assessments of ADAS and AD.
\begin{itemize}
\item{For rear-end crash reconstructions, despite a relatively accurate estimation of the impact speed, it is rarely possible to reconstruct the speed profile of the vehicles during the pre-crash phase if no recorded data are available. Post-crash interviews and evidence from the on-scene investigation may be the only source of information. Usually, the lead vehicle would be assumed to be moving with a constant acceleration/deceleration before the crash when no information to the contrary is available. Synthetic data can provide alternative speed profiles given the speed at impact and other available constraints. The use of synthetic data will make the reconstruction easier and more reliable since they can provide prior knowledge of the lead-vehicle speed profile based on actual collected pre-crash data.}
\item{For the safety assessments of ADAS and AD, the synthetic data can be used to create virtual crashes for testing whether the crash can be avoided with a given ADAS or AD technology.}
\end{itemize}

\section{Future Work} \label{section:futurework}
This study is the first step in generating rear-end crash scenarios for the safety assessments of ADAS and AD. Future work will use models of the following vehicle's behavior together with the lead-vehicle kinematics model from this work to generate rear-end crash scenarios.

After completing these steps for the rear-end crash scenario, we will move on to other crash scenarios. Moreover, the parameterized data for the additional scenarios will be added to the same online combined incident dataset.

\bibliographystyle{IEEEtran}
\bibliography{Main}
\begin{IEEEbiography}[{\includegraphics[width=1in,height=1.25in,clip,keepaspectratio]{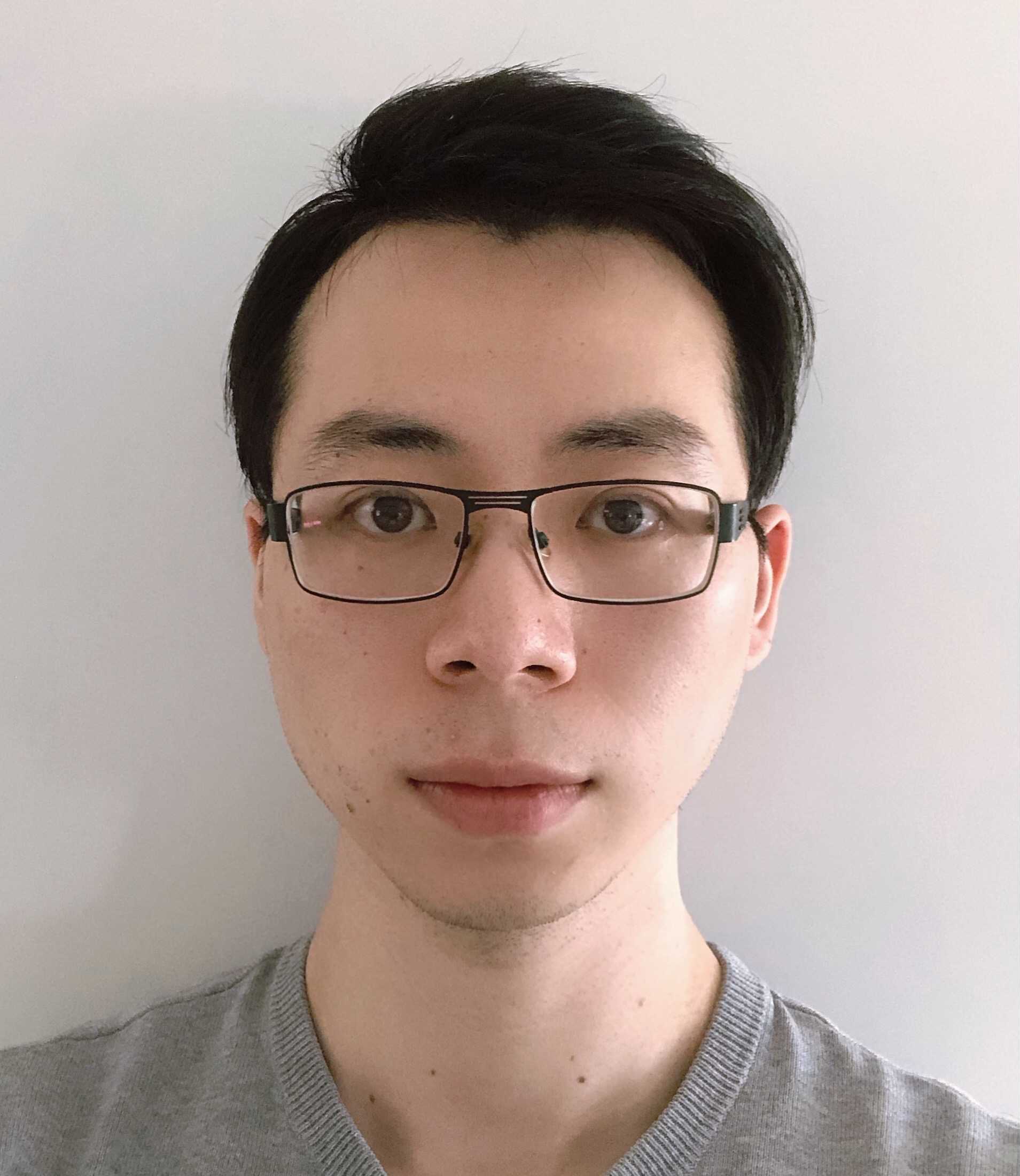}}]{Jian Wu}
received his B.S. and M.S. degrees in automotive engineering from Tsinghua University, Beijing, China, in 2013 and 2016. He is now an industrial Ph.D. candidate with Volvo Cars Safety Center and the Department of Mechanics and Maritime Sciences, Chalmers University of Technology, Göteborg, Sweden. He is the author or coauthor of four journal papers and two conference papers. His current research interests include driver behavior modeling, crash data synthesis, and safety assessments of ADAS and AD.
\end{IEEEbiography}

\begin{IEEEbiography}[{\includegraphics[width=1in,height=1.25in,clip,keepaspectratio]{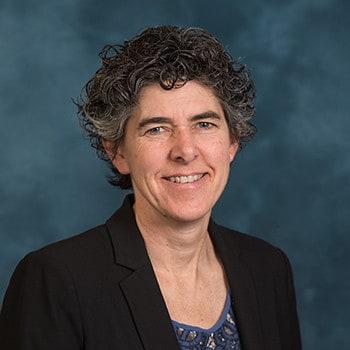}}]{Carol Flannagan}
received an M.A. in Statistics and a Ph.D. in Mathematical Psychology from the University of Michigan. She is a Research Professor at the University of Michigan Transportation Research Institute (UMTRI) in Ann Arbor, Michigan, USA, and an Affiliated Associate Professor at Chalmers University, Göteborg, Sweden. Her work in transportation research encompasses the analysis of a wide variety of transportation-related data and the development of innovative statistical methods for transportation research. She is currently working on a number of projects related to safety assessment and benefits assessment for advanced technologies, including ADS.
\end{IEEEbiography}

\begin{IEEEbiography}[{\includegraphics[width=1in,height=1.25in,clip,keepaspectratio]{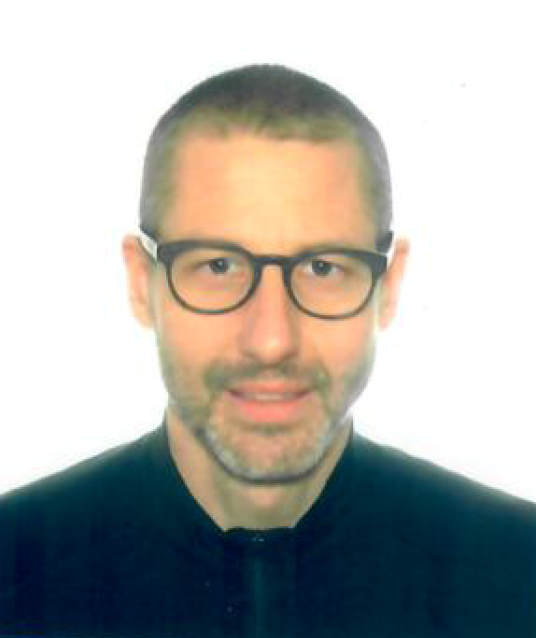}}]{Ulrich Scander}
received his B.S. degree in Biomedical Engineering from the University of Aachen, Germany, in 1997 and his M.S. degree in Accident Research from Graz University of Technology, Austria, in 2008. He received his Ph.D. in Machine and Vehicle Systems from Chalmers University of Technology, Gothenburg, Sweden. He has worked for over 20 years in different positions, such as data analyst and senior principal researcher at Autoliv Research in Germany and Sweden. Since 2022 he has been a Technical Expert at the Safety Centre of Volvo Cars, leading the analysis of field data with a focus on crashes and their consequences.
\end{IEEEbiography}

\begin{IEEEbiography}[{\includegraphics[width=1in,height=1.25in,clip,keepaspectratio]{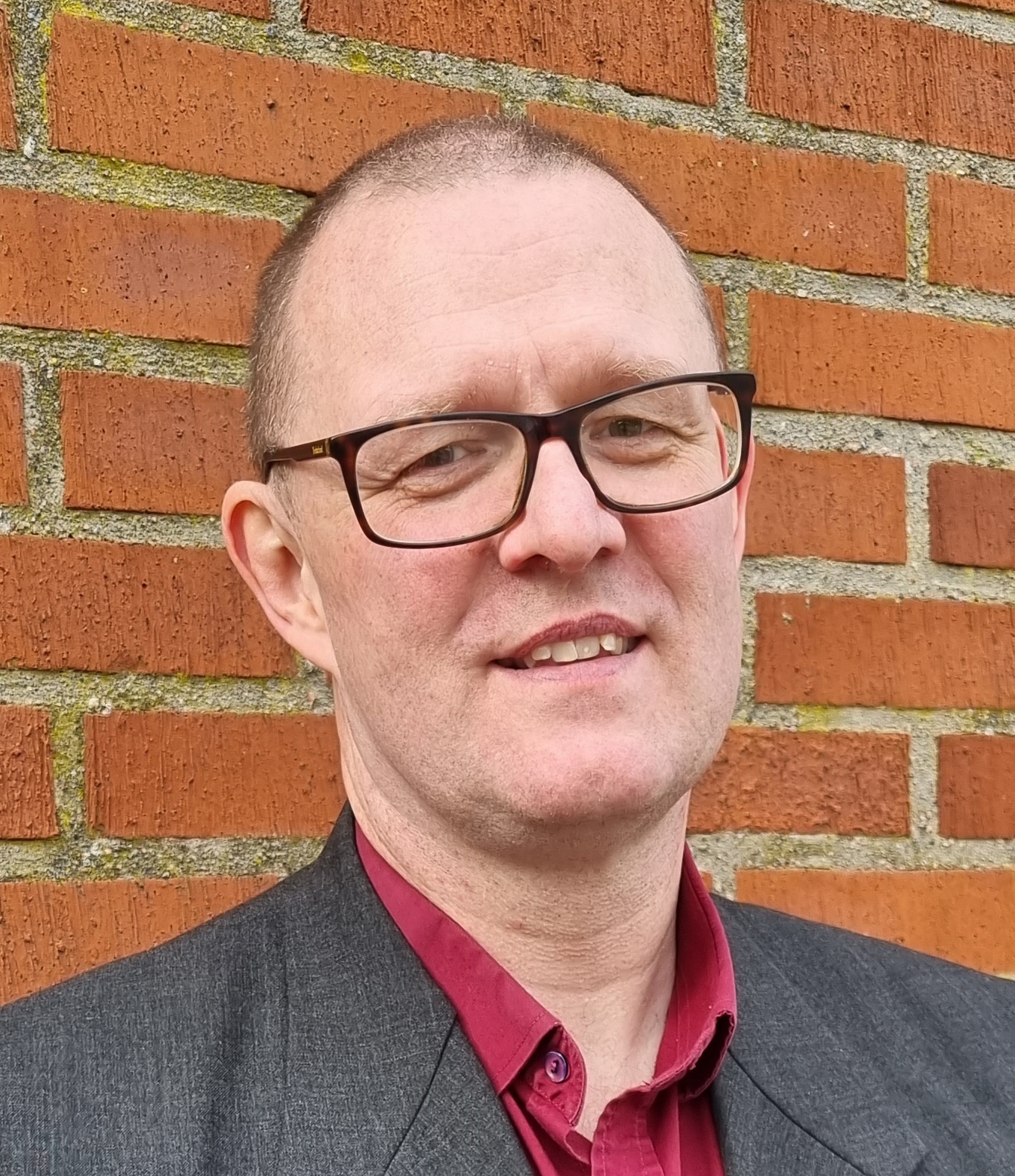}}]{Jonas Bärgman}
received his M.S. degree in Mechanical Engineering at Chalmers University of Technology, Gothenburg, Sweden, in 1997. After his degree, he worked as an industry researcher in in-crash safety at Autoliv Research for three years and as a software developer at AB Volvo for two years. At this point, he continued his career at Autoliv Research (again) in the domain of pre-crash safety, focusing on human factors and driver behavior. In 2009 he started working at Chalmers University of Technology to build a research group on Active Safety. He received his Ph.D. in Machine and Vehicle Systems from the same university in 2016 and is currently an Associate Professor. He is also the examiner for the course “Vehicle and Traffic Safety” in the master’s degree program in Mobility Engineering. His main research interests are virtual safety assessment and its components, including driver behavior modeling, scenario generation, and statistical methods.
\end{IEEEbiography}


\vfill

\end{document}


\onecolumn
\begingroup
\fontsize{20pt}{20pt}\selectfont
IEEE Copyright Notice\\
\endgroup
\\
© 2023 IEEE. Personal use of this material is permitted. Permission from IEEE must be obtained for all other uses, in any current or future media, including reprinting/republishing this material for advertising or promotional purposes, creating new collective works, for resale or redistribution to servers or lists, or reuse of any copyrighted component of this work in other works.
\twocolumn

\renewcommand\thefigure{A\arabic{figure}}
\renewcommand\thetable{A\Romannum{\arabic{table}}}
\title{Supplement of "Modeling Lead-vehicle Kinematics\\ For Rear-end Crash Scenario Generation"}

\author{Jian Wu, Carol Flannagan, Ulrich Sander, and Jonas Bärgman}

\markboth{IEEE Transactions on Intelligent Transportation Systems}%
{Shell \MakeLowercase{\textit{et al.}}: A Sample Article Using IEEEtran.cls for IEEE Journals}

\maketitle

\section{Datasets}
The SQL code for selecting rear-end pre-crashes (the subject vehicle is the one struck) in the CISS dataset is shown in Algorithm 1. The driver’s age information is collected as well.

The query code for selecting rear-end pre-crashes/near-crashes in the SHRP2 dataset is shown in Algorithm 2. In all cases, the subject vehicle is the one struck. One data log can contain two events, Events 1 and 2 (Event 1 happens first). The cases, in which Event 1 is a rear-end crash/near-crash or Event 2 is a rear-end crash/near-crash while Event 1 is not a crash, are selected.

\section{Methodology}
\subsection{Parameterization of the Lead-vehicle Speed Profiles}
\subsubsection{Sample weights}
\begin{figure*}[!t] 
    \centering
    \includegraphics[width=0.5\textwidth]{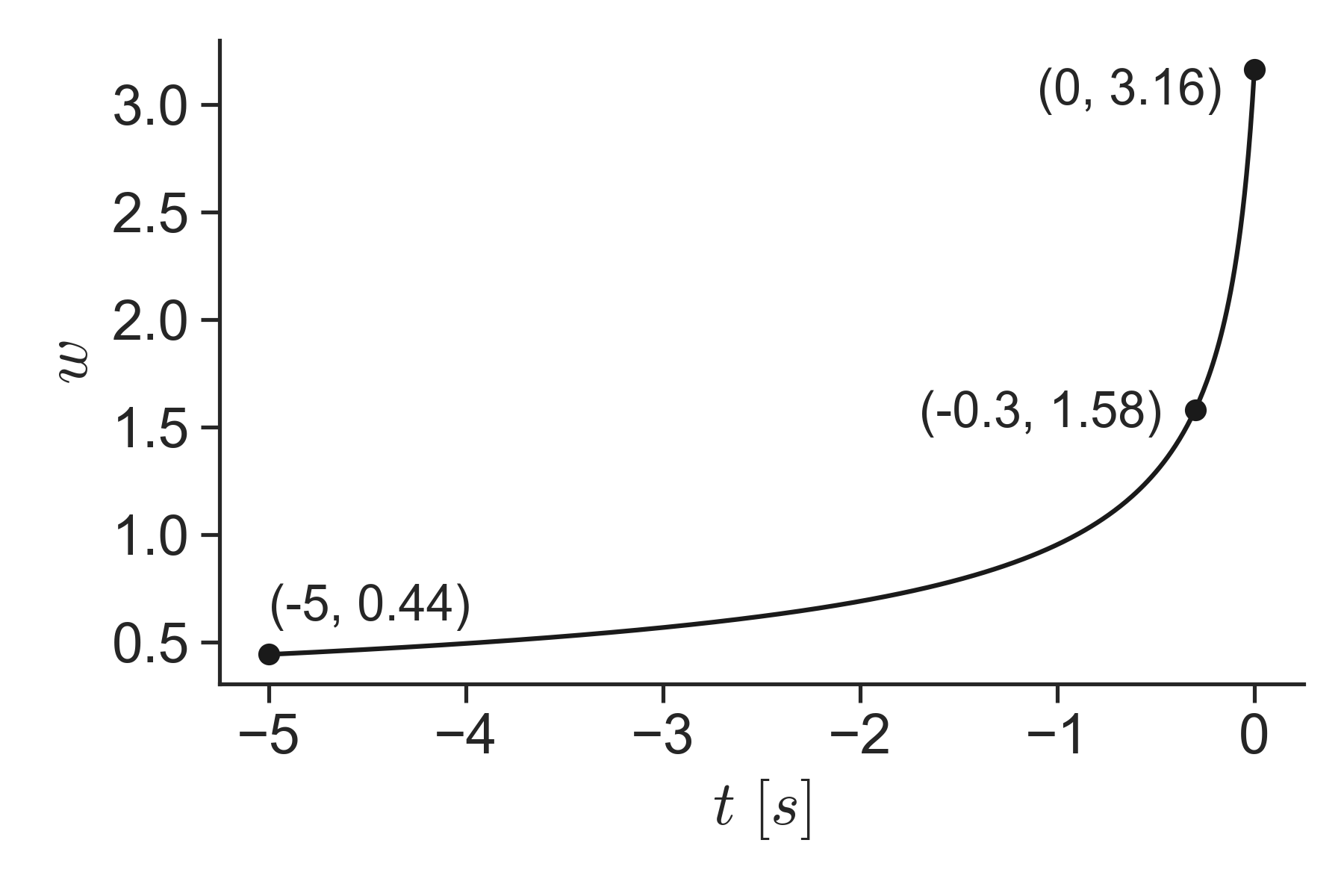}
    \caption{Sample weight against time.}
    \label{fig:smpwgt}
\end{figure*}

In each lead-vehicle speed profile, the weight of sample $i$ is defined as a function of $t(i)$,
\begin{equation}
        w_{i} = (0.1 - t_{i})^{-0.5}.
\end{equation}
The function is plotted in Fig. \ref{fig:smpwgt}; the weight ranges from 0.44 to 3.16 for near-crashes and from 0.44 to 1.58 for crashes. (The weight differences aren't huge.)

We wanted to fit the lead-vehicle speed profile into a piecewise linear model with as few breakpoints as possible, so we limited the maximum amount of breakpoints. Without sample weights, the piecewise linear model treats the speed changes close to time zero and those close to the start of the incident equally. In other words, the chance of the piecewise linear model adding a breakpoint is the same throughout the time period. However, with the sample weights considered, the model will prioritize capturing the speed changes closer to time zero by adding a new breakpoint.

\subsubsection{Loss function}
In the piecewise linear model, a larger number of breakpoints covers more details (segments) of the speed changes with more parameters, leading to a more accurate fitting. However, as dimensionality (the number of parameters) increases, the complexity of the multivariate distribution modeling increases and the amount of data needed often grows exponentially \cite{koppen2000curse}. Therefore, we introduced the loss function
\begin{equation}
    L = (\epsilon + \lambda \cdot \frac{\text{max}(v)}{\Delta v + \epsilon}) \cdot n_b - R^2.
\end{equation}
to balance accuracy and complexity.

The loss function selects the weighted piecewise linear regression with the smallest loss among several with various numbers of breakpoints. The function contains two parts: 1) a penalty for the number of breakpoints and 2) a reward for fitting accuracy.

The penalty part, $(\epsilon + \lambda \cdot \frac{\text{max}(v)}{\Delta v + \epsilon}) \cdot n_b$, is based on the perceptibility of the lead vehicle’s speed change to the following vehicle’s driver. An easier perceptible speed change corresponds with a smaller penalty given the same number of breakpoints. According to Lee \cite{lee1976theory}, the changing rate of the optical size (width) of the lead vehicle on the following vehicle's driver’s retina is computed as
\begin{equation}
    \theta^\prime = \frac{d}{dt}(\text{tan}^{-1}\frac{W}{d})= \frac{W}{W^2 + d^2} \cdot (v_f - v_l),
\end{equation}
where $\theta$ is the retinal image size, $W$ is the width of the lead vehicle, $d$ is the following distance, $v_f$ is the following vehicle's speed, and $v_l$ is the lead vehicle's speed. The lead vehicle's speed change will affect $\theta^\prime$ and, therefore, be perceived by the following vehicle's driver. Considering that 1) a larger lead vehicle's speed change ($\Delta v$) leads to a larger change of $\theta^\prime$, 2) a higher lead vehicle's speed ($\text{max}(v)$) is usually associated with a larger following distance leading to a smaller change of $\theta^\prime$, and 3) a lead speed change associated with a larger change of $\theta^\prime$ is more perceptible, the penalty should increase with the decrease of $\Delta v$ or the increase of $\text{max}(v)$ given the same number of breakpoints $n_b$. Regarding the third point, it is worth mentioning that Tian et al. \cite{tian2022explaining} discovered an interesting aspect regarding pedestrian crossing. As the approaching vehicle gets closer, the pedestrian's change rate of the vehicle's optical size initially increases, but after reaching approximately one meter, it decreases. This is because the pedestrian and the vehicle are not in the same lane. In contrast, our study focuses on two vehicles in the same lane, and we observed a monotonic increase in $\theta^\prime$, thus confirming the third point. In addition, instead of $\text{max}(v^2)$, $\text{max}(v)$ is used in the penalty part because the following distance does not increase linearly with speed \cite{loulizi2019steady}.

The reward part, $-R^2$, rewards fitting accuracy. Given the same penalty, a larger R-squared indicates better fitting accuracy and, thus, a smaller loss.

\subsubsection{Selection of pre-configured parameters}
\begin{figure*}[!t] 
    \centering
    \includegraphics[width=0.5\textwidth]{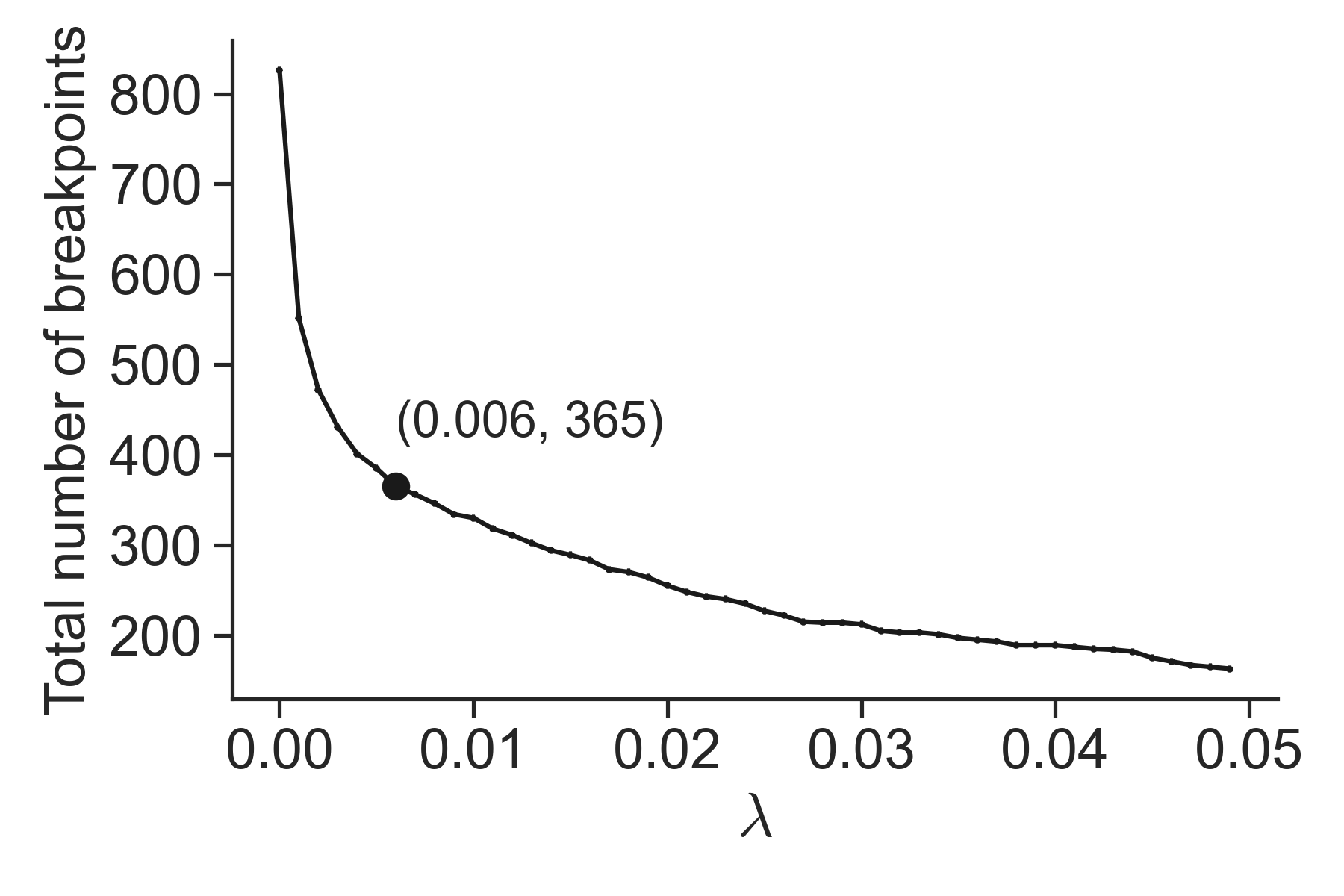}
    \caption{Total number of breakpoints against $\lambda$. $\lambda$ was set as the elbow of the curve, 0.006.}
    \label{fig:seloflam}
\end{figure*}

The two pre-configured parameters in the piecewise linear model are $\lambda$ and $n_{b,max}$. To determine $n_{b,max}$, we randomly selected a small sub-dataset (30 events) across the CISS and SHRP2 datasets and manually annotated the preferred number of breakpoints for each event in the sub-dataset. It was found that most of the events can be covered with no more than two breakpoints, and very few require three breakpoints. Therefore, $n_{b,max}$ was set as 3. While $\lambda$ was set as the elbow of the curve shown in Fig. \ref{fig:seloflam} (the total number of breakpoints for all events against $\lambda$).

\subsection{Comparison between CISS$_{sc}$ and SHRP2$_{sc}$}
\begin{figure*}[!t]
\centering
\subfloat[]{\includegraphics[width=0.3\textwidth]{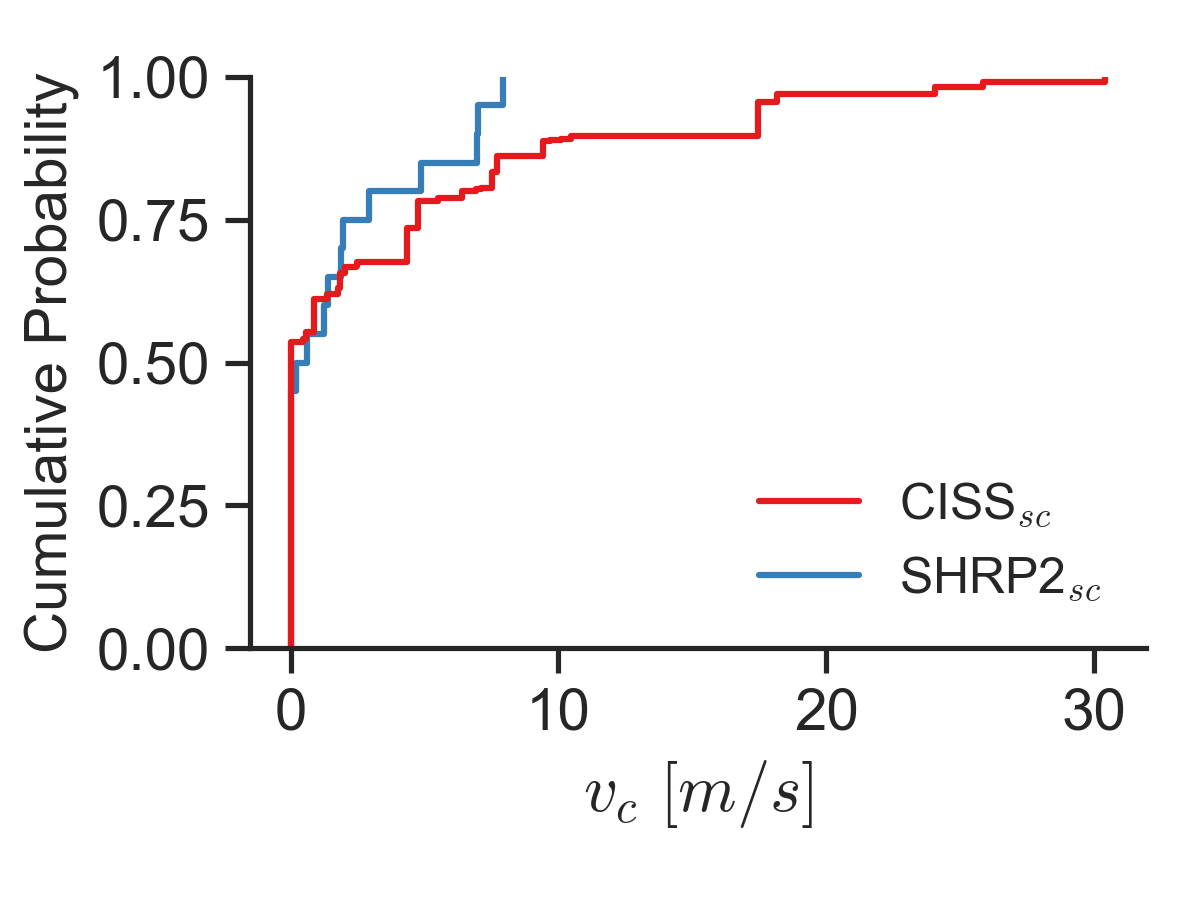}}
\hfil
\subfloat[]{\includegraphics[width=0.3\textwidth]{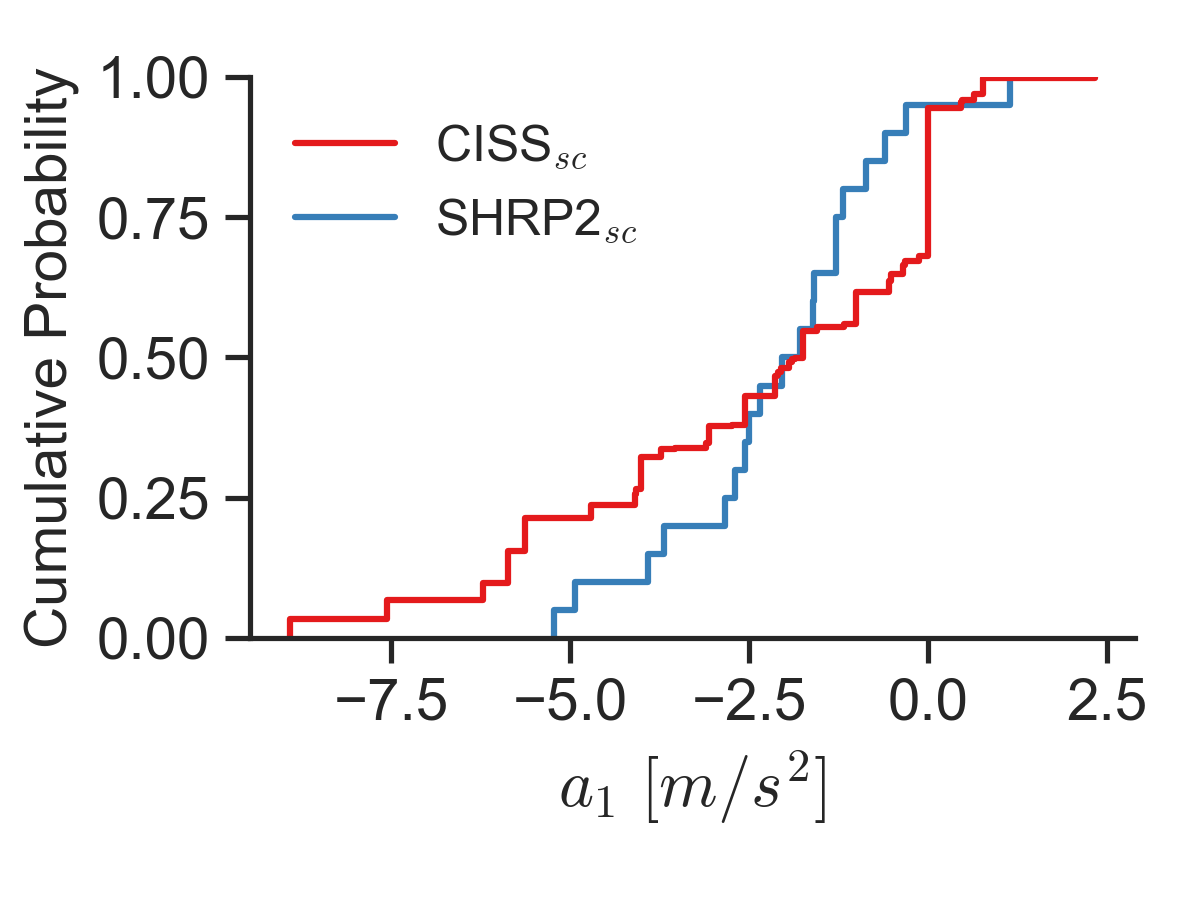}}
\hfil
\subfloat[]{\includegraphics[width=0.3\textwidth]{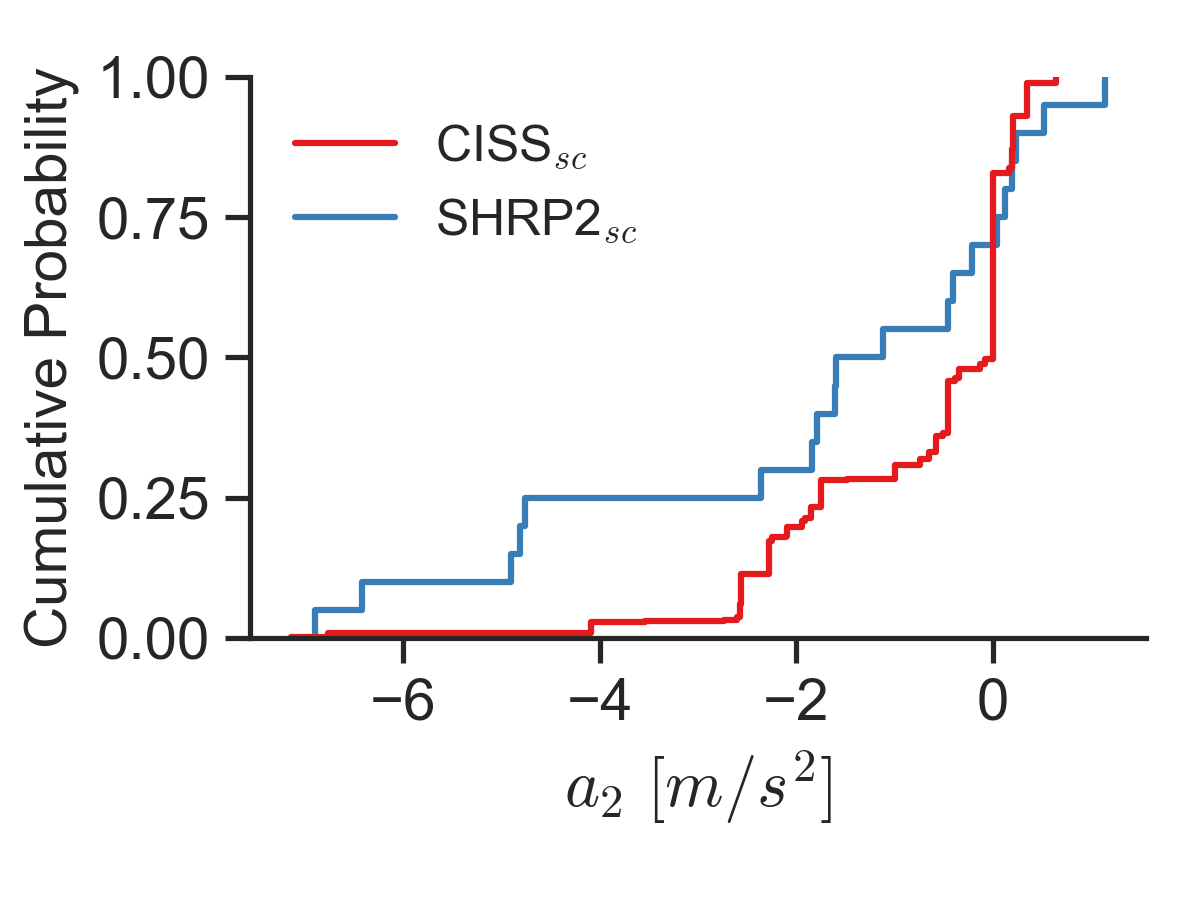}}
\vfil
\subfloat[]{\includegraphics[width=0.3\textwidth]{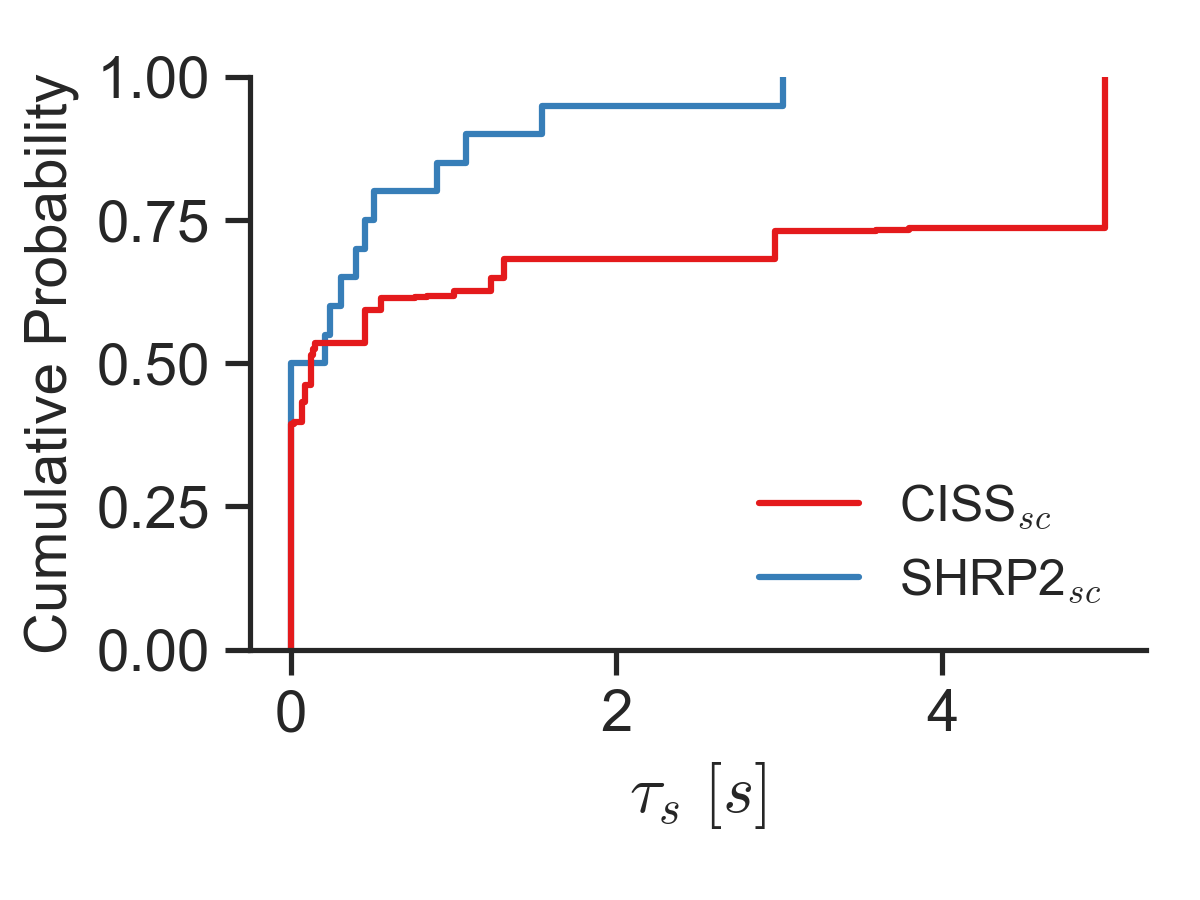}}
\hfil
\subfloat[]{\includegraphics[width=0.3\textwidth]{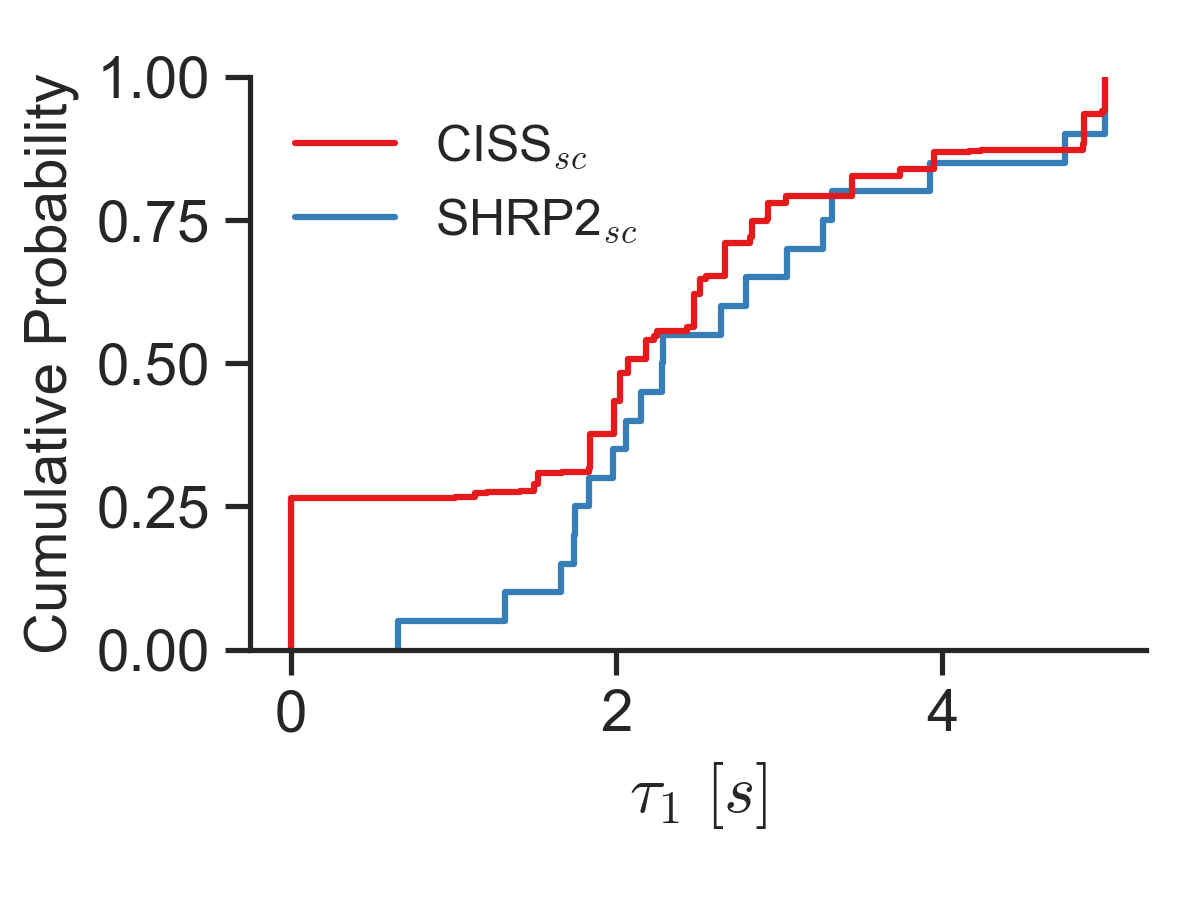}}
\hfil
\subfloat[]{\includegraphics[width=0.3\textwidth]{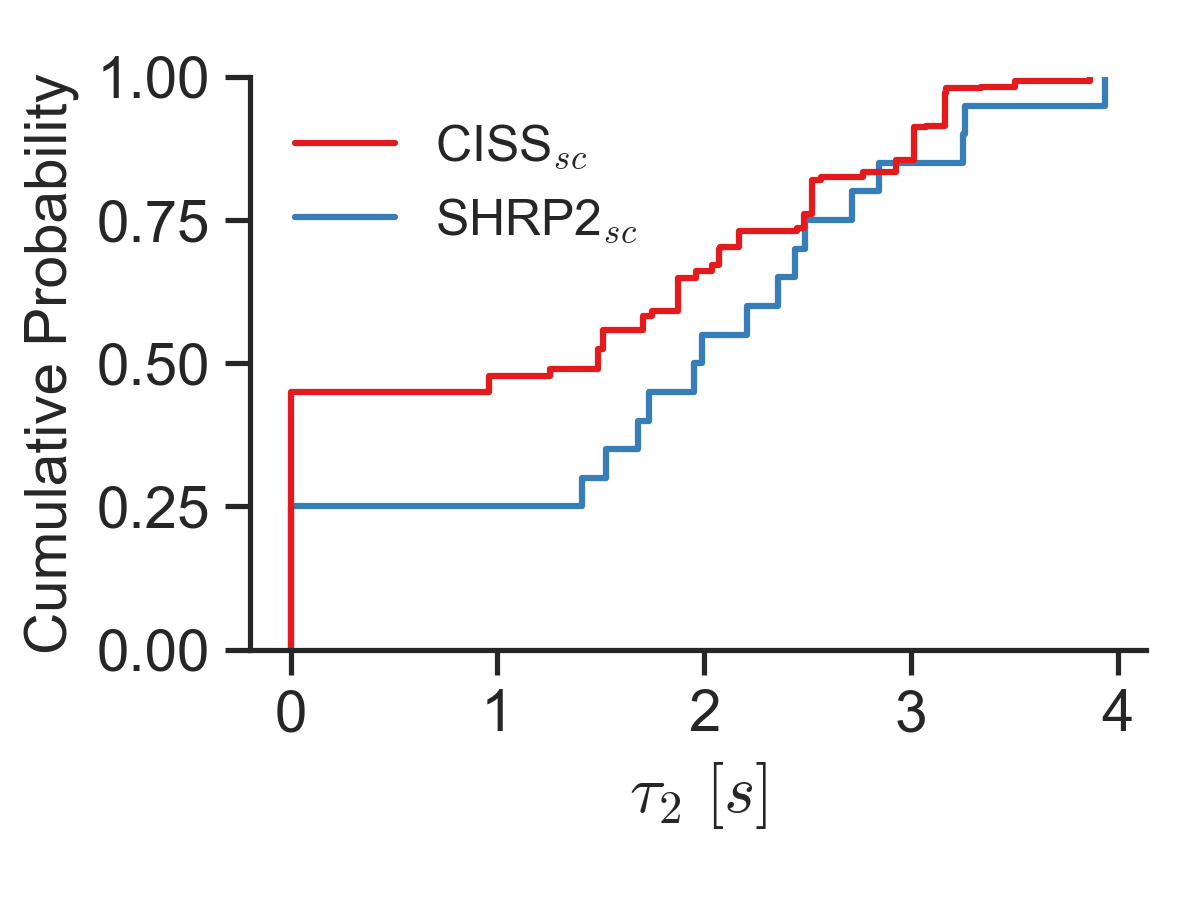}}
\caption{Comparison between the weighted CDFs of the six parameters of CISS$_{sc}$ and SHRP2$_{sc}$.}
\label{fig:scparamcmp}
\end{figure*}

The weighted two-sample Kolmogorov–Smirnov (KS) tests of each of the six parameters were conducted to determine whether the severe crashes in the CISS and SHRP2 datasets are from the same distribution. The results in Table \ref{tab:sccomparison} show no significant differences. The weighted cumulative distribution functions (CDFs) of the six parameters of CISS$_{sc}$ and SHRP2$_{sc}$ are shown in Fig. \ref{fig:scparamcmp}.

\subsection{Weight Trimming of Raw Sample Weights in CISS$_{sc}$}
The weight-trimming approach used in the study is proposed by Van de Kerckhove et al.. In weight trimming, weights exceeding a specified cut-point are trimmed to that value, as expressed in (\ref{eq:wt1}).

\begin{equation} \label{eq:wt1}
    w_{jt} = \begin{cases}w_0,&{\text{if}}\ w_j > w_0; \\ 
               {w_j,}&{\text{otherwise.}}
    \end{cases},
\end{equation}
where $w_j$ is the weight prior to trimming, and $w_0$ is the trimming cut-point which is defined as

\begin{equation}
    w_0 = 3.5\sqrt{1 + CV^2(w_j)} \cdot \text{median}(w_j),
\end{equation}
where $CV^2(w_j)$ is the coefficient of variation of $w_j$. In the case of CISS$_{sc}$, the trimming cut-point is 1797.3.

\subsection{Adding Selected Near-crashes as Variations of Crashes}
\begin{figure*}[!t] 
    \centering
    \includegraphics[width=0.5\textwidth]{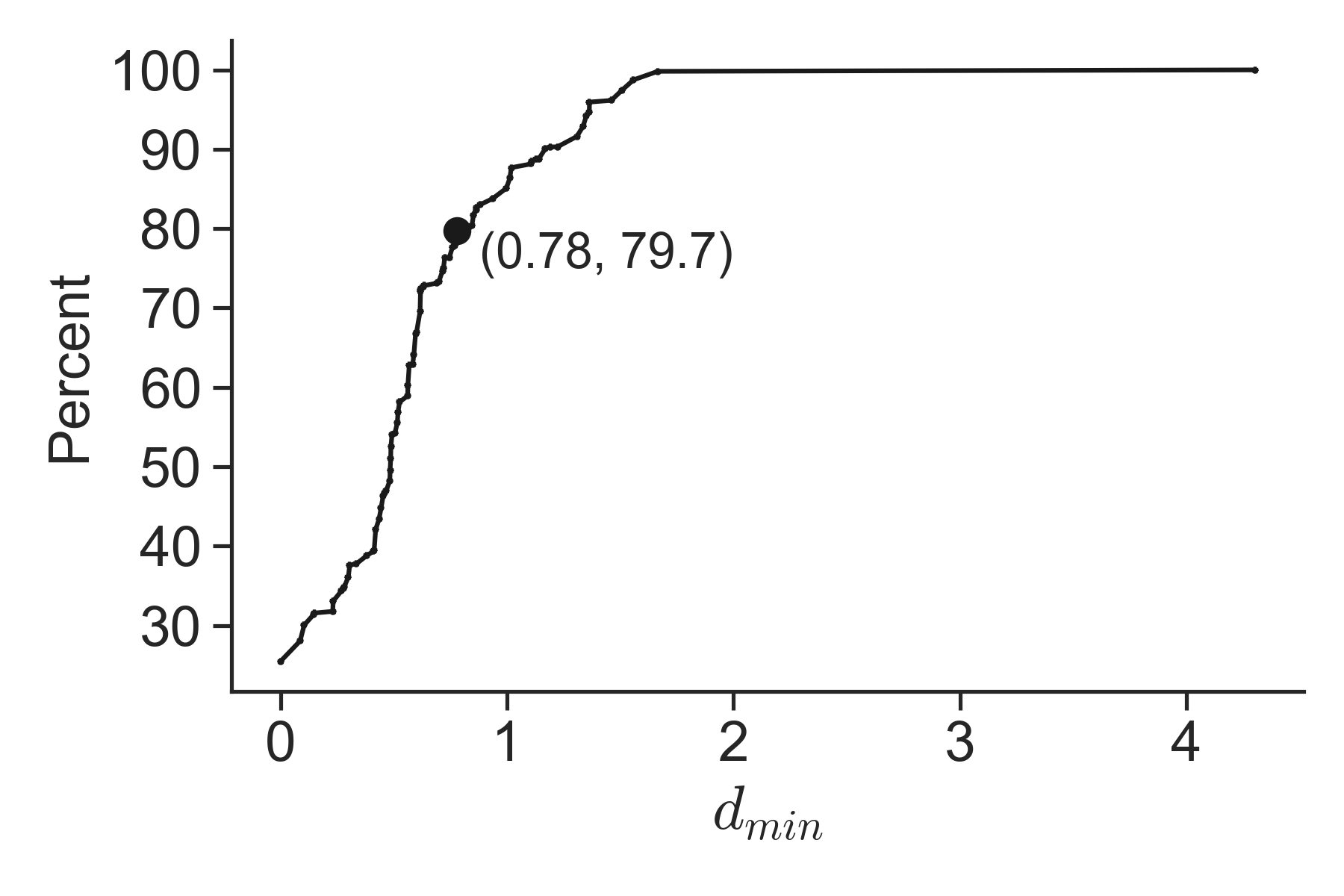}
    \caption{CDF of the minimum Euclidean distance of the combined crashes. The threshold $d_{thd}$ is set as the elbow of the curve, 0.780.}
    \label{fig:cdfofdistance}
\end{figure*}

\begin{figure*}[!t]
    \centering
    \includegraphics[width=0.5\textwidth]{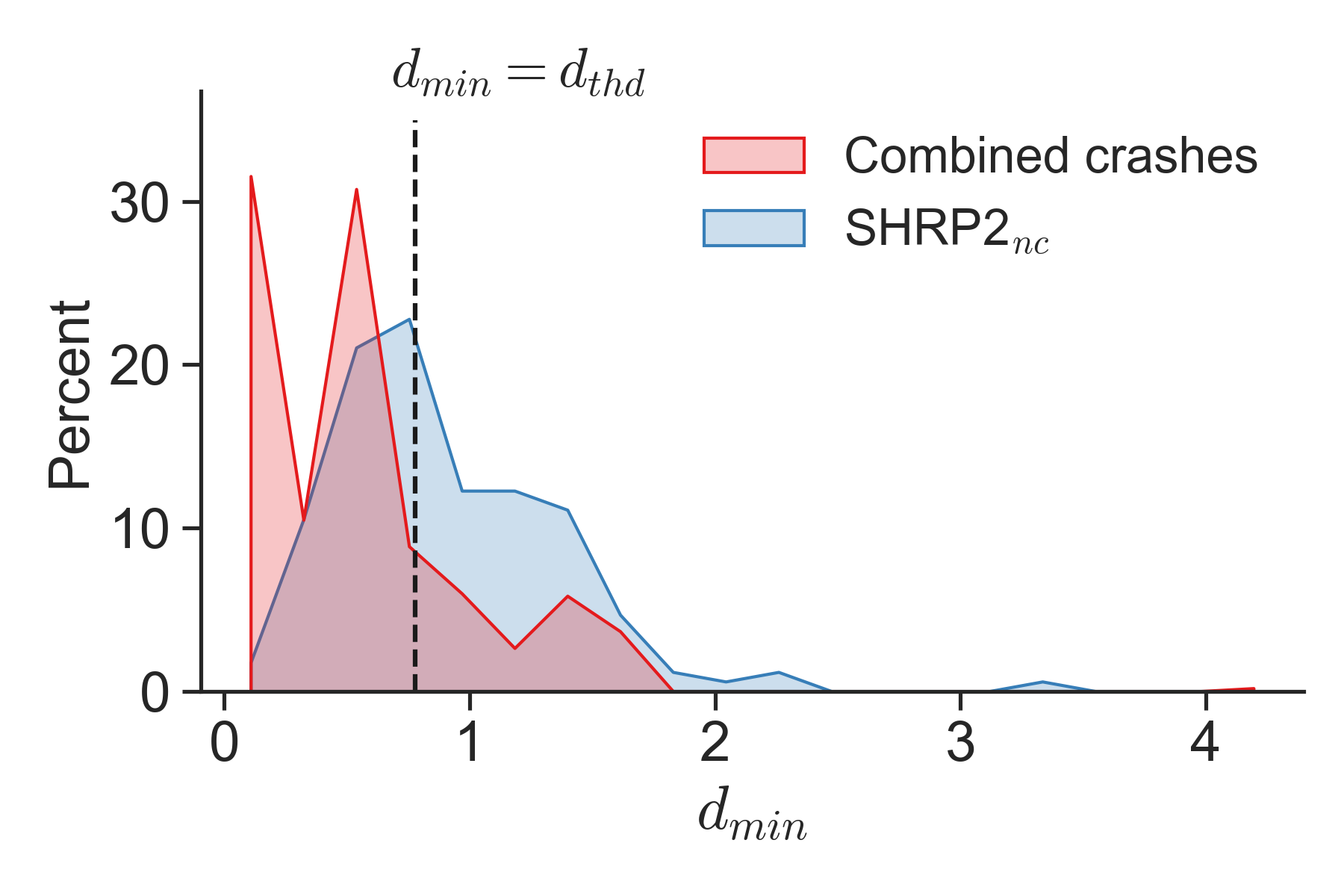}
    \caption{Distributions of the minimum Euclidean distance of the combined crashes and SHRP2$_{nc}$.}
    \label{fig:disofdistance}
\end{figure*}

The CDF curve of the minimum Euclidean distance of the combined crashes is shown in Fig. \ref{fig:cdfofdistance}. The pre-configured parameter $d_{thd}$ was set as the elbow of the curve, where the curve slope slows down, and it can be seen as the start of the long tail. Moreover, the distributions of the minimum Euclidean distance of the combined crashes and SHRP2$_{nc}$ are shown in Fig. \ref{fig:disofdistance}.

\section{Results}
\subsection{Data Combination}
\begin{figure*}[!t]
    \centering
    \subfloat[]{\includegraphics[width=0.3\textwidth]{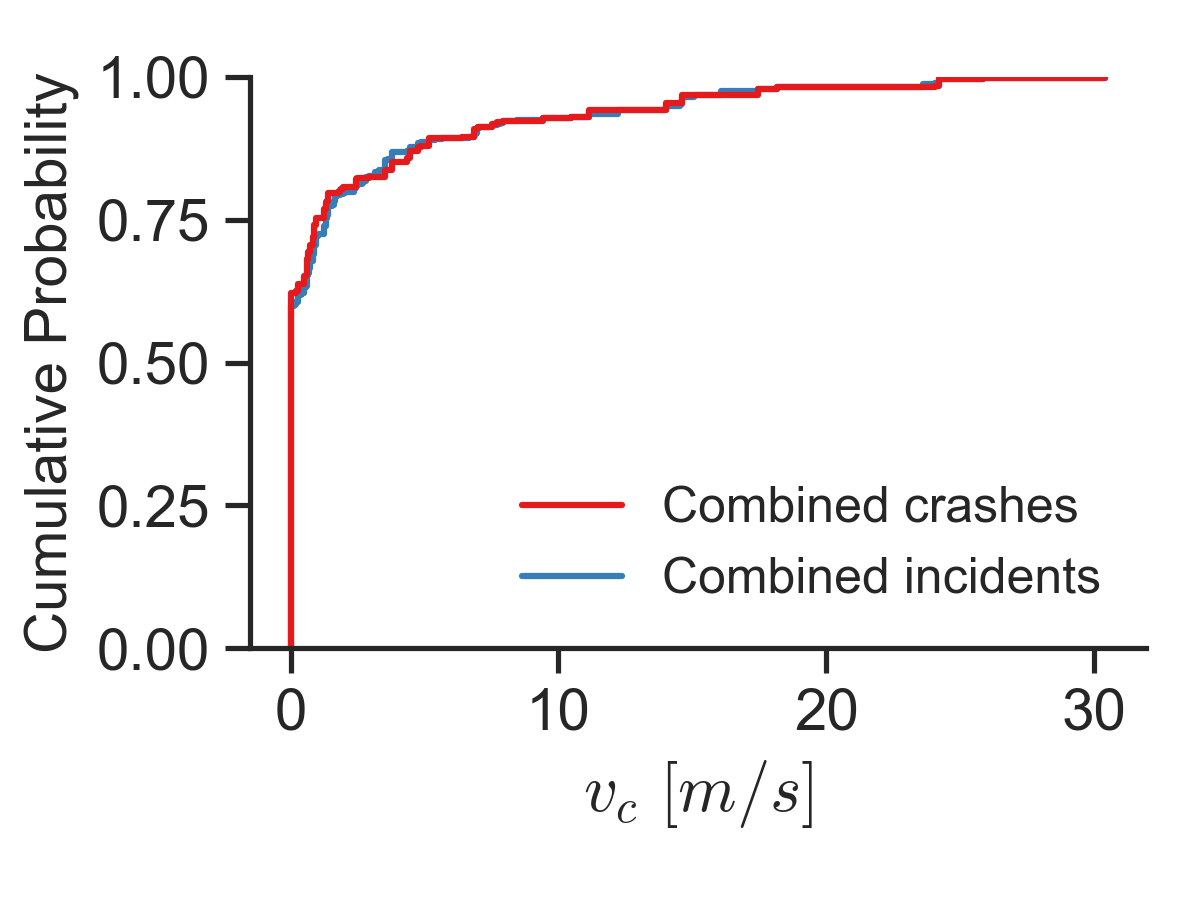}}
    \hfil
    \subfloat[]{\includegraphics[width=0.3\textwidth]{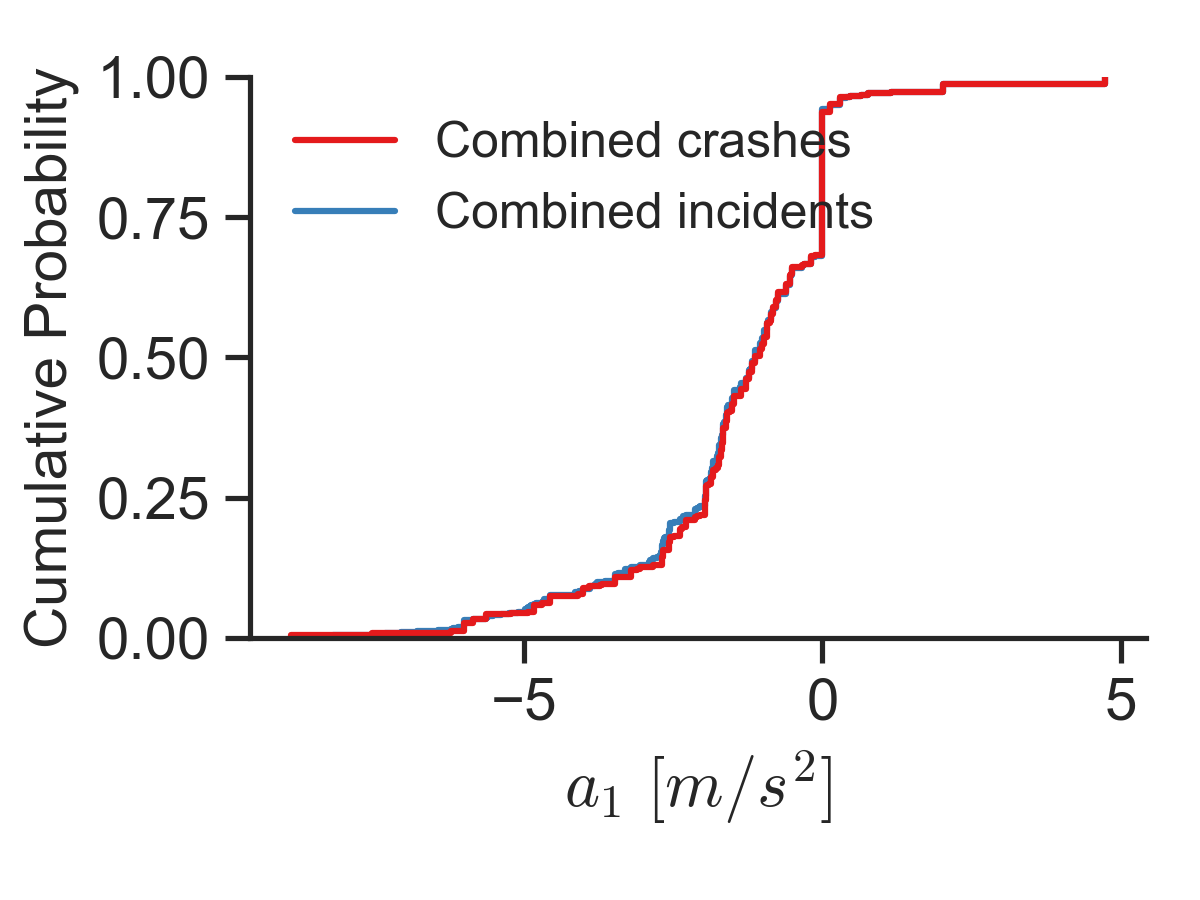}}
    \hfil
    \subfloat[]{\includegraphics[width=0.3\textwidth]{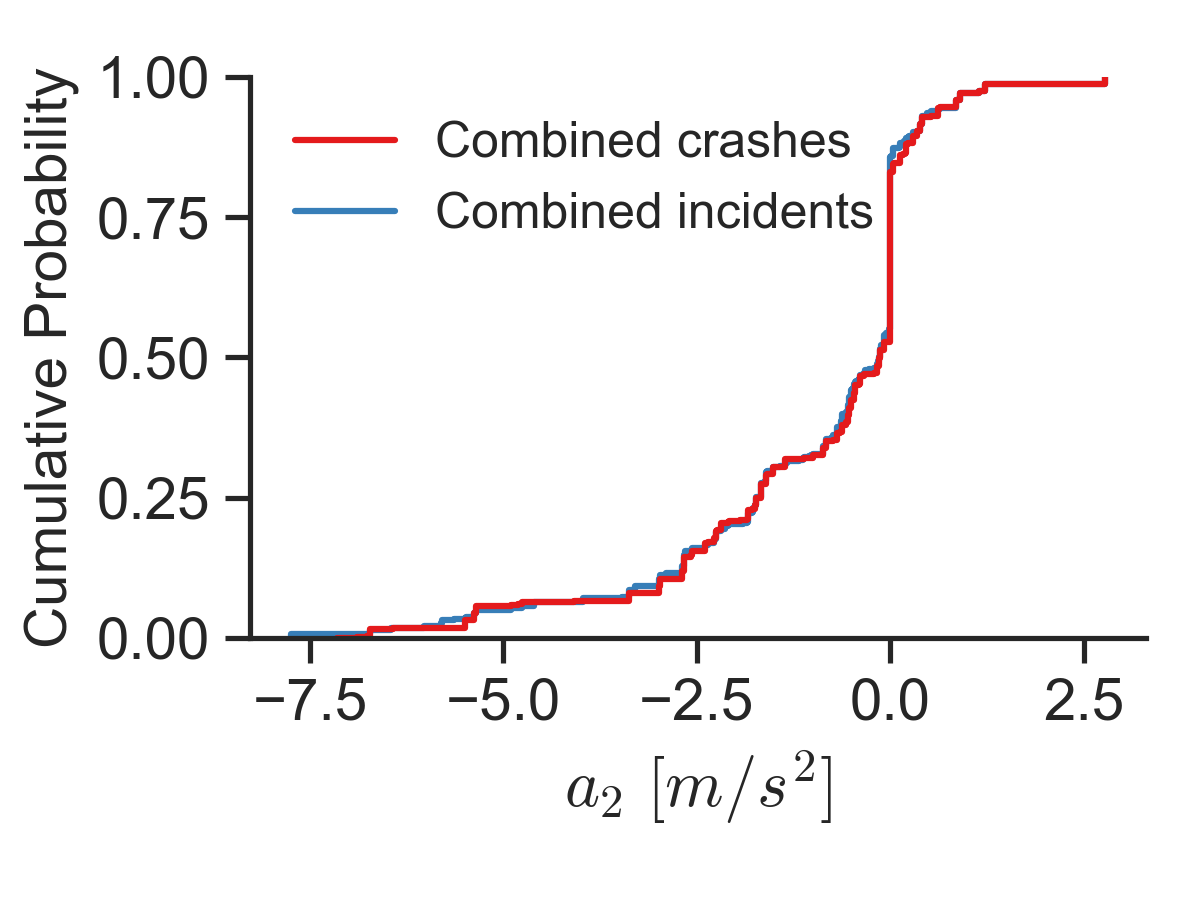}}
    \vfil
    \subfloat[]{\includegraphics[width=0.3\textwidth]{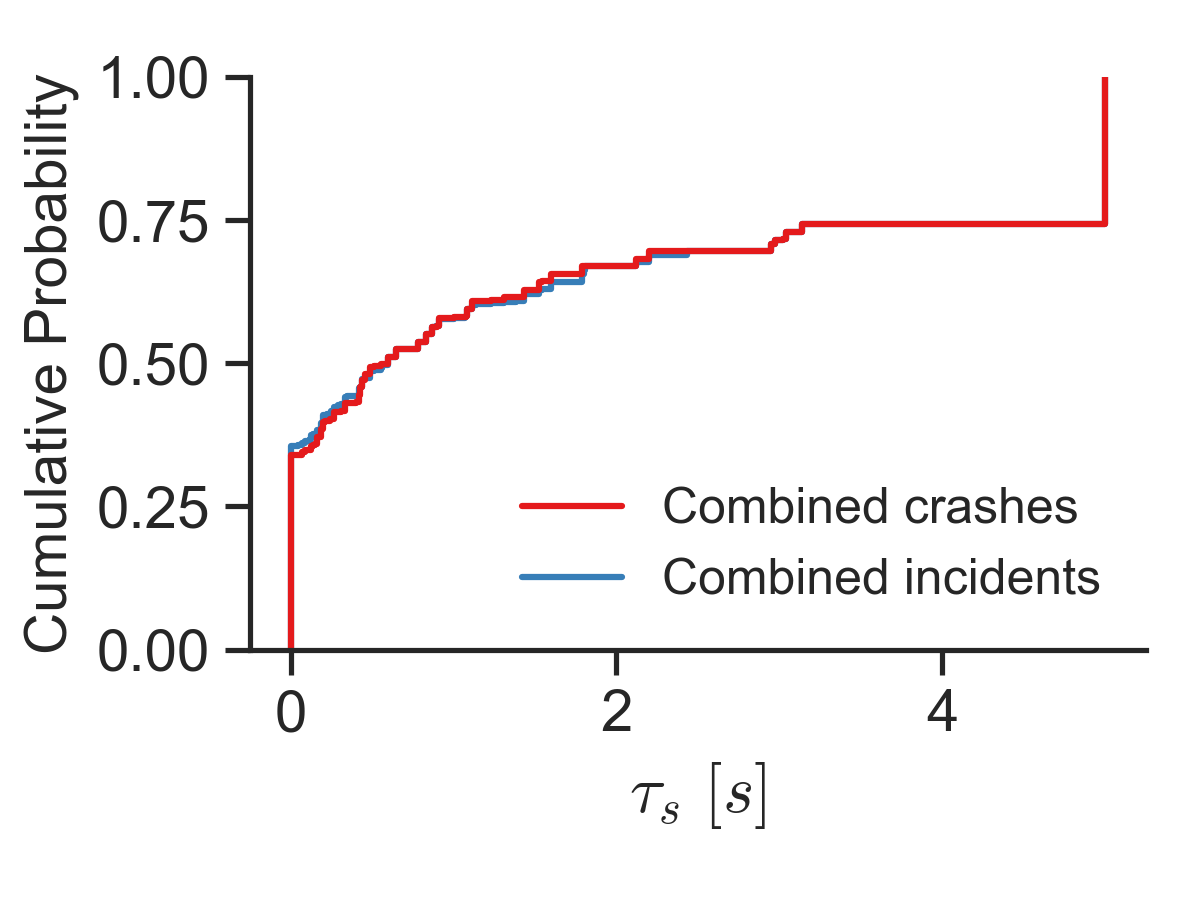}}
    \hfil
    \subfloat[]{\includegraphics[width=0.3\textwidth]{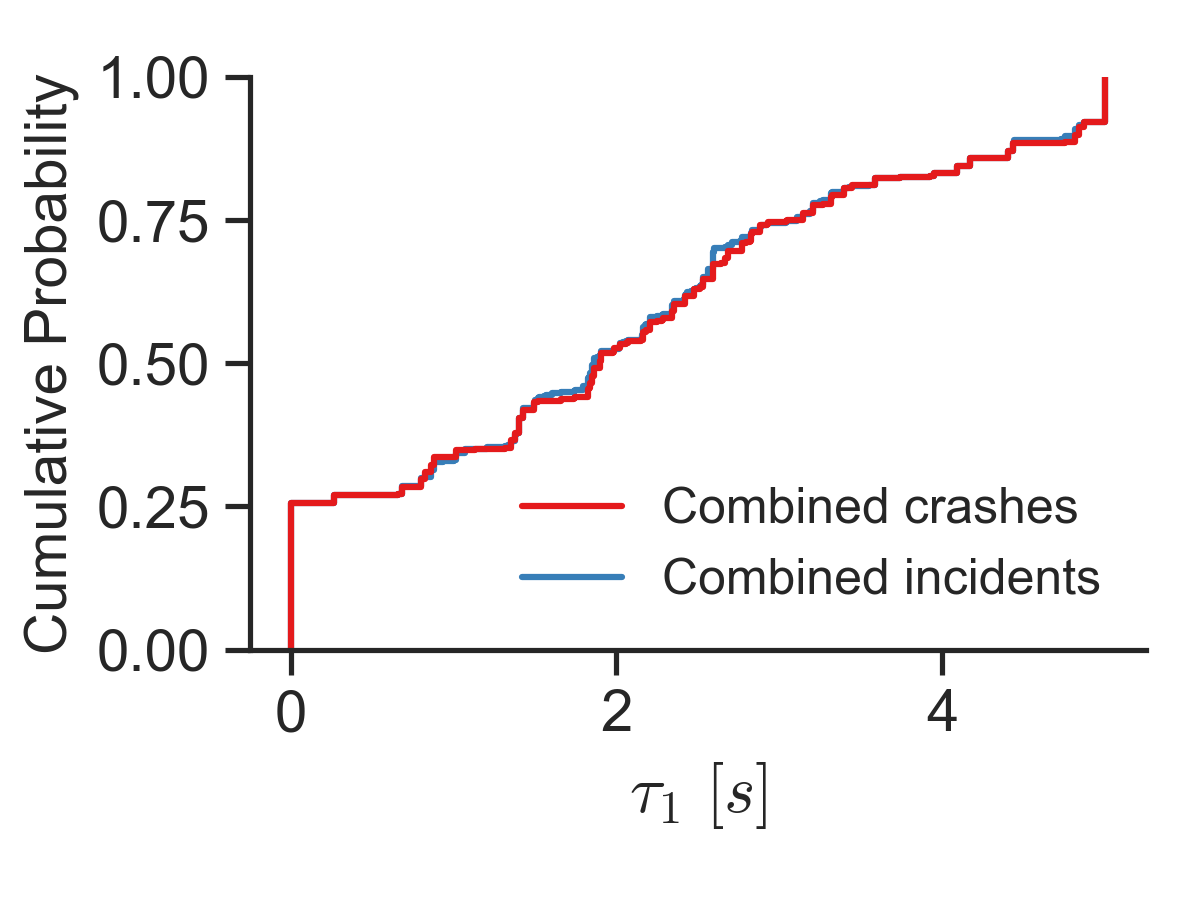}}
    \hfil
    \subfloat[]{\includegraphics[width=0.3\textwidth]{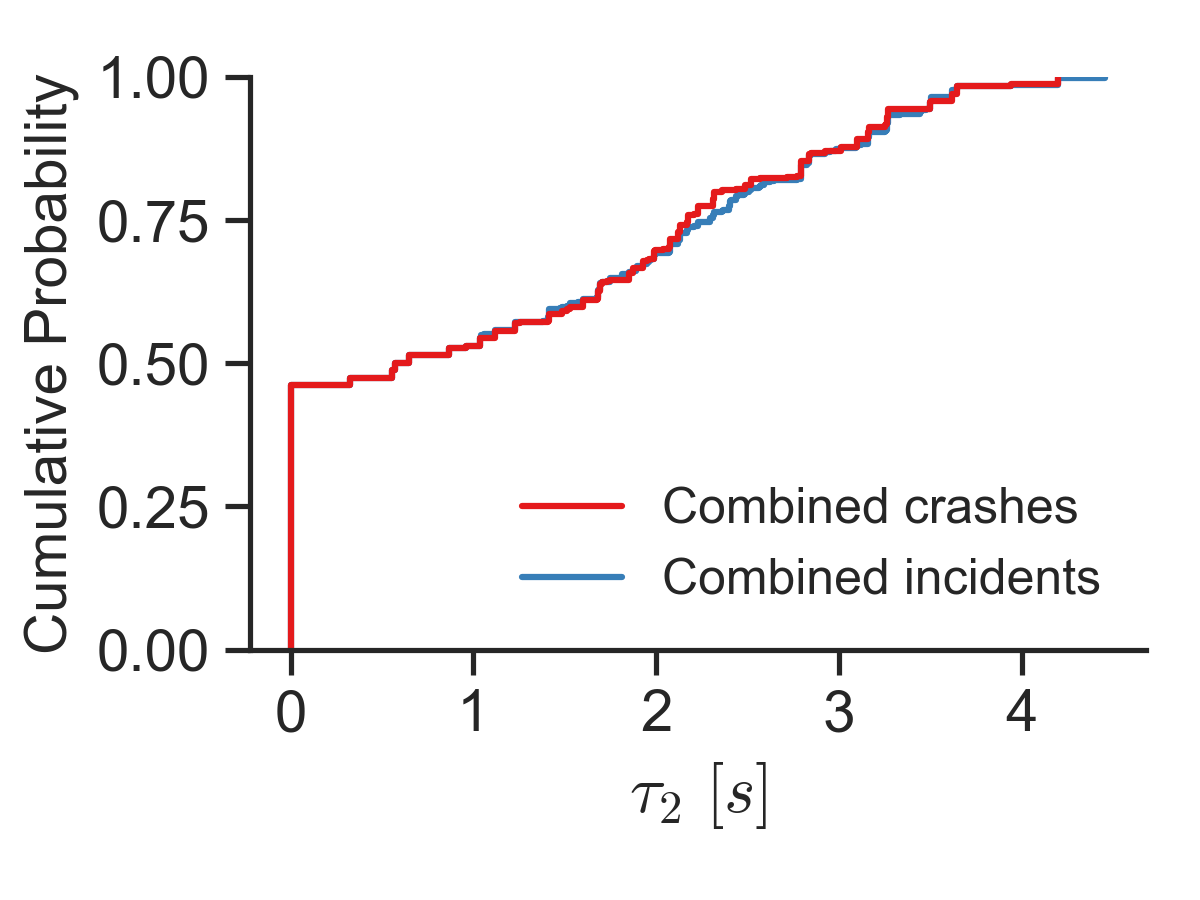}}
    \caption{Weighted CDFs of the six parameters of the combined crash and the combined incident datasets. The difference between the two datasets in terms of a single parameter's marginal distribution is negligible.}
    \label{fig:wgtCDFsofdatasets}
\end{figure*}

The weighted CDFs of the six parameters of the combined crash and the combined incident datasets are shown in Fig. \ref{fig:wgtCDFsofdatasets}.

\subsection{Comparison between the Synthetic and Raw incidents}
\begin{figure*}[!t]
    \centering
    \subfloat[]{\includegraphics[width=0.3\textwidth]{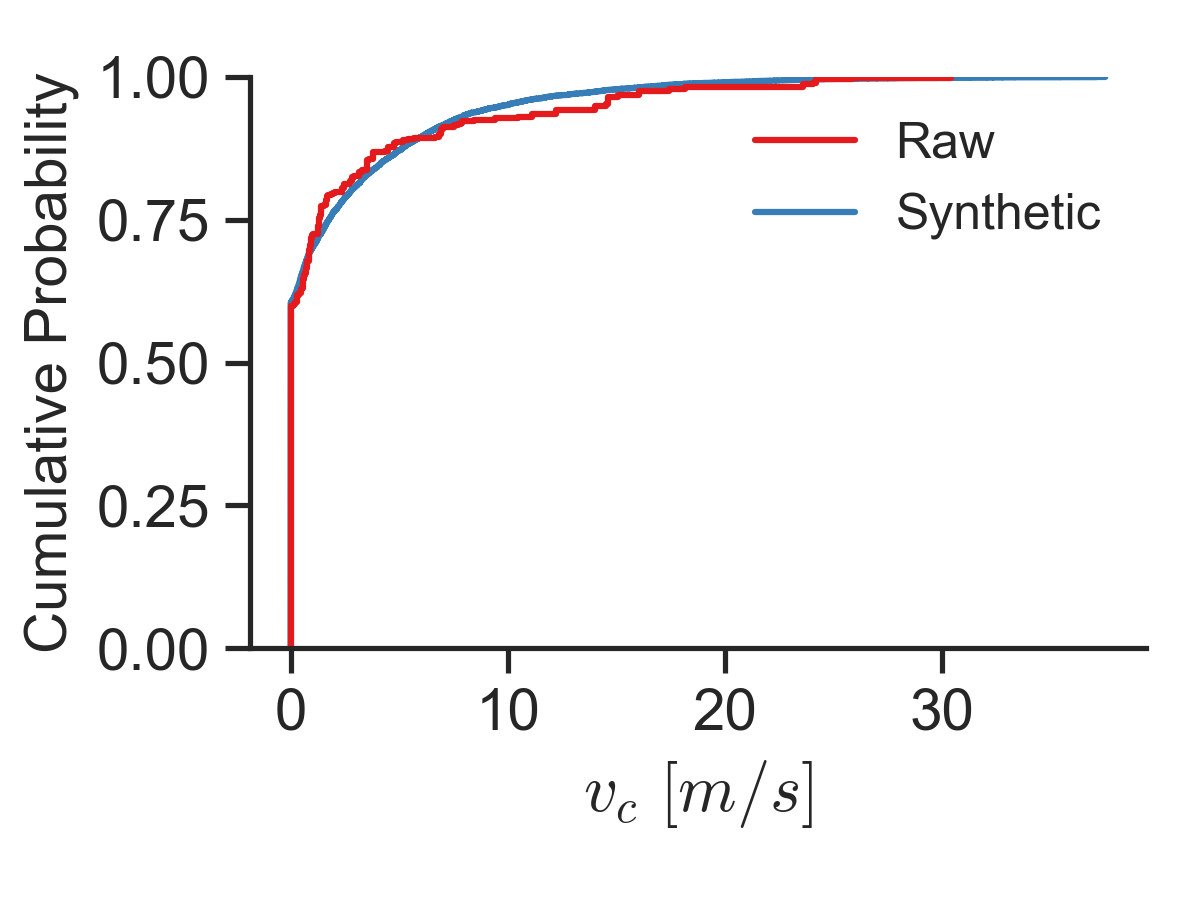}}
    \hfil
    \subfloat[]{\includegraphics[width=0.3\textwidth]{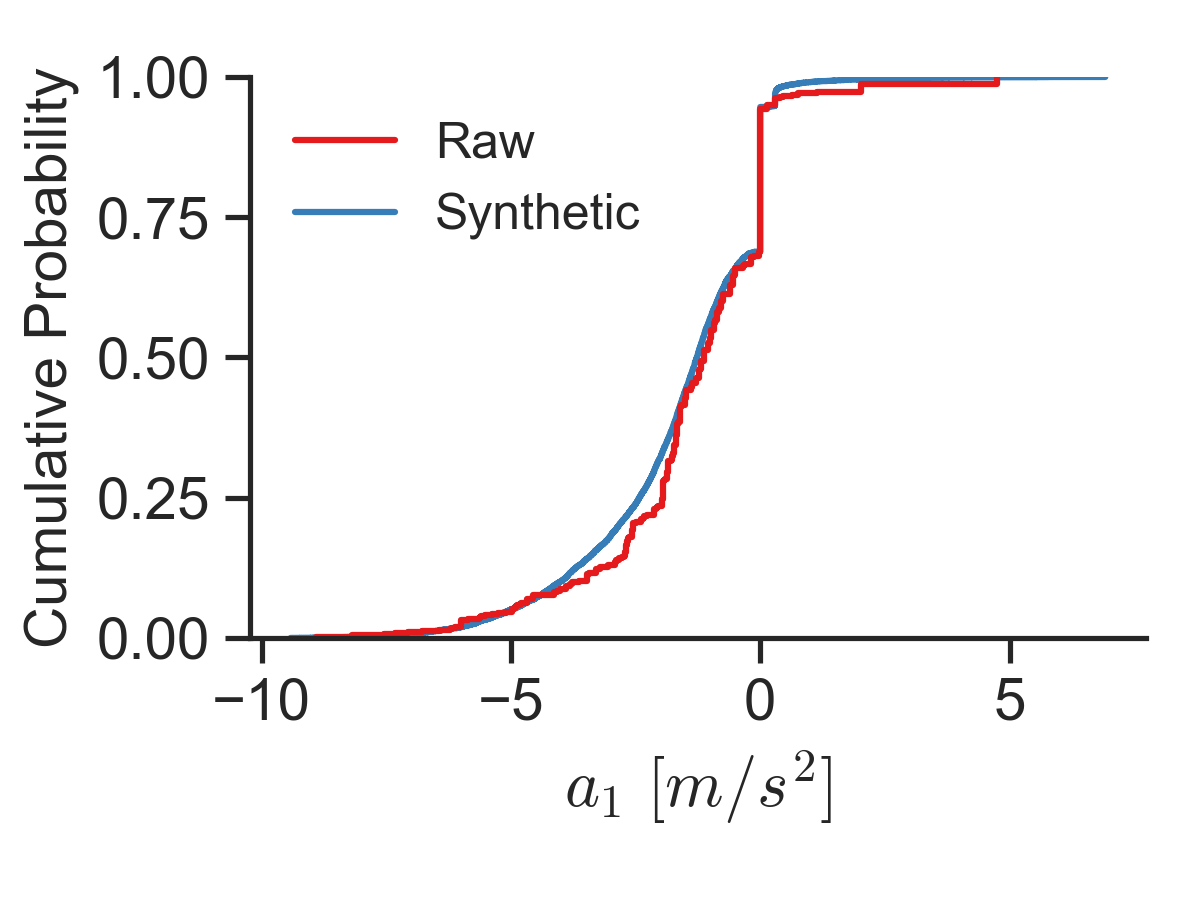}}
    \hfil
    \subfloat[]{\includegraphics[width=0.3\textwidth]{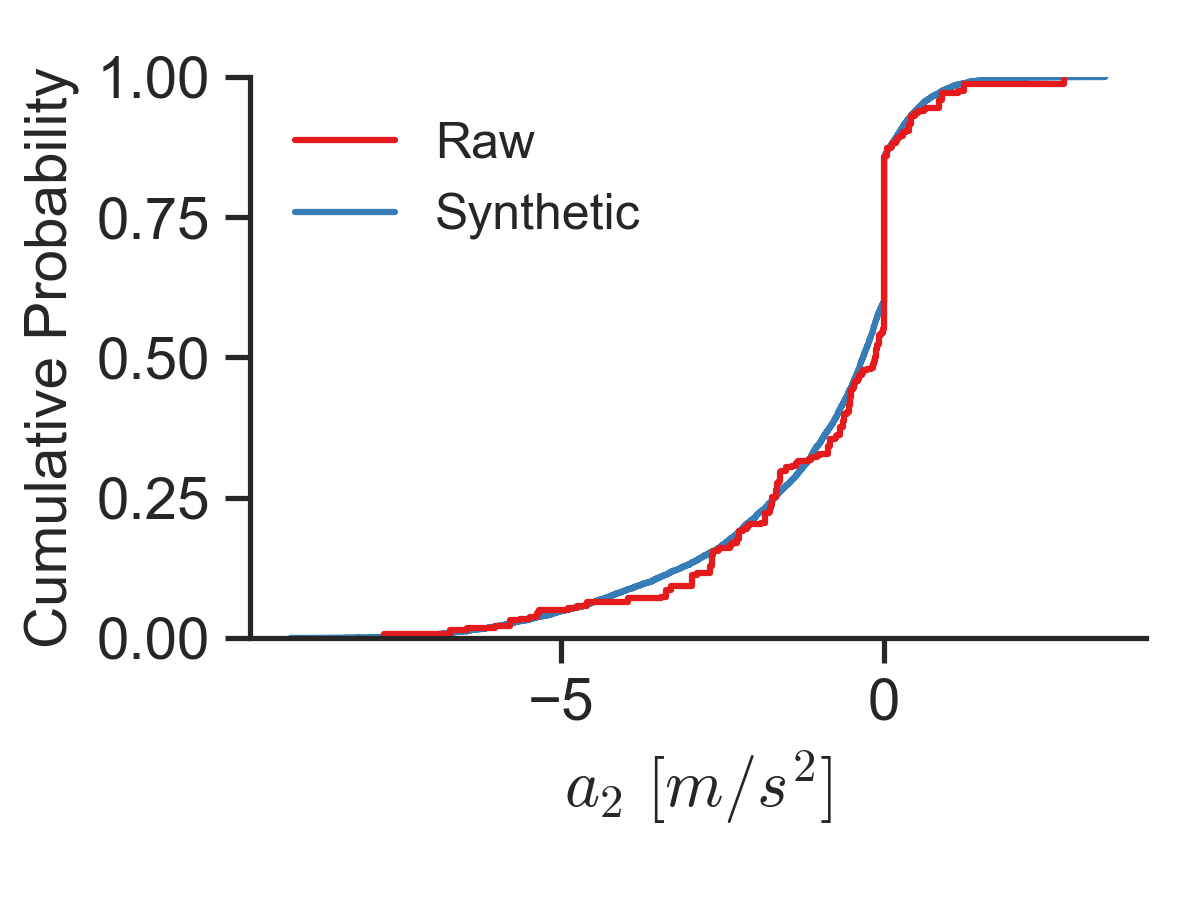}}
    \vfil
    \subfloat[]{\includegraphics[width=0.3\textwidth]{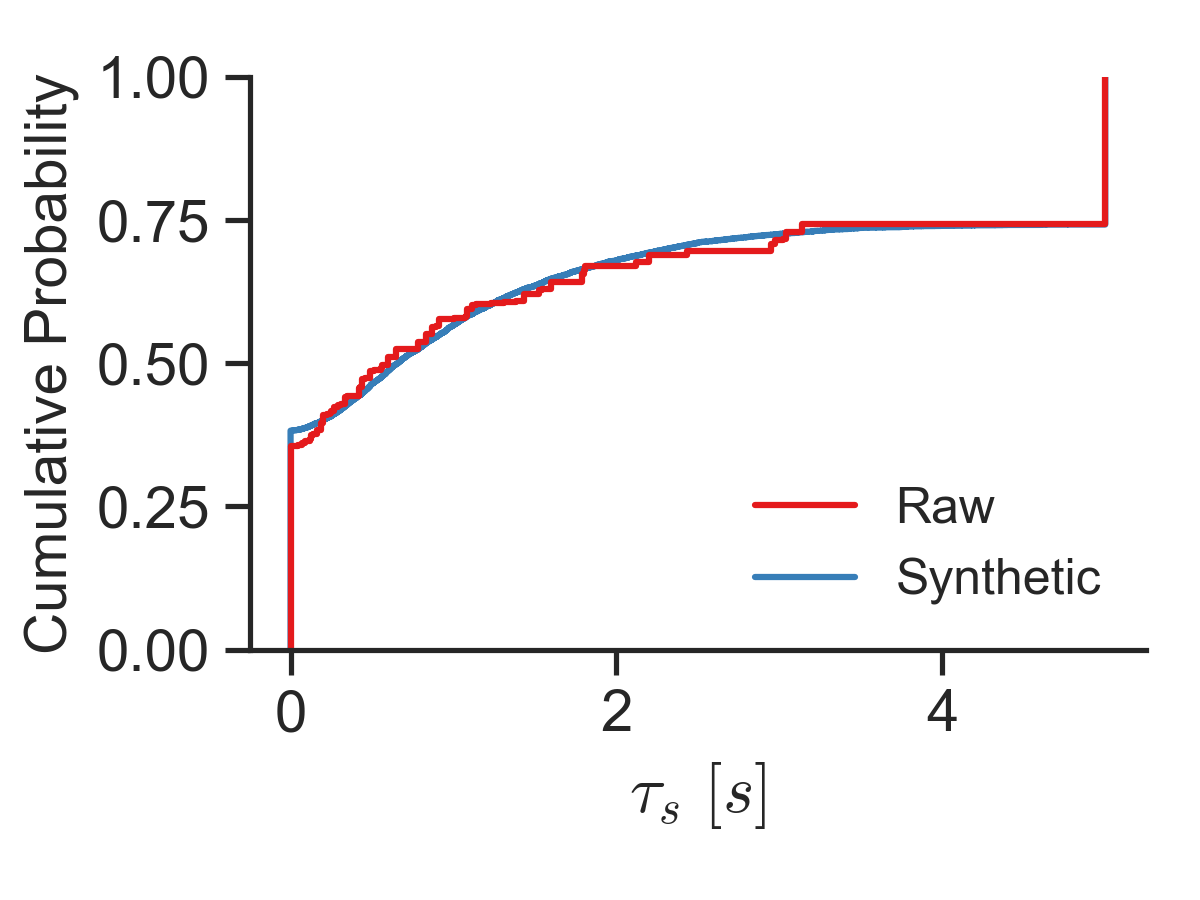}}
    \hfil
    \subfloat[]{\includegraphics[width=0.3\textwidth]{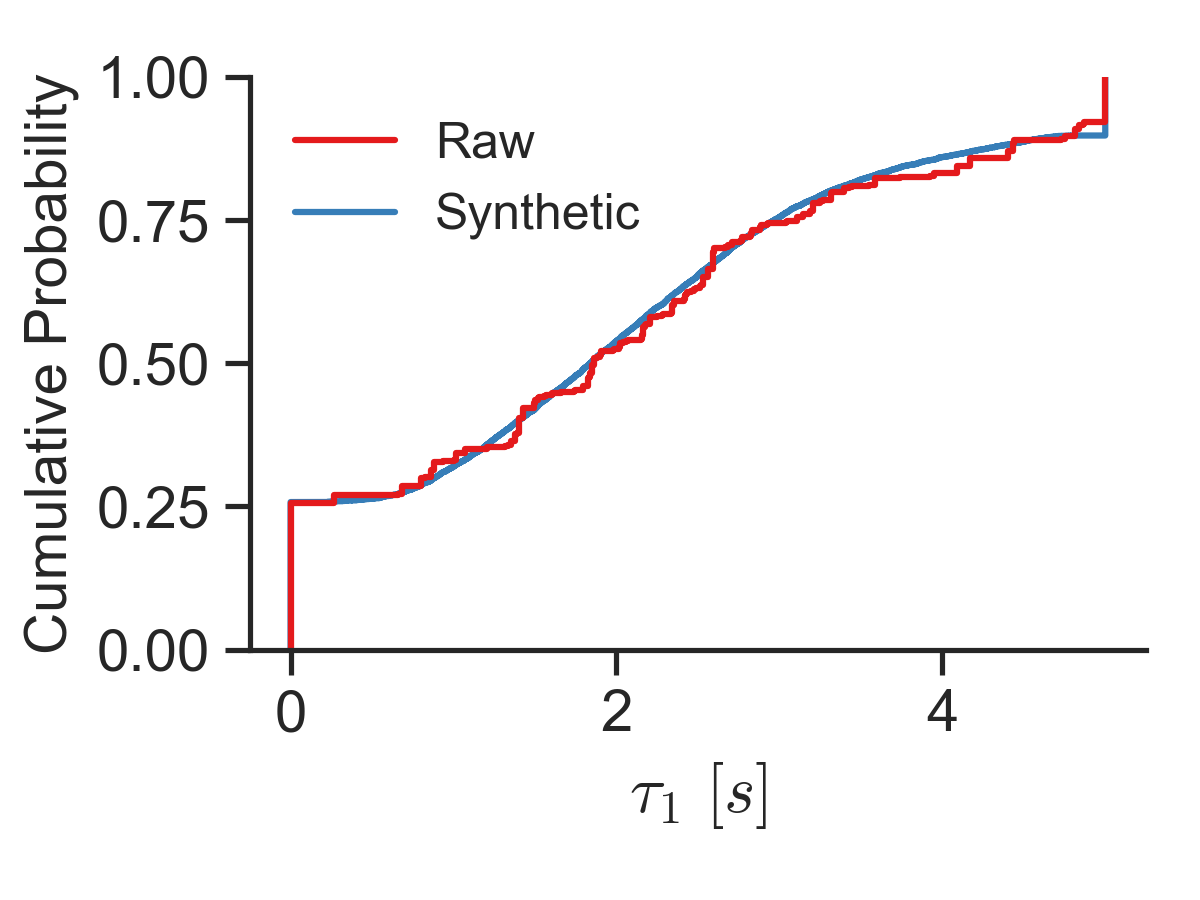}}
    \hfil
    \subfloat[]{\includegraphics[width=0.3\textwidth]{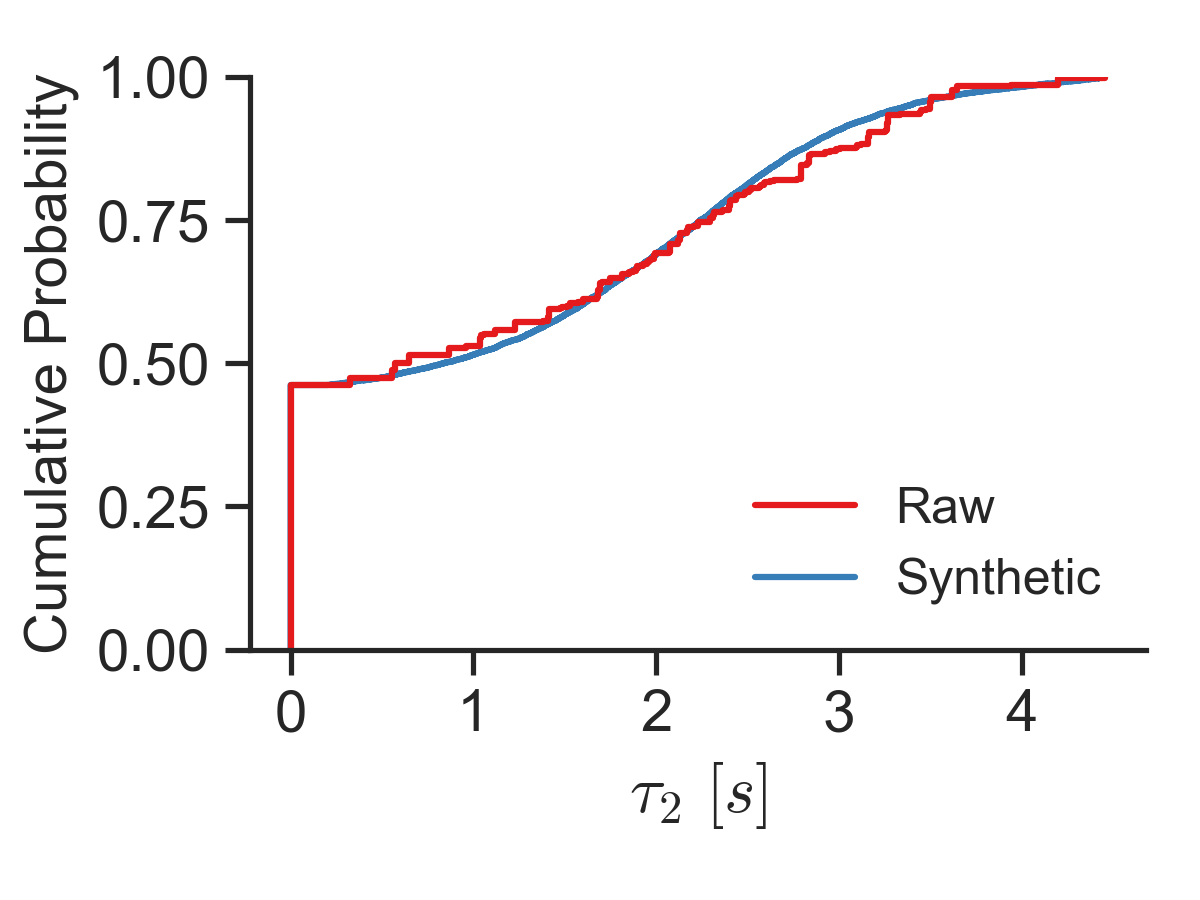}}
    \caption{Weighted CDFs of the six parameters of the raw and synthetic incidents.}
    \label{fig:wgtCDFsofrawandsyn}
\end{figure*}

\begin{figure*}[!t]
    \centering
    \subfloat[]{\includegraphics[width=0.26\textwidth]{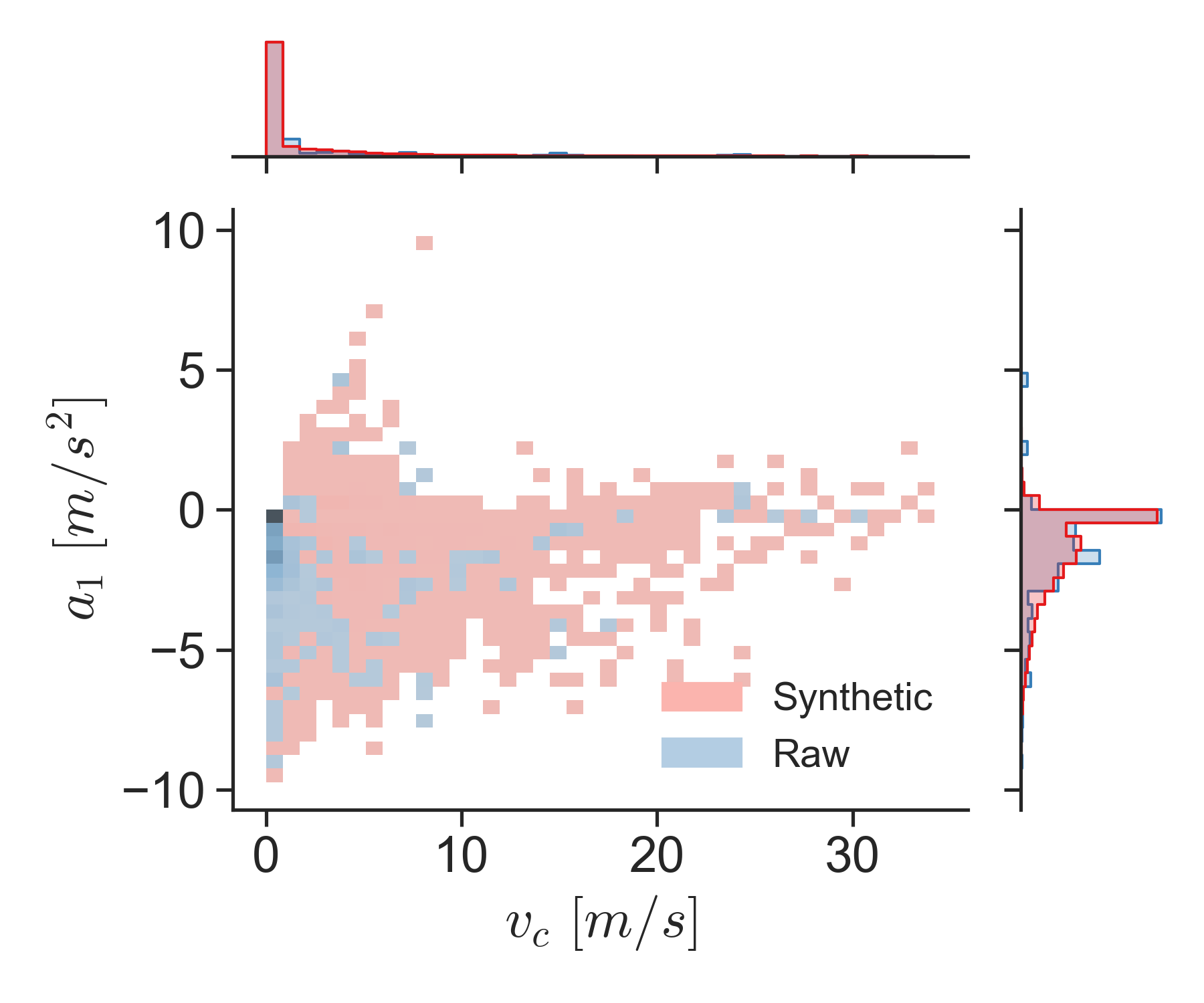}}
    \hfil
    \subfloat[]{\includegraphics[width=0.26\textwidth]{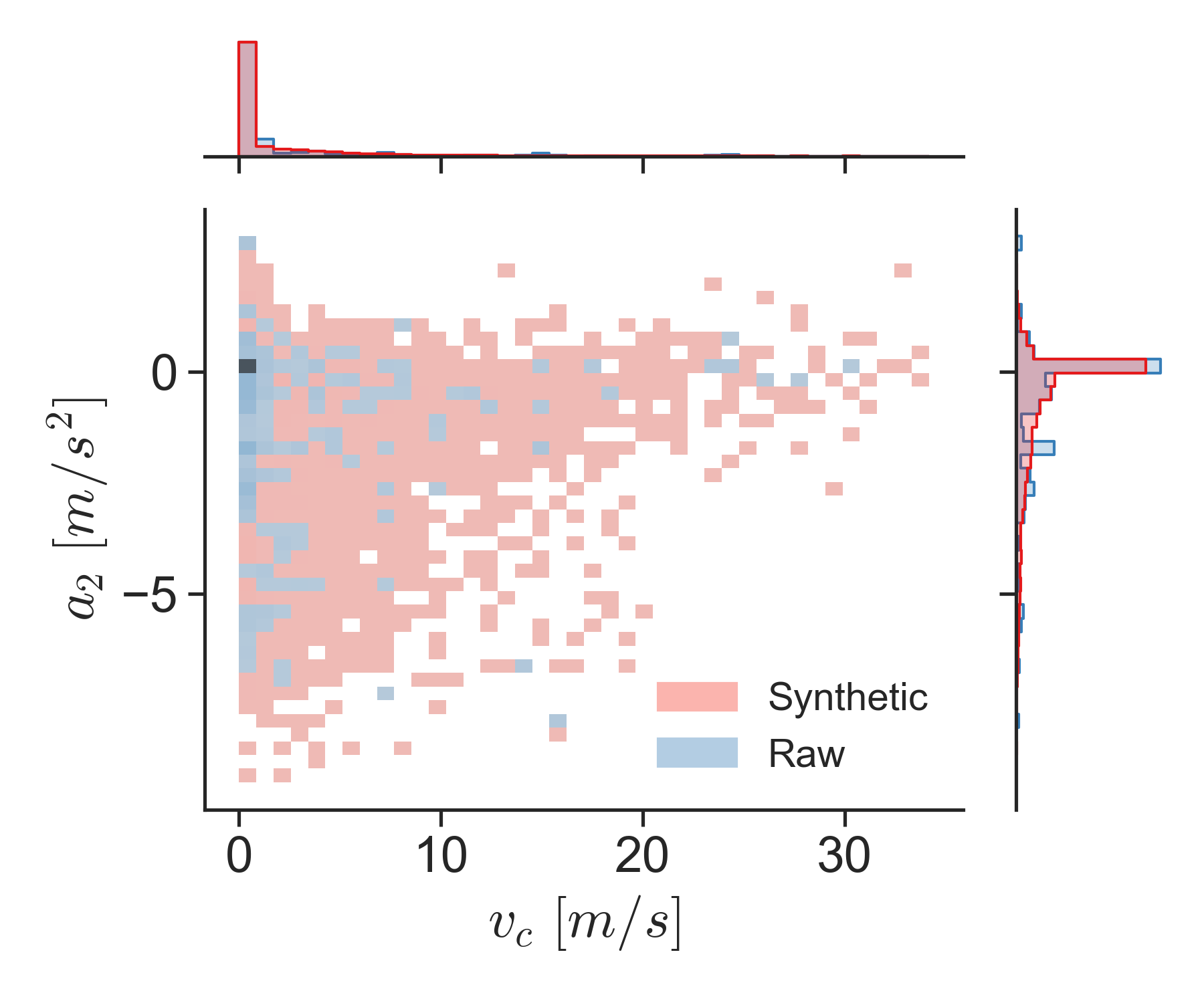}}
    \hfil
    \subfloat[]{\includegraphics[width=0.26\textwidth]{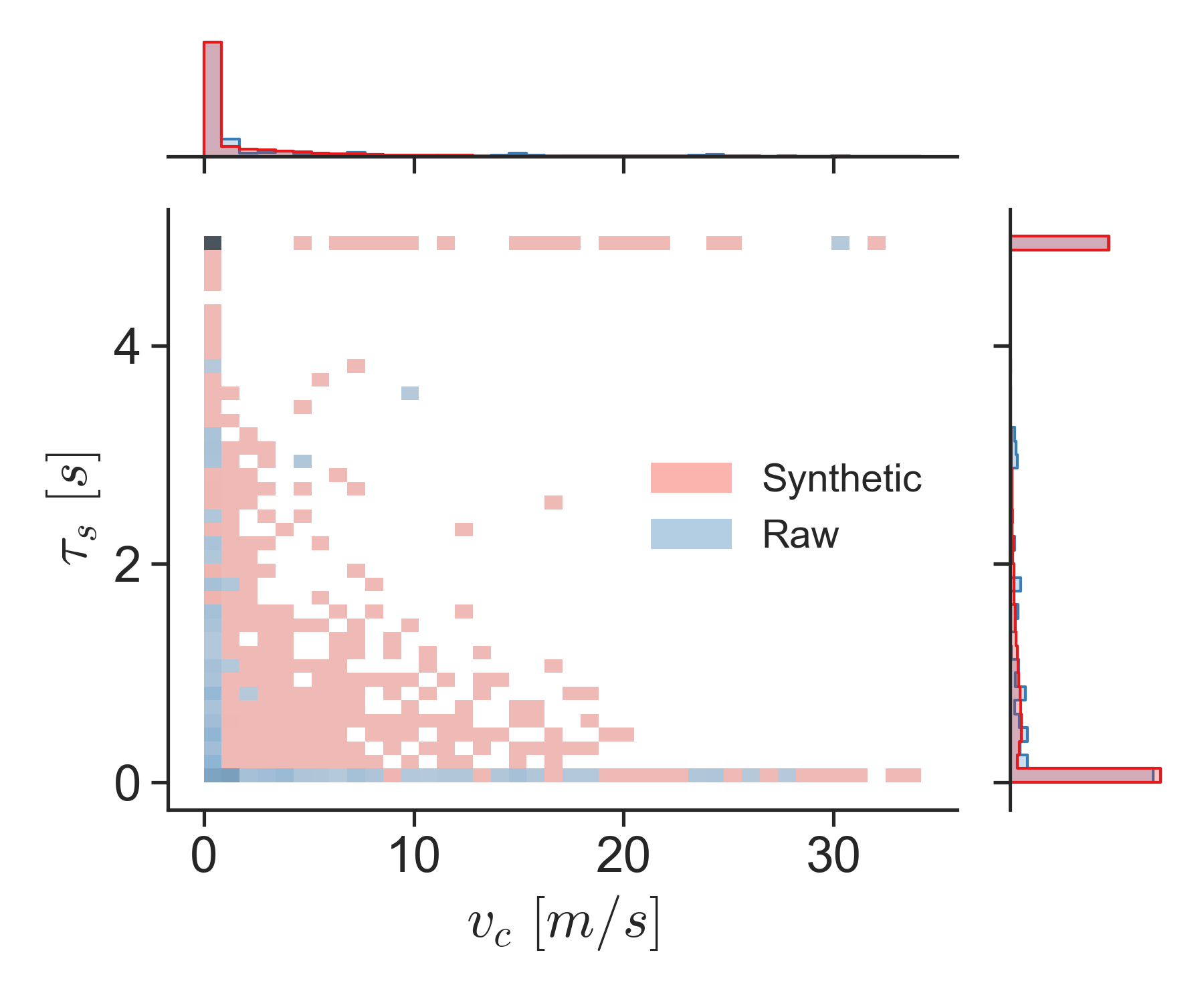}}
    \vfil
    \subfloat[]{\includegraphics[width=0.26\textwidth]{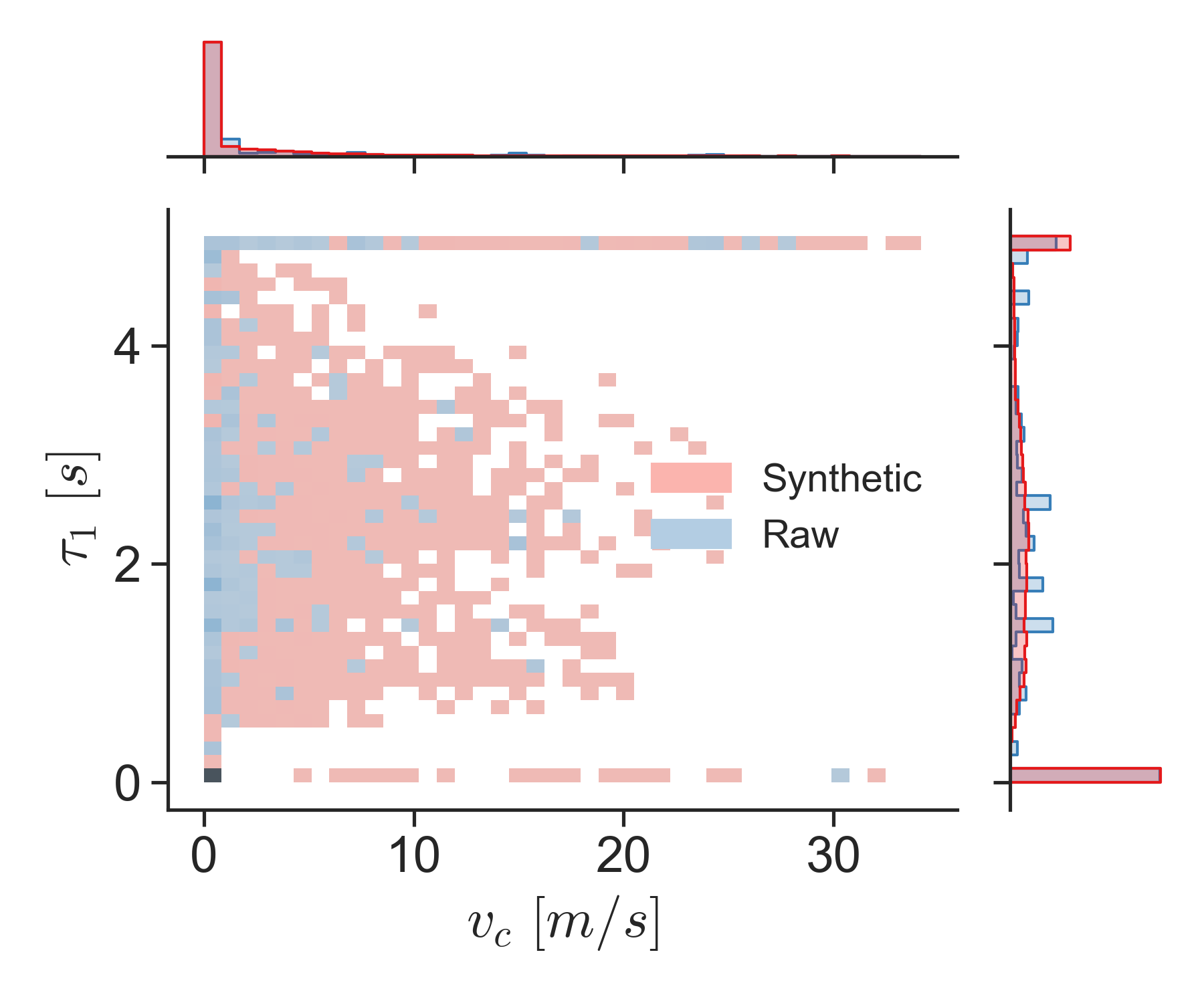}}
    \hfil
    \subfloat[]{\includegraphics[width=0.26\textwidth]{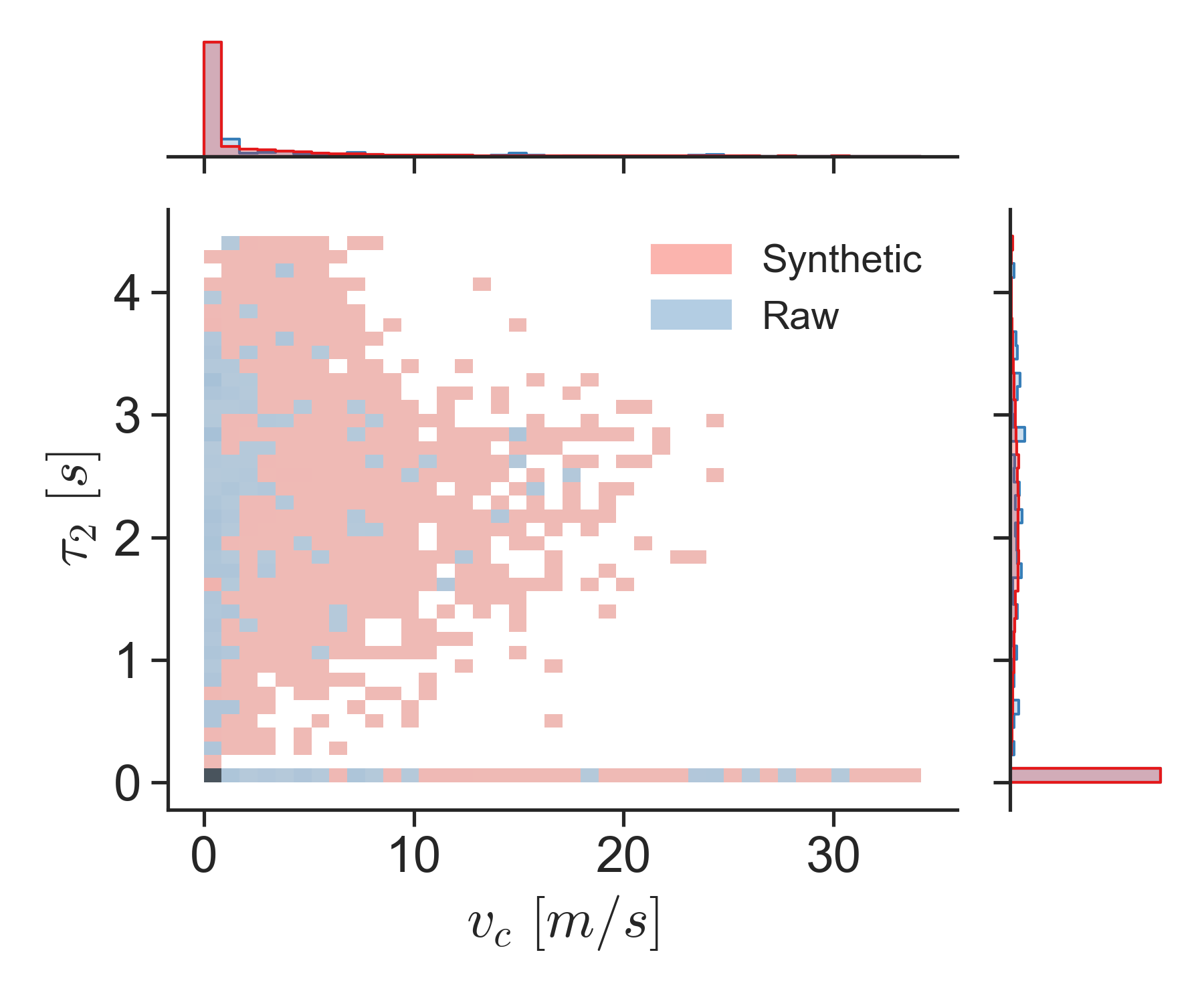}}
    \hfil
    \subfloat[]{\includegraphics[width=0.26\textwidth]{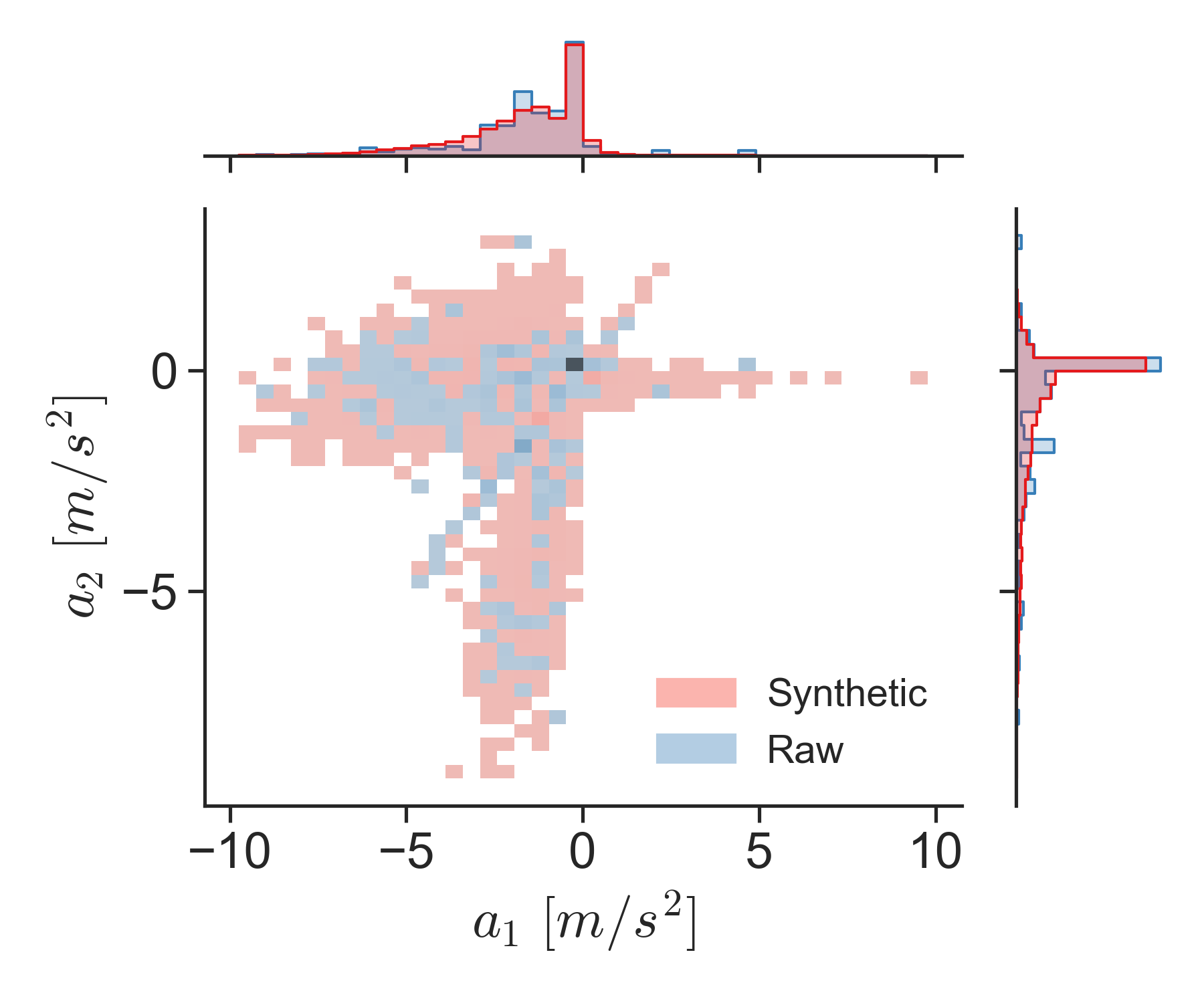}}
    \vfil
    \subfloat[]{\includegraphics[width=0.26\textwidth]{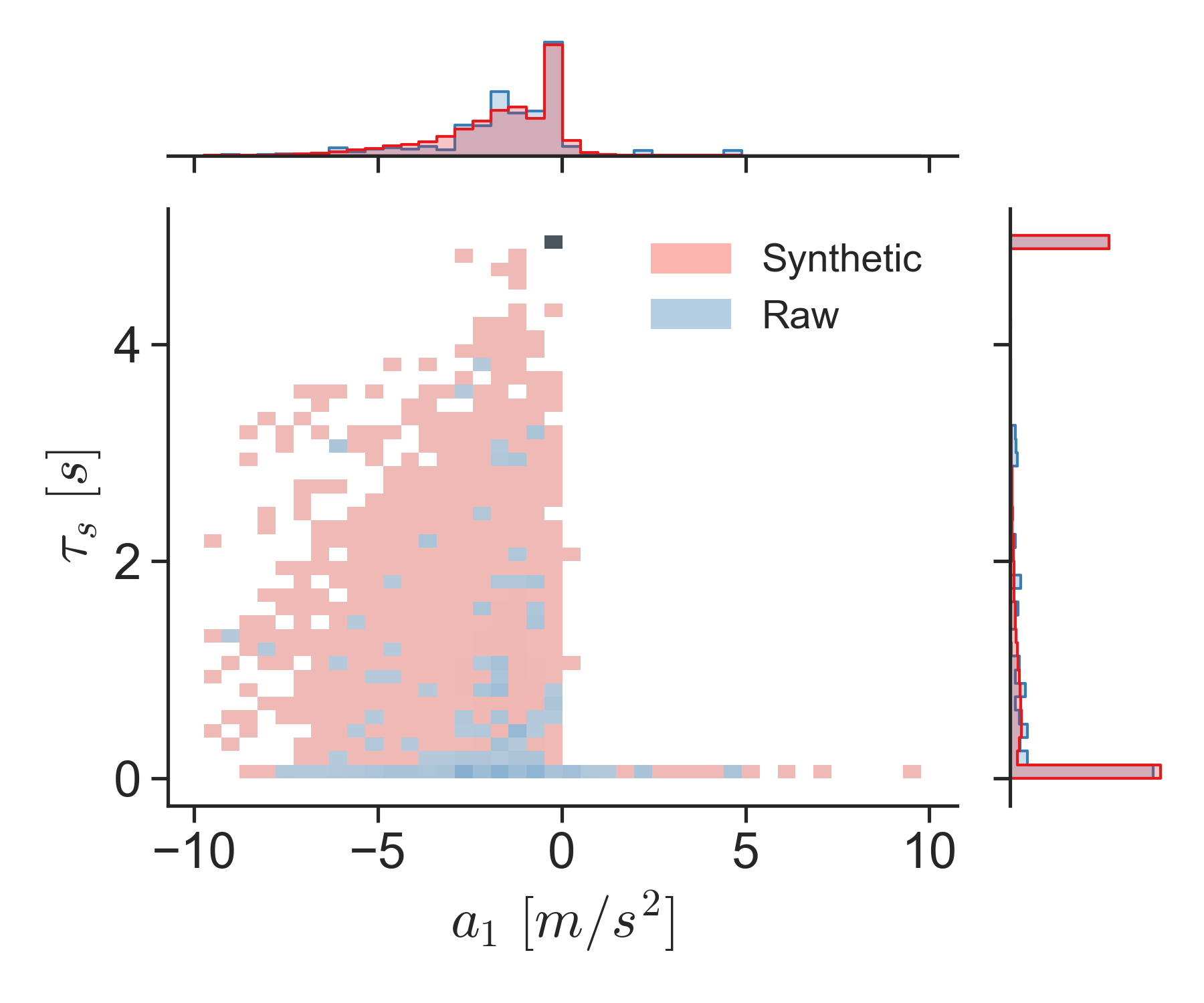}}
    \hfil
    \subfloat[]{\includegraphics[width=0.26\textwidth]{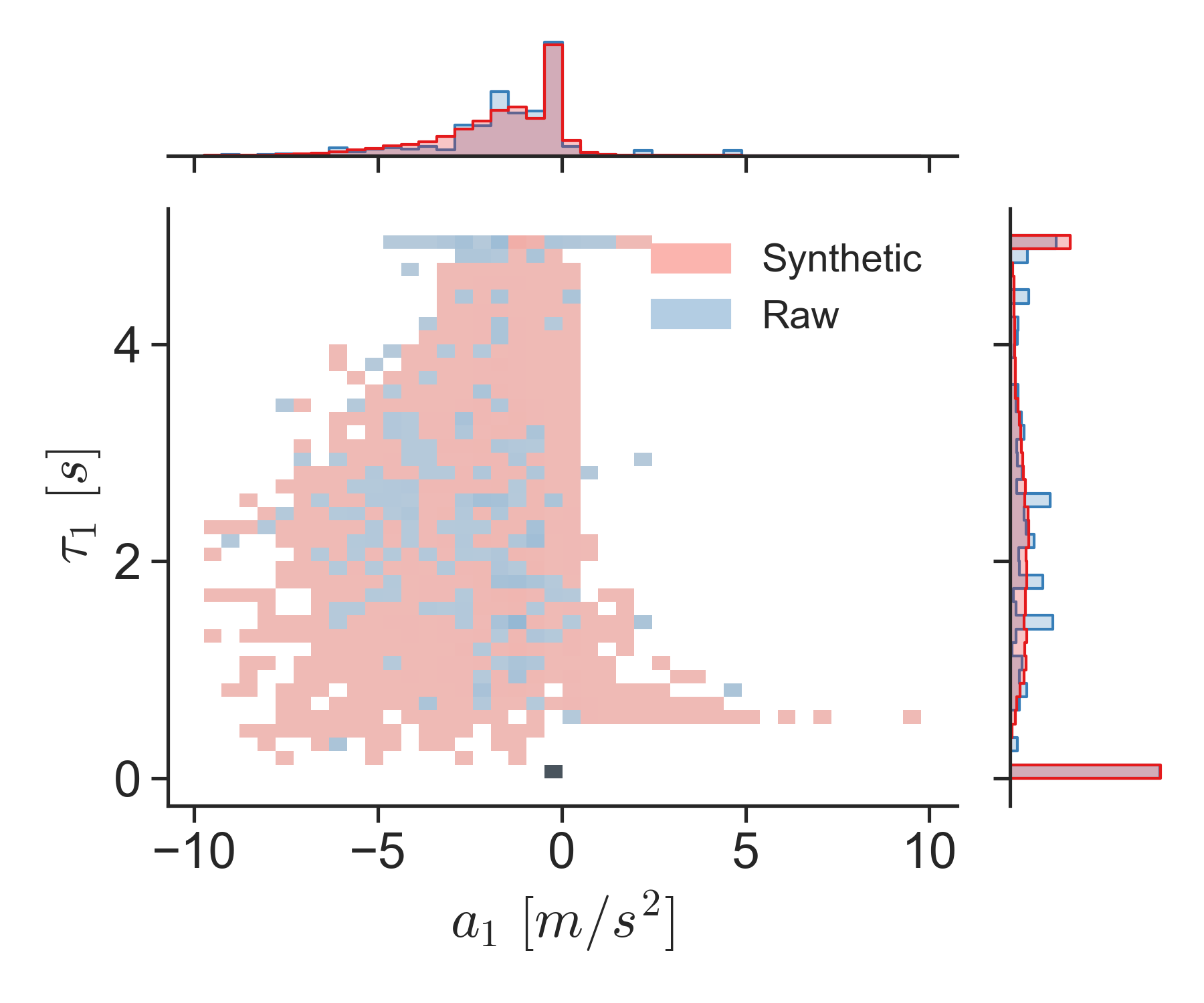}}
    \hfil
    \subfloat[]{\includegraphics[width=0.26\textwidth]{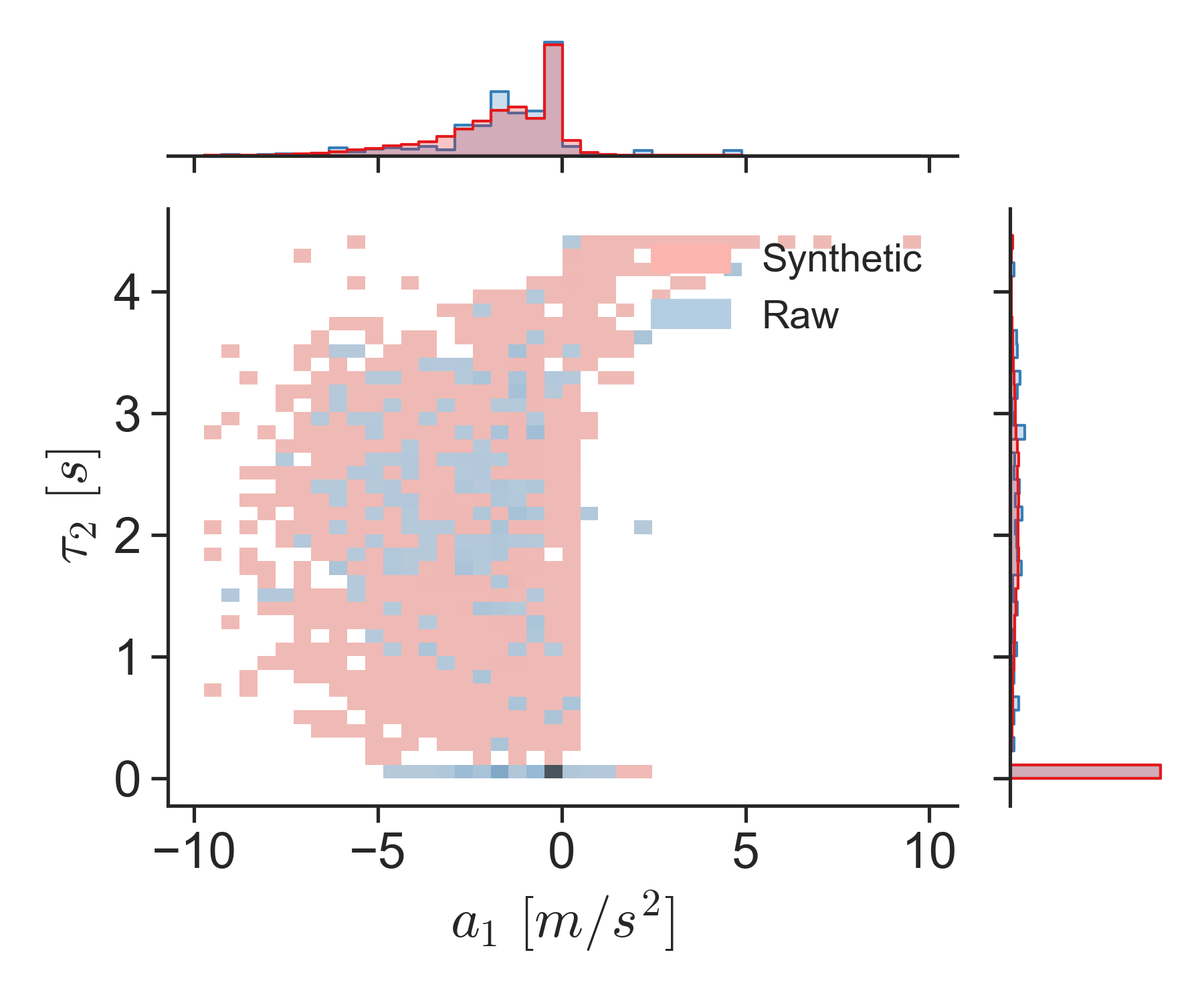}}
    \vfil
    \subfloat[]{\includegraphics[width=0.26\textwidth]{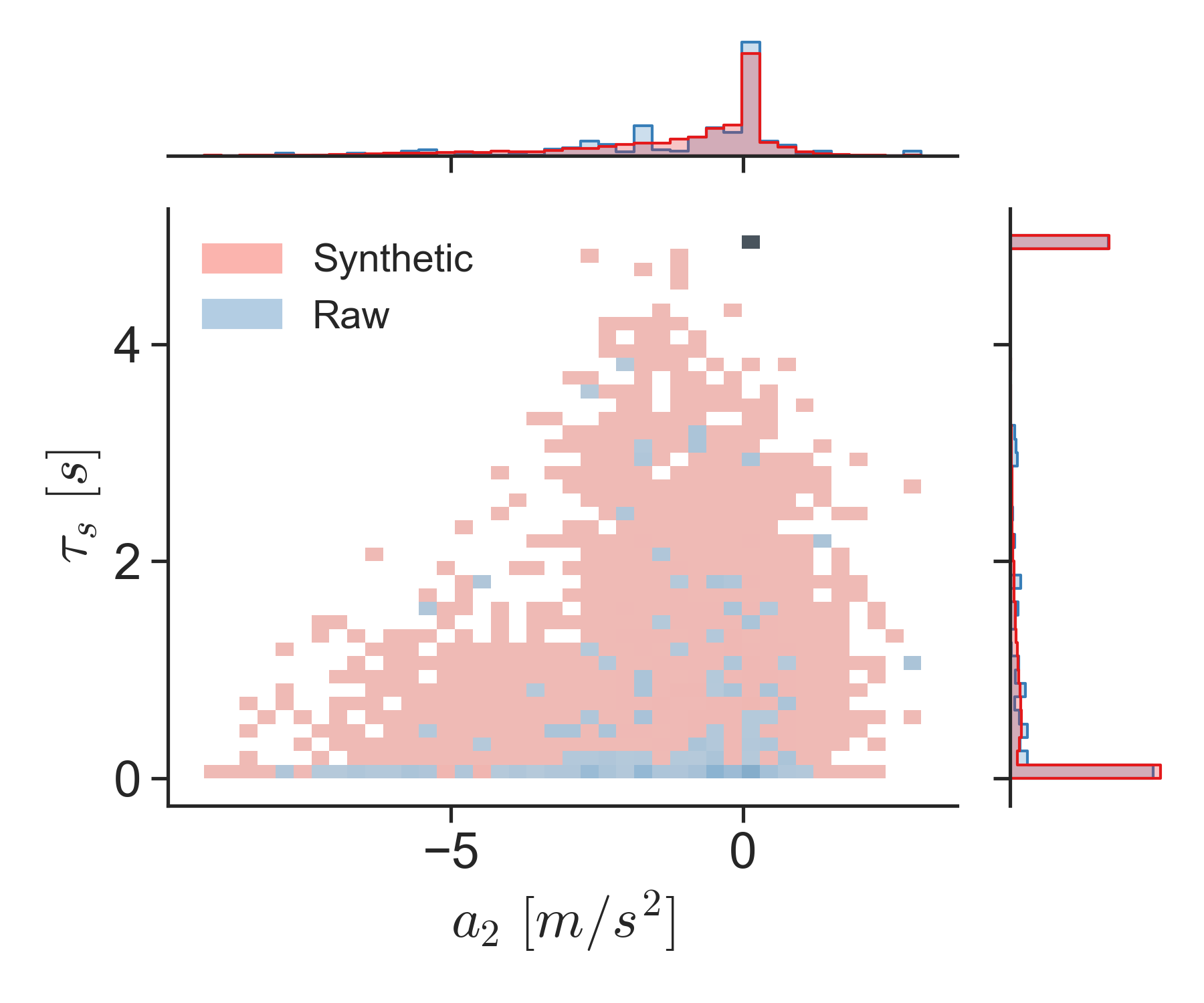}}
    \hfil
    \subfloat[]{\includegraphics[width=0.26\textwidth]{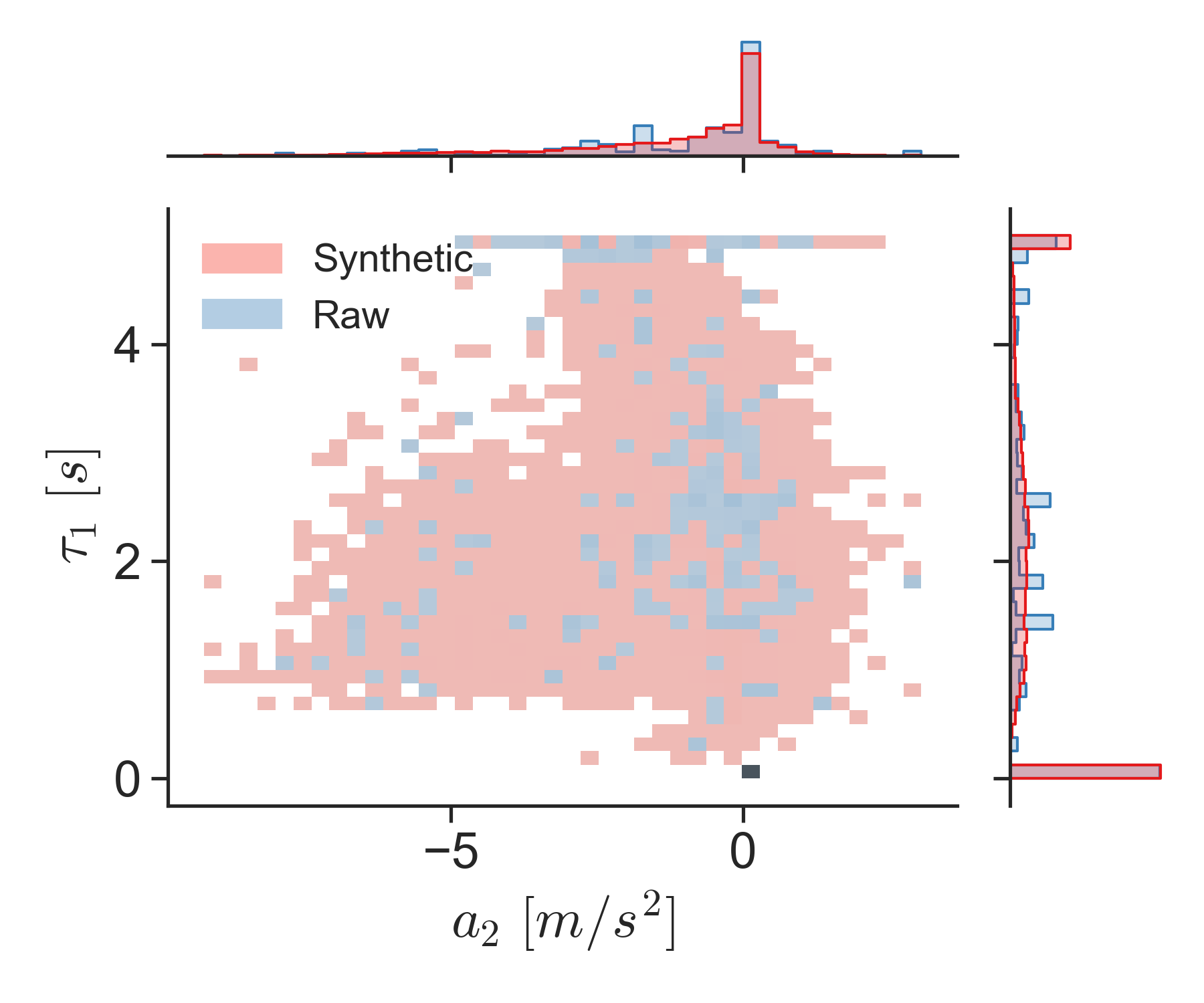}}
    \hfil
    \subfloat[]{\includegraphics[width=0.26\textwidth]{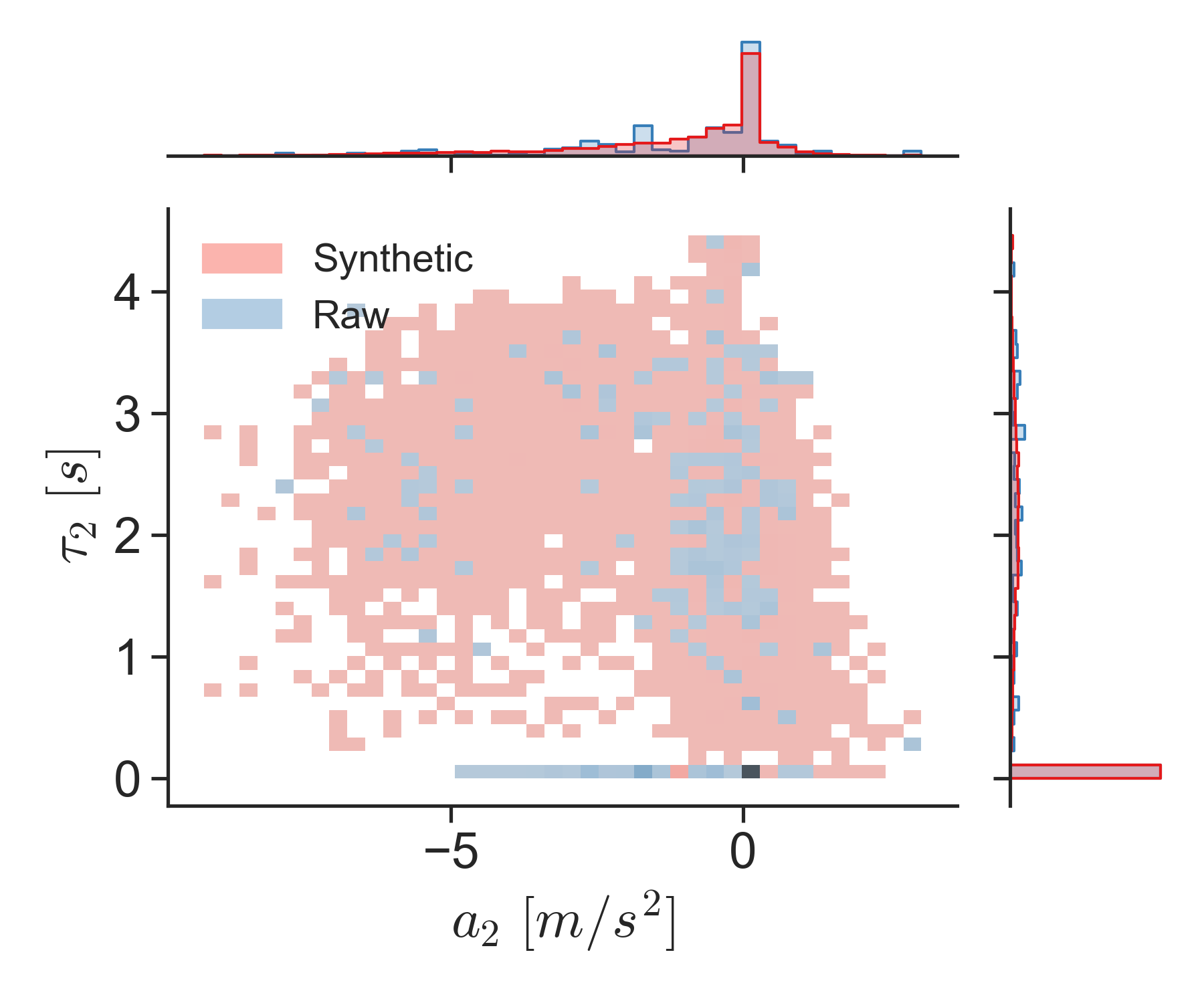}}
    \vfil
    \subfloat[]{\includegraphics[width=0.26\textwidth]{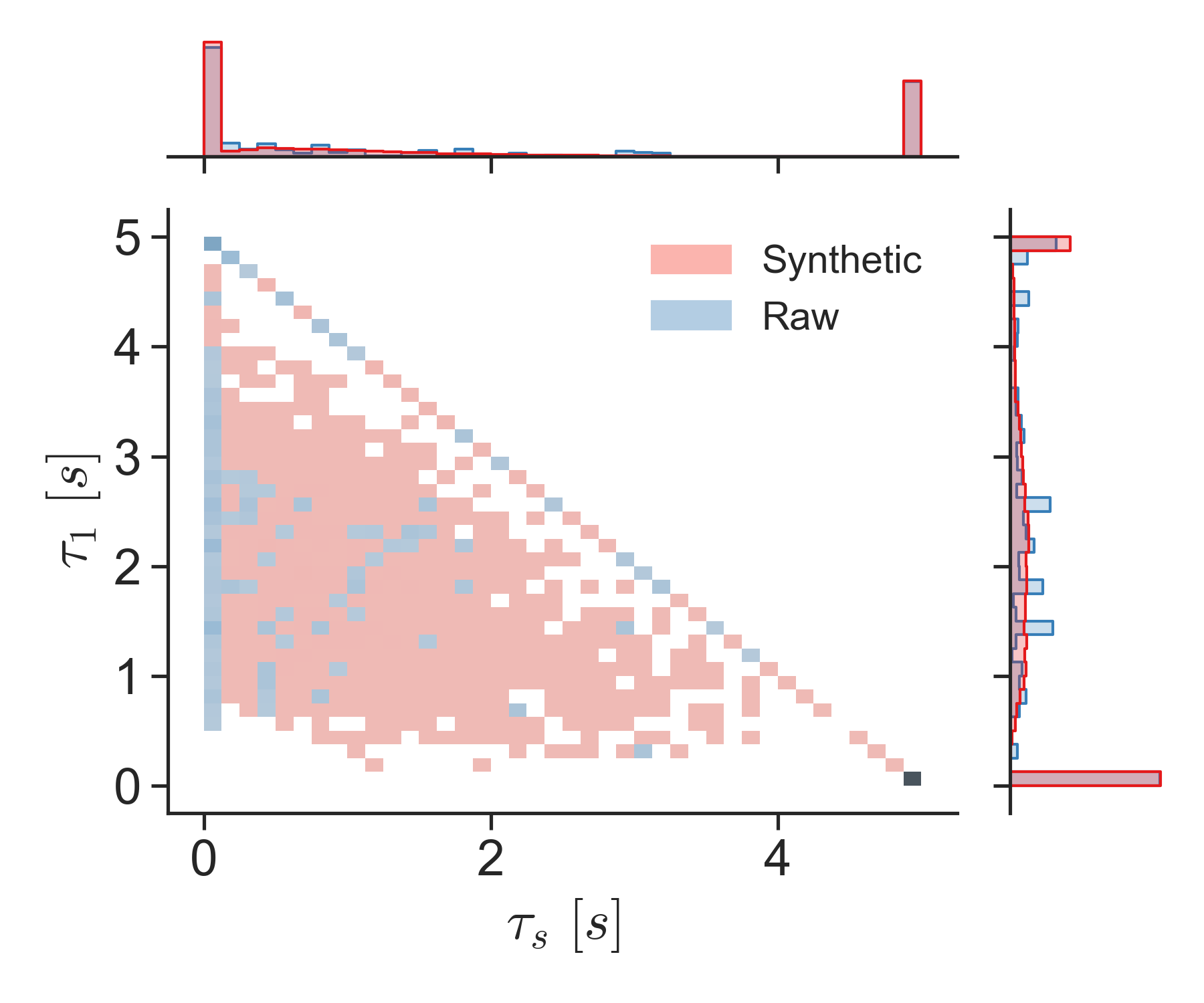}}
    \hfil
    \subfloat[]{\includegraphics[width=0.26\textwidth]{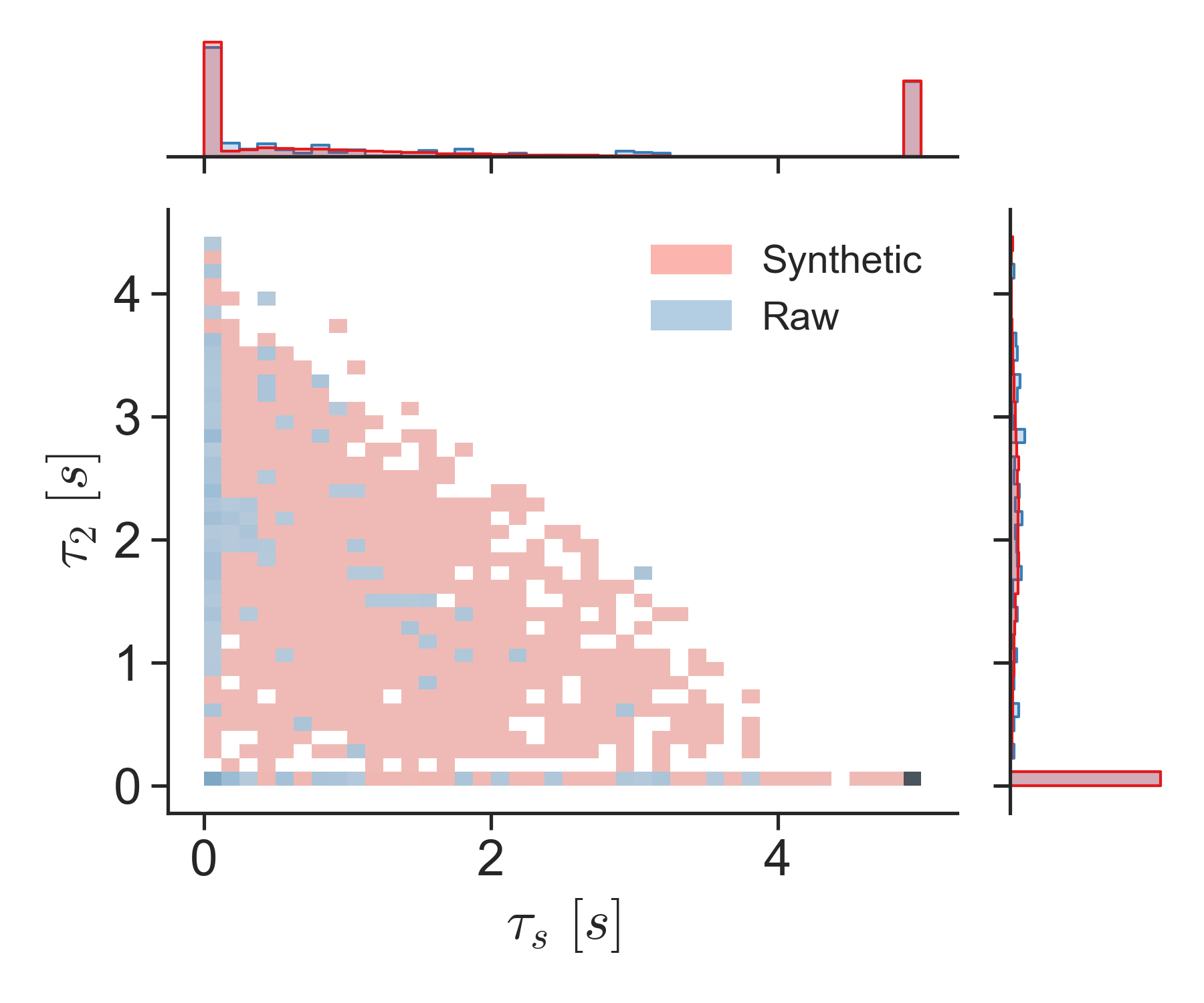}}
    \hfil
    \subfloat[]{\includegraphics[width=0.26\textwidth]{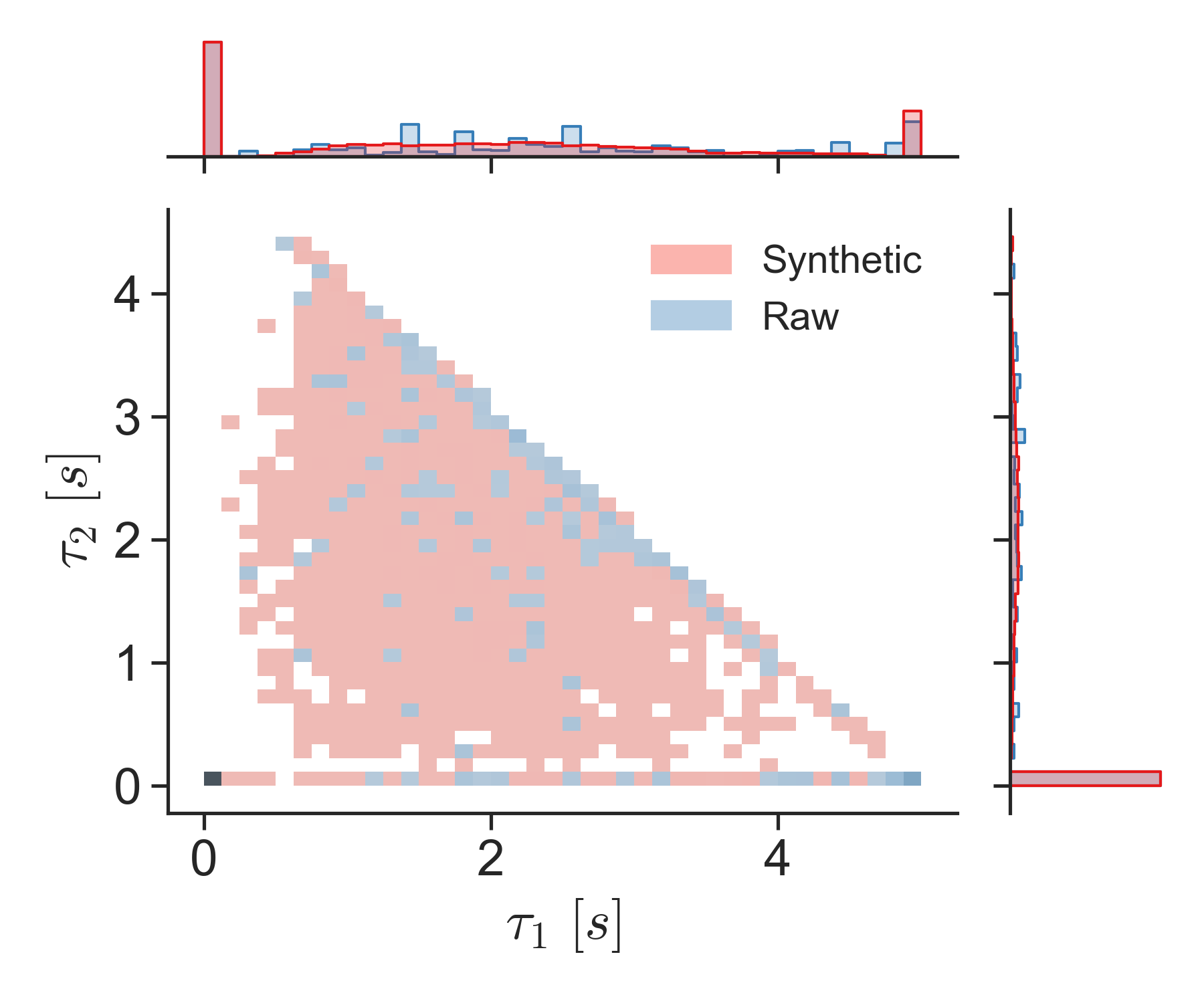}}
    \caption{Weighted joint distributions for every pair of parameters for the raw and synthetic incidents.}
    \label{fig:wgtjointtwoparam}
\end{figure*}

\begin{figure*}[!t]
    \centering
    \subfloat[]{\includegraphics[width=0.3\textwidth]{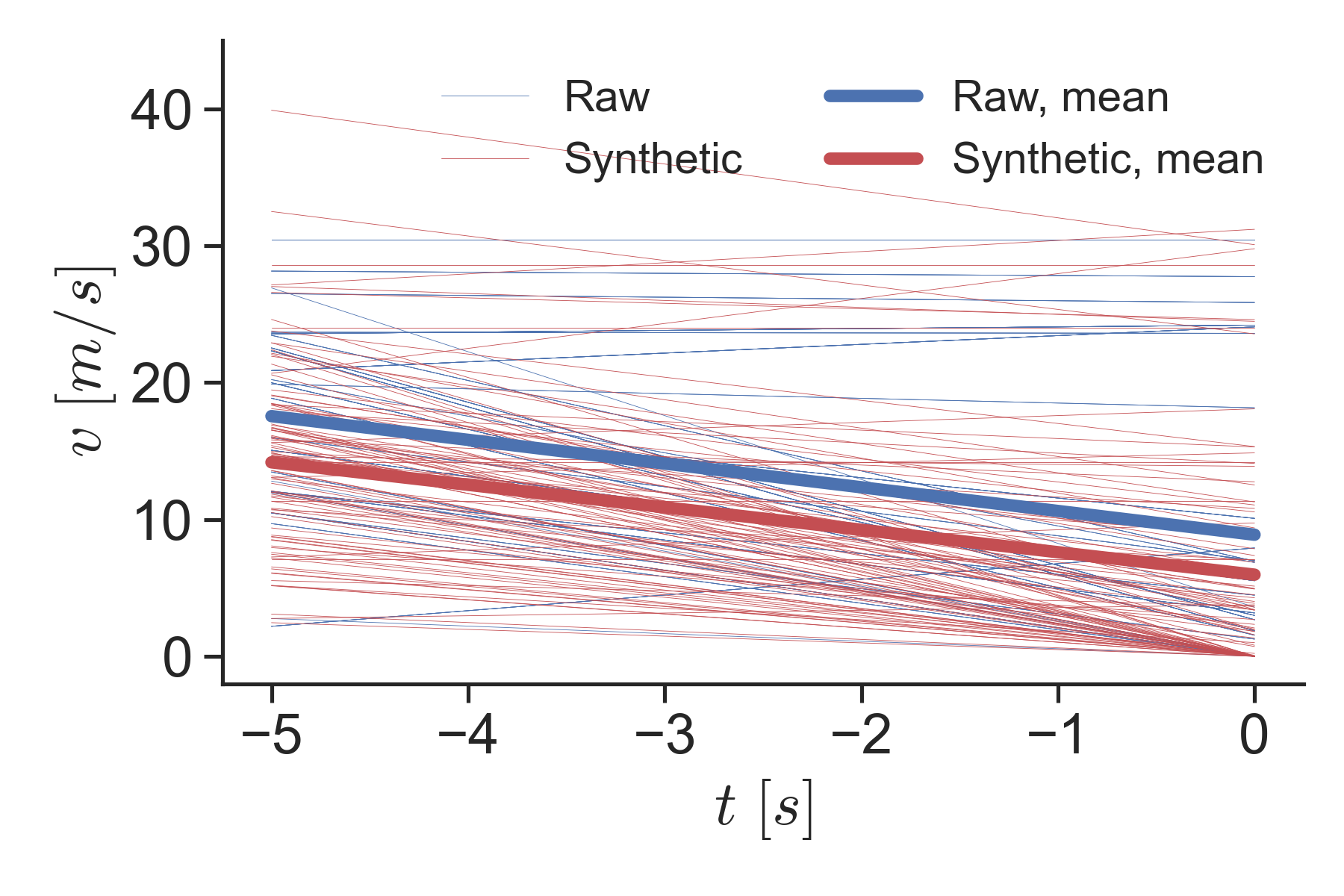}}
    \hfil
    \subfloat[]{\includegraphics[width=0.3\textwidth]{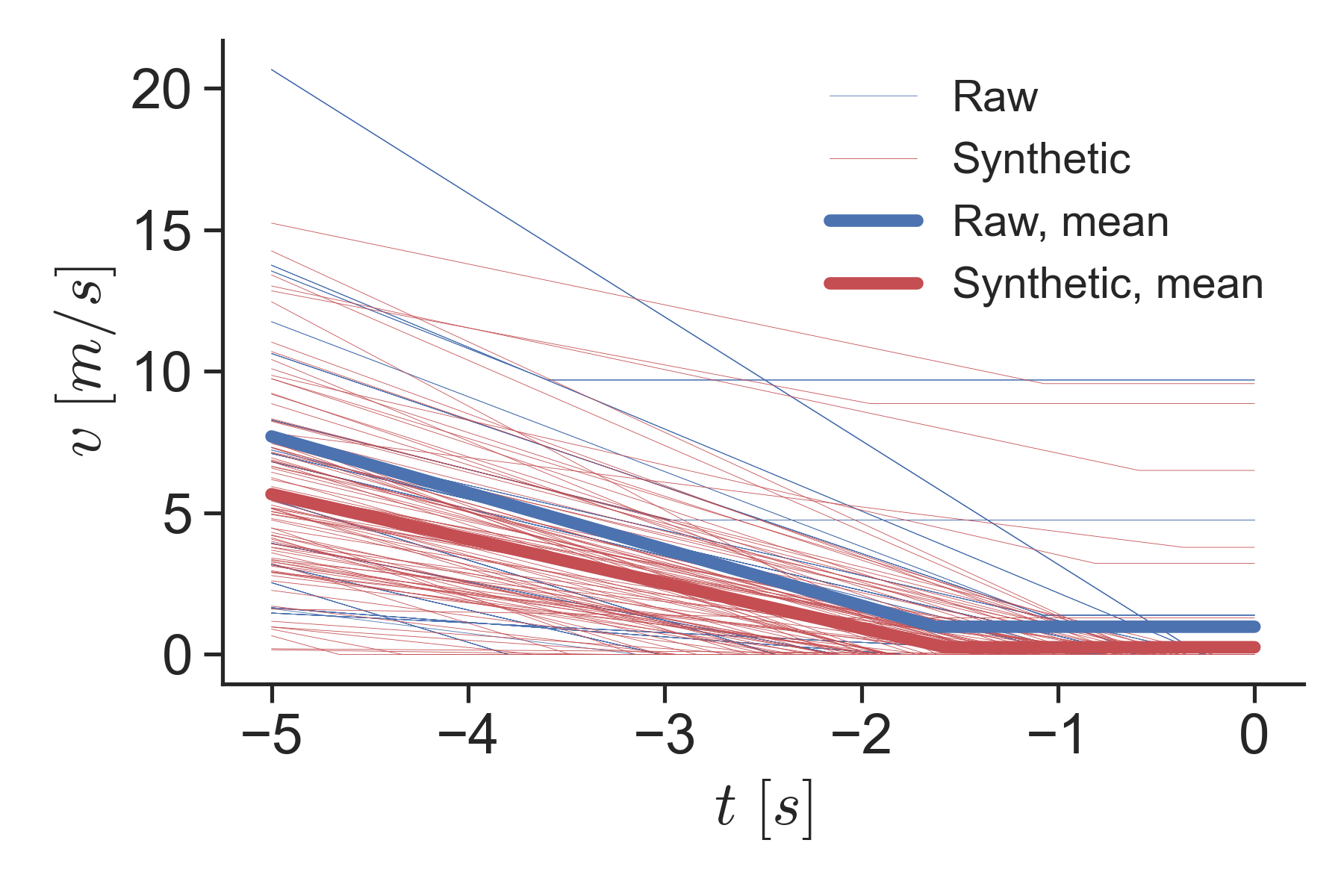}}
    \hfil
    \subfloat[]{\includegraphics[width=0.3\textwidth]{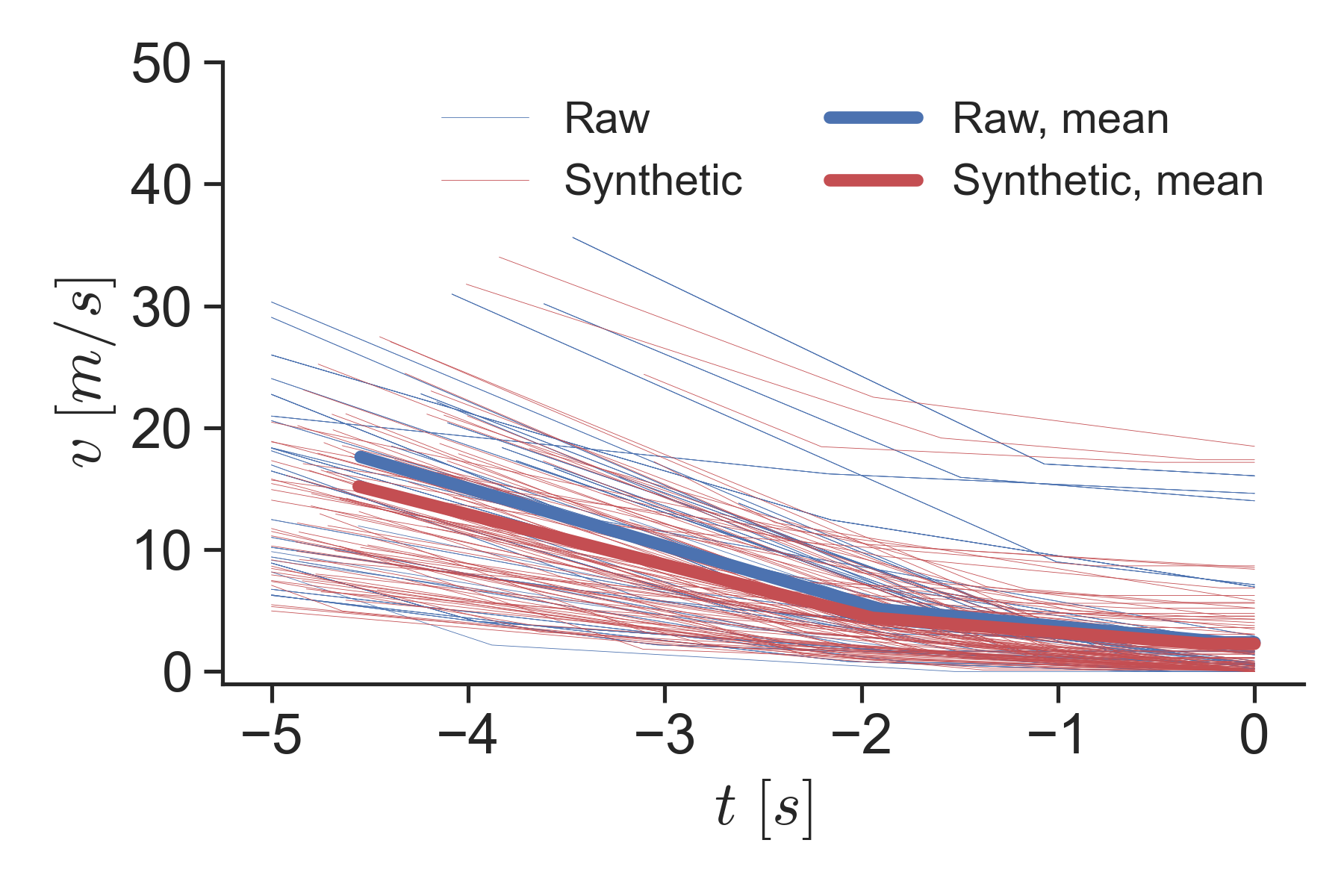}}
    \vfil
    \subfloat[]{\includegraphics[width=0.3\textwidth]{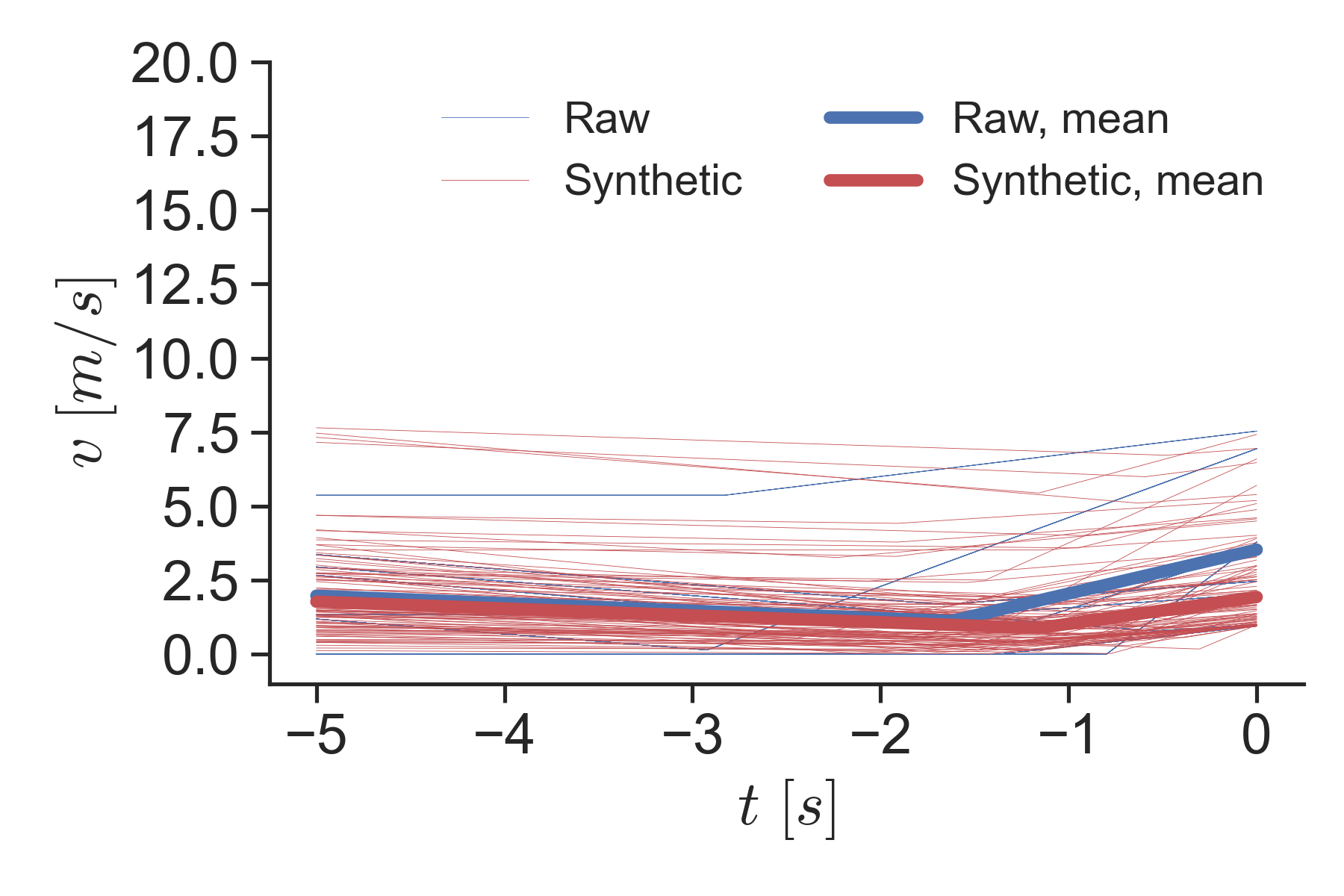}}
    \hfil
    \subfloat[]{\includegraphics[width=0.3\textwidth]{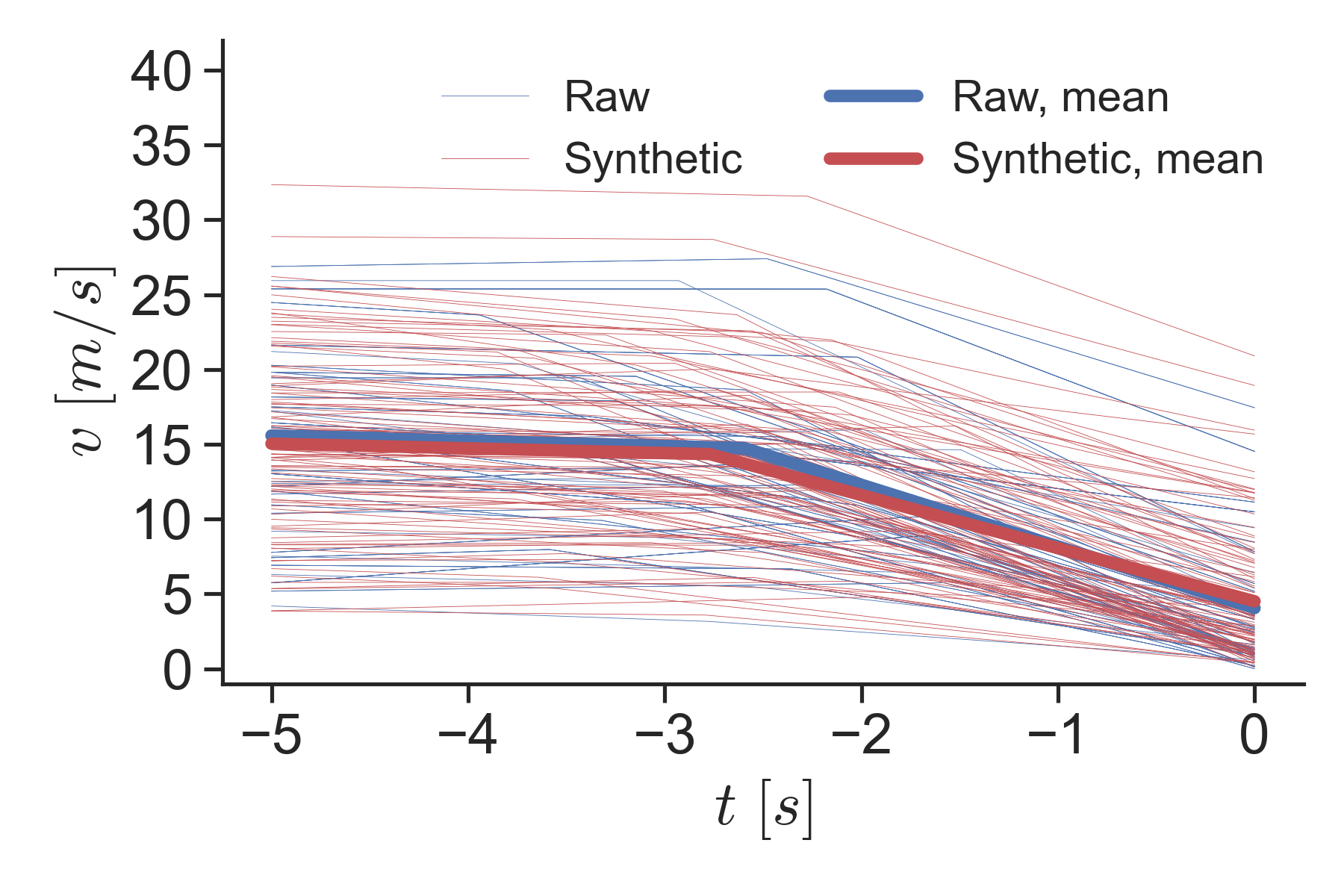}}
    \hfil
    \subfloat[]{\includegraphics[width=0.3\textwidth]{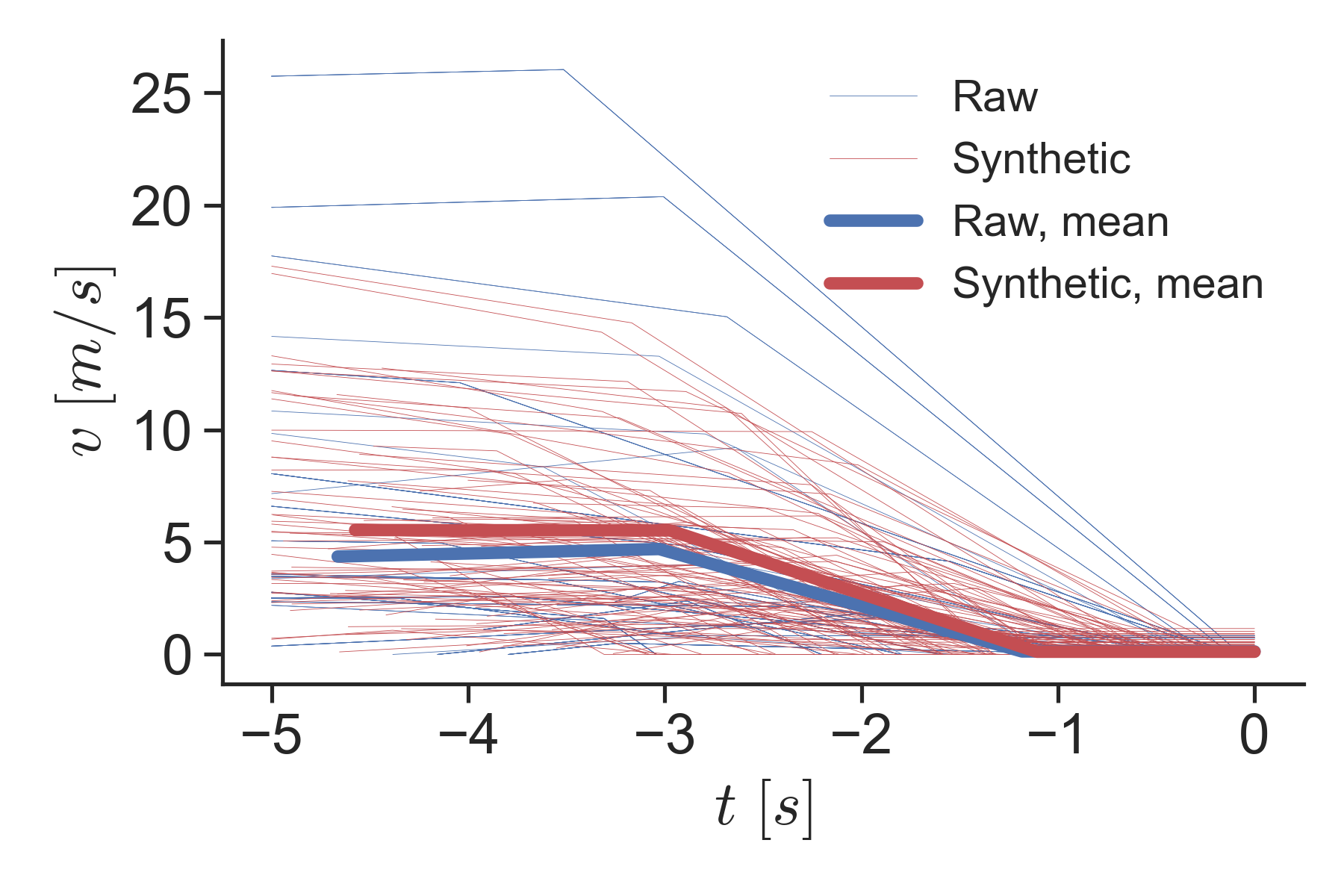}}
    \caption{Lead-vehicle speed profiles of the raw and synthetic incidents of all sub-datasets except S1. (a)-(b) show results from S2 to S7. The bold lines are with the weighted mean values of parameters describing the speed profile. The thin lines are 100 randomly sampled profiles for the raw and synthetic incidents respectively.}
    \label{fig:spdprofiles}
\end{figure*}

The weighted CDFs of the six parameters of the raw and synthetic incidents are shown in Fig. \ref{fig:wgtCDFsofrawandsyn}. Fig. \ref{fig:wgtjointtwoparam} shows the weighted joint distributions for every pair of parameters for the raw and synthetic incidents, and Fig. \ref{fig:spdprofiles} shows the lead-vehicle speed profiles for the raw and synthetic incidents of all sub-datasets except S1.

\section{Disscussion and Conclusion}
\begin{figure*}[!t]
    \centering
    \subfloat[]{\includegraphics[width=0.3\textwidth]{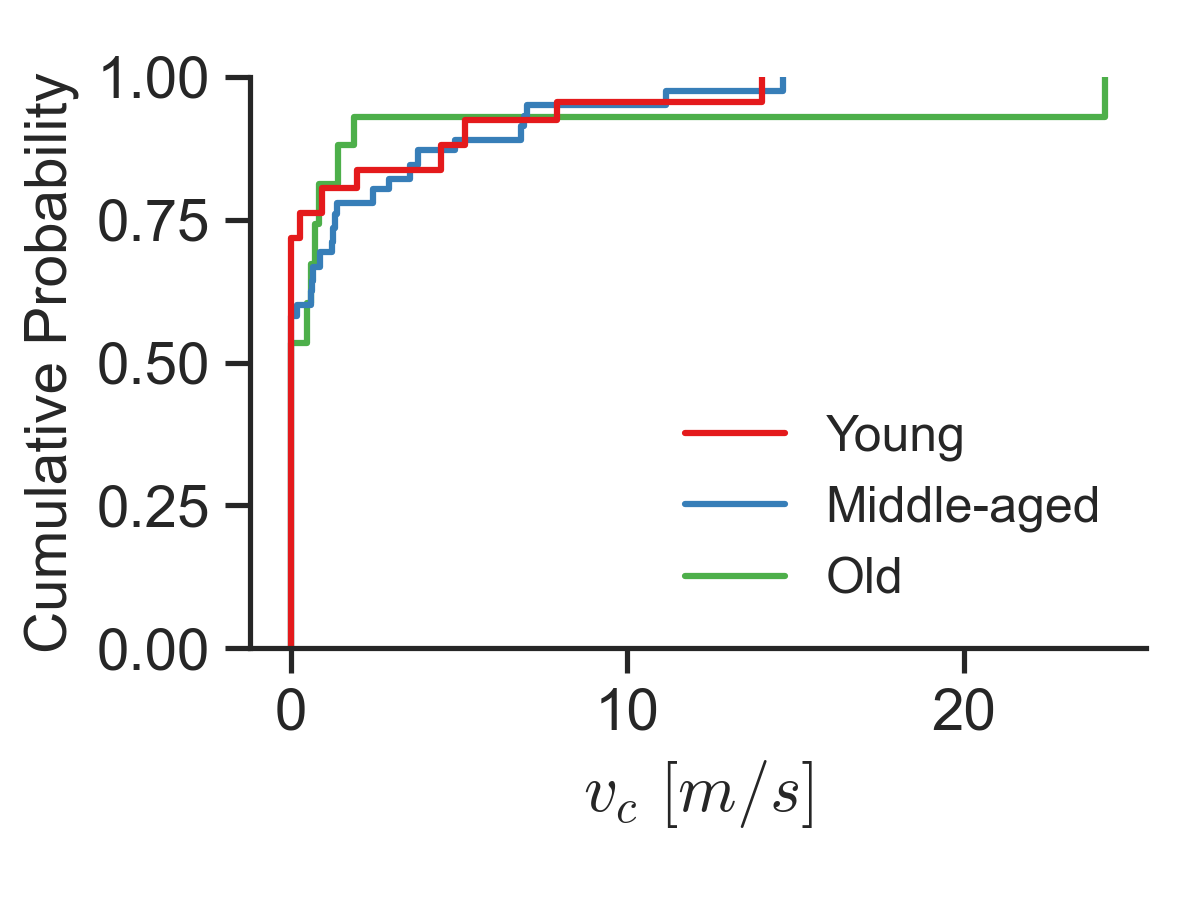}}
    \hfil
    \subfloat[]{\includegraphics[width=0.3\textwidth]{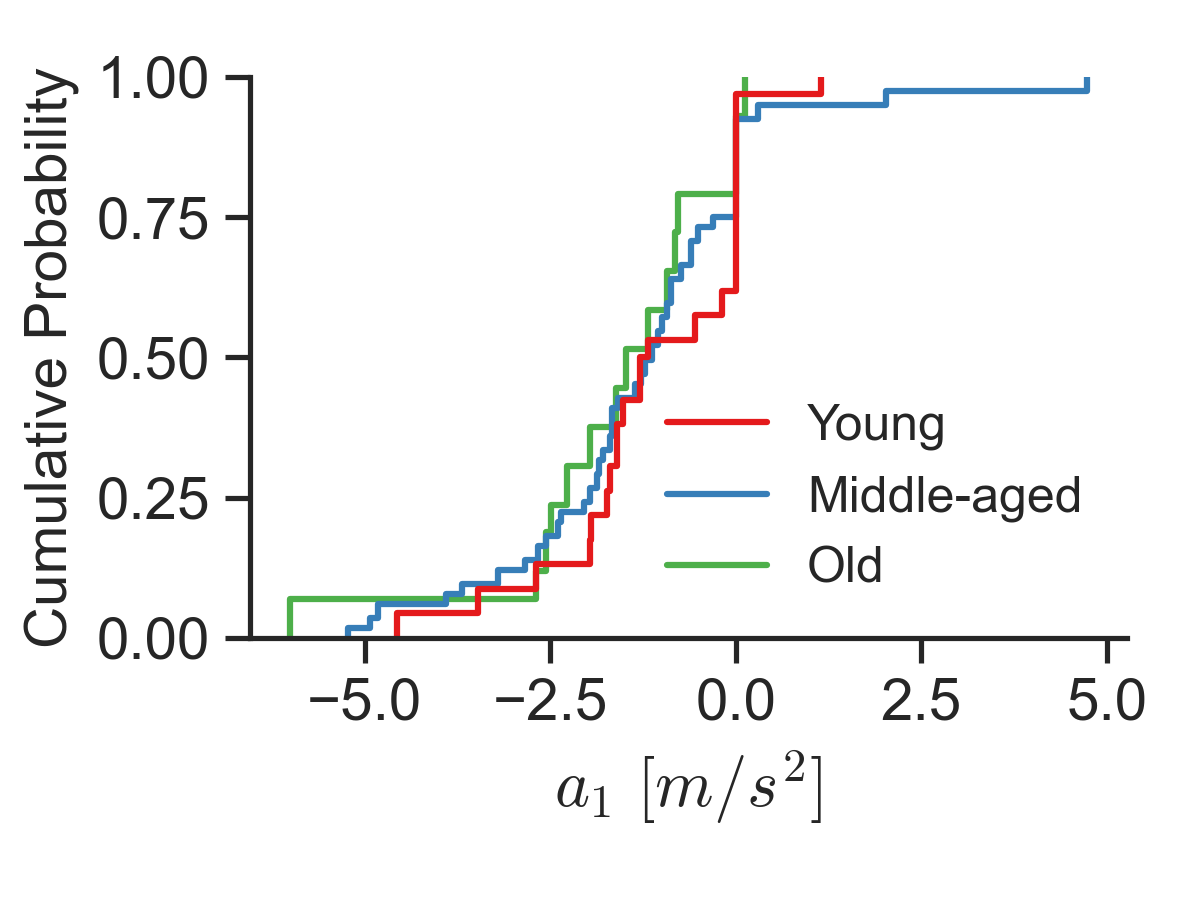}}
    \hfil
    \subfloat[]{\includegraphics[width=0.3\textwidth]{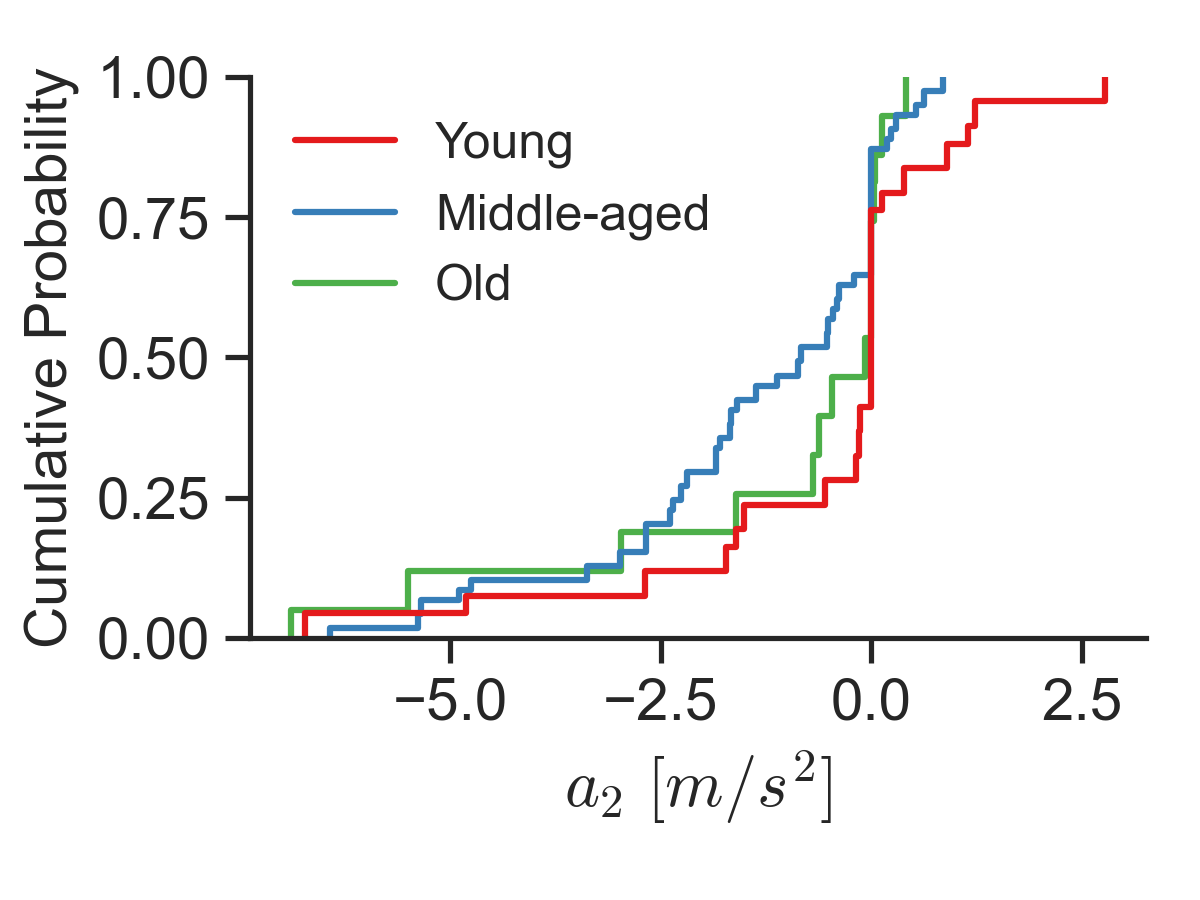}}
    \vfil
    \subfloat[]{\includegraphics[width=0.3\textwidth]{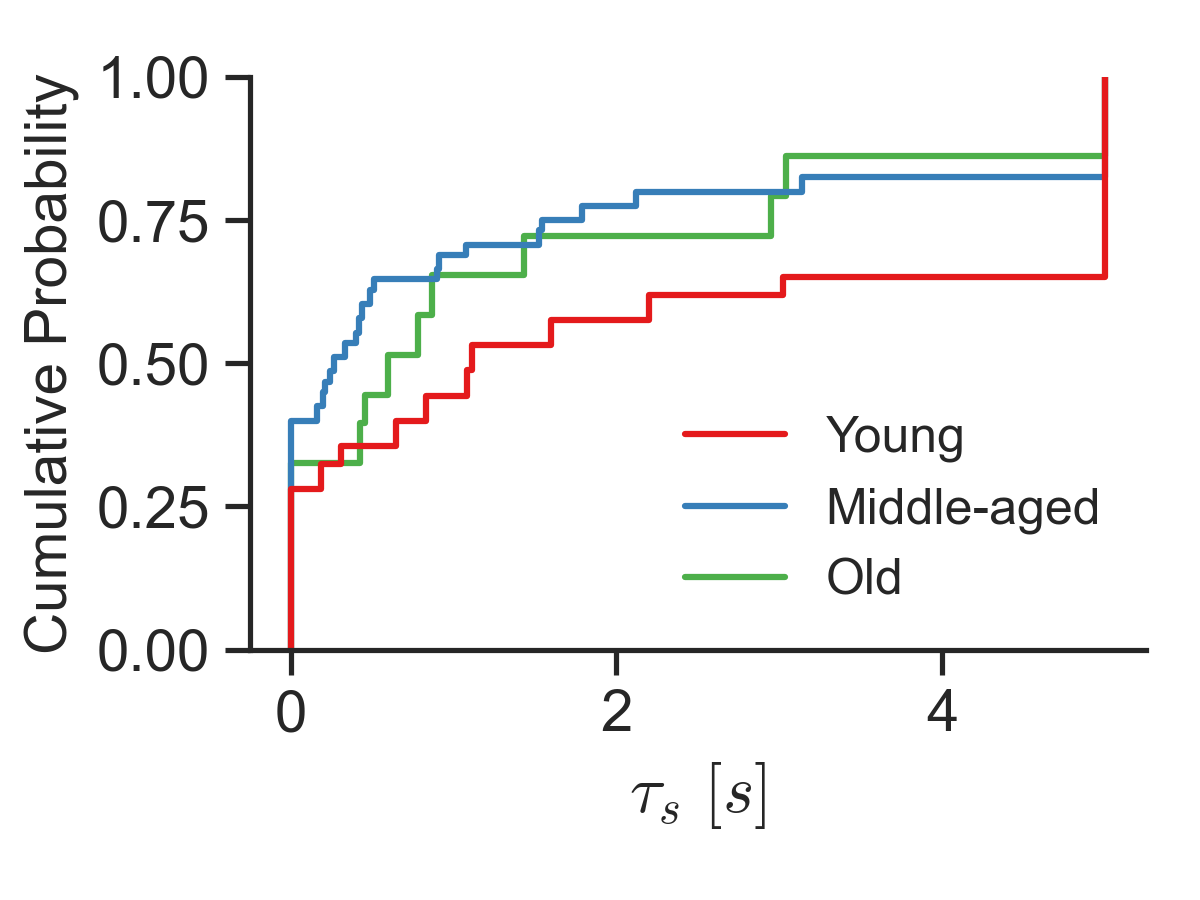}}
    \hfil
    \subfloat[]{\includegraphics[width=0.3\textwidth]{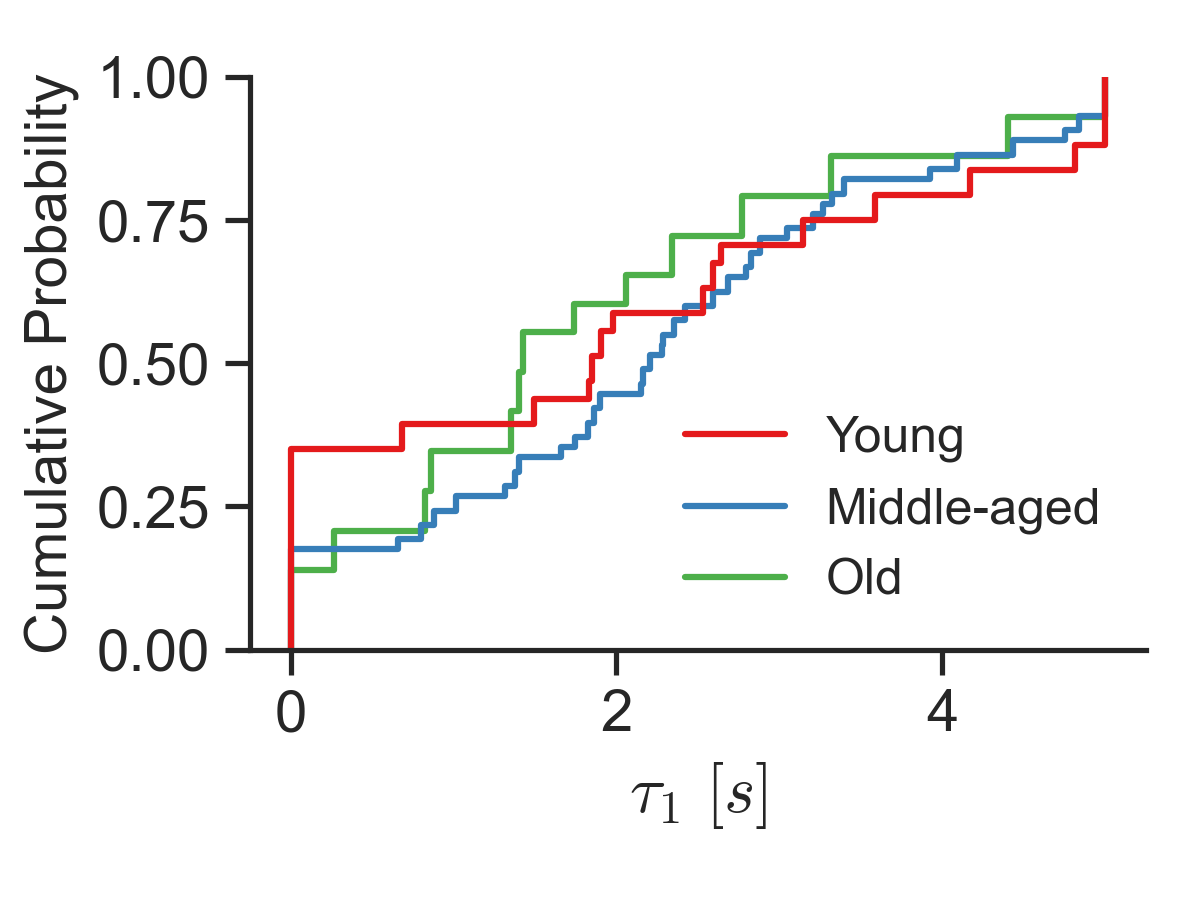}}
    \hfil
    \subfloat[]{\includegraphics[width=0.3\textwidth]{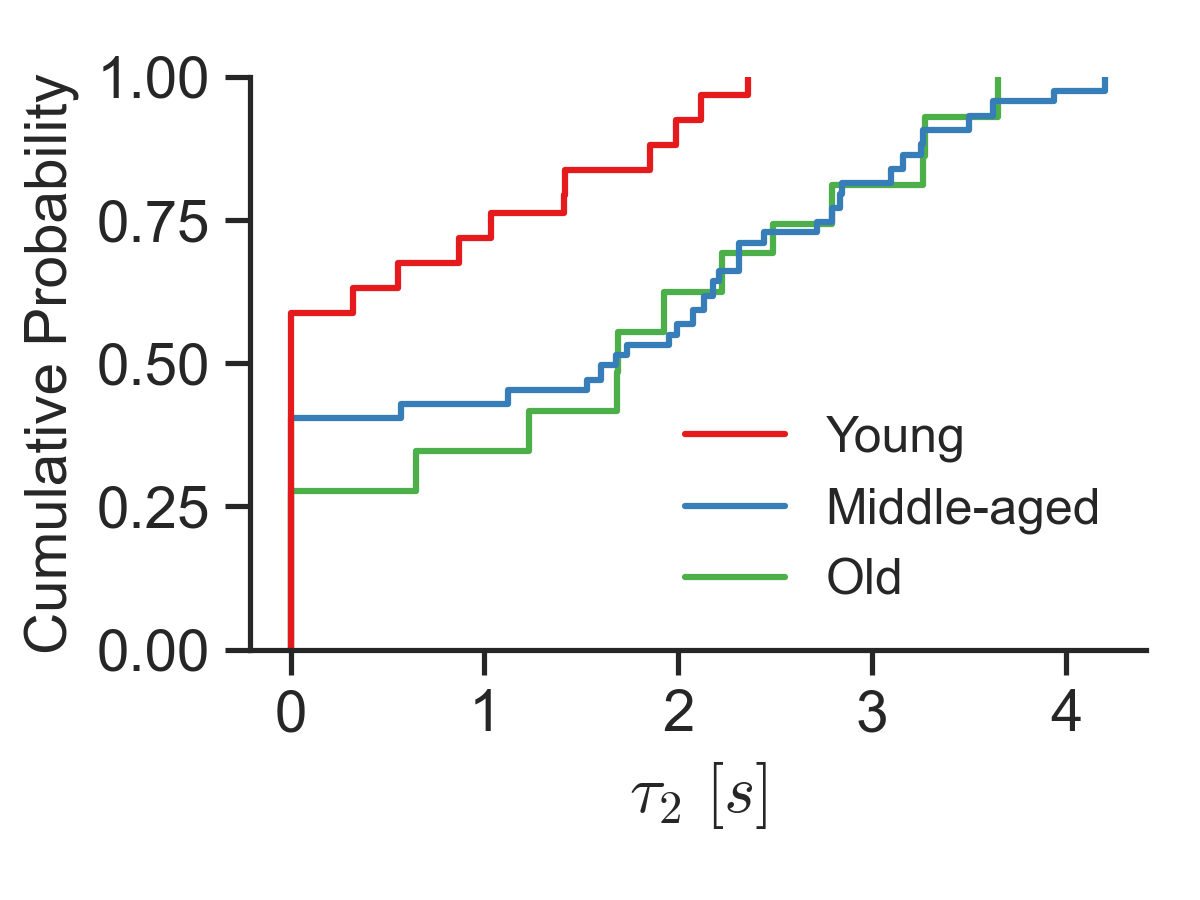}}
    \caption{Comparison among the weighted CDFs of the six parameters of different age groups in the combined crash dataset.}
    \label{fig:ageinfluence}
\end{figure*}

According to previous research \cite{flannagan2019replacement}, the SHRP2 dataset shows an over-representation of drivers under age 25 or over 64. To investigate the effects of the sampling bias on our study, the SHRP2 crash dataset is divided into three groups accordingly: young (under 25), middle-aged (over 24 and under 65), and senior (over 64), as shown in Table \ref{tab:agegroups}. The weighted two-sample KS tests were applied to determine whether there is any significant difference between any two of the three groups for each of the six parameters [$v_c$, $a_1$, $a_2$, $\tau_s$, $\tau_1$, $\tau_2$]. Based on the results presented in Table \ref{tab:kstestsagegroups}, it can be observed that only three out of eighteen comparisons depict notable dissimilarities at a significance level of 0.10. In addition, most of the weighted CDFs of the six parameters of different age groups shown in Fig. \ref{fig:ageinfluence} are well-aligned, indicating decent similarities. Therefore, we argue that the sampling bias of driver age in the SHRP2 dataset can be ignored in this study.

\begin{algorithm*} [!t]
\centering
\caption{SQL code for selecting rear-end crashes (subject as the struck vehicle) in the CISS dataset.}
\begin{algorithmic}
\STATE 
\STATE SELECT e.CASEID, e.VEHNO, e.EDREVENTNO, e.PTIME, e.PVALUE, o.AGE, o.ROLE, cra.CASEWGT FROM EDRPRECRASH as e
\STATE LEFT JOIN EDREVENT as event on e.CASEID = event.CASEID AND e.EDREVENTNO = event.EDREVENTNO AND e.VEHNO = event.VEHNO
\STATE LEFT JOIN GV as g on e.CASEID = g.CASEID AND e.VEHNO = g.VEHNO AND event.CDCEVENT = g.DVEVENT
\STATE LEFT JOIN CDC as c on e.CASEID = c.CASEID AND c.EVENTNO = event.CDCEVENT AND e.VEHNO = c.VEHNO
\STATE LEFT JOIN OCC as o on e.CASEID = o.CASEID AND e.VEHNO = o.VEHNO
\STATE WHERE g.CRASHCONF = ‘D’ AND e.PCODE = 1010 AND c.CDCPLANE = ‘B’
\STATE ORDER BY e.CASEID
\end{algorithmic}
\end{algorithm*}

\begin{algorithm*} [!t]
\centering
\caption{Query code for selecting rear-end pre-crashes/near-crashes (subject as the struck vehicle) in the SHRP2 dataset.}
\begin{algorithmic}
\STATE 
\STATE (Event Severity 1 = ‘Crash’ or ‘Near-crash’ AND Incident Type 1 = ‘Rear-end, struck’ AND Event Nature 1 = ‘Conflict with a following vehicle’)
\STATE \textbf{OR}
\STATE (Event Severity 1 != 'Crash' AND Event Severity 2 = ‘Crash’ or ‘Near-crash’ AND Incident Type 2 = 'Rear-end, struck' AND Event Nature 2 = ‘Conflict with a following vehicle’)
\end{algorithmic}
\end{algorithm*}

\begin{table*} [!t]
\caption{Weighted two-sample KS tests between CISS$_{sc}$ and SHRP2$_{sc}$\label{tab:sccomparison}} 
\centering
\begin{threeparttable}
\begin{tabular}{ccccccc}
\hline
Parameter & $v_c$ & $a_1$ & $a_2$ & $\tau_s$ & $\tau_1$ & $\tau_2$\\
\hline
statistic & 0.15 & 0.28 & 0.27 & 0.27 & 0.26 & 0.26\\
p-value & 0.97 & 0.33 & 0.40 & 0.37 & 0.41 & 0.44\\
\hline
\end{tabular}
\begin{tablenotes}
\RaggedRight
\item $^*$The sample sizes are as follows: 49 for CISS$_{sc}$ and 20 for SHRP2$_{sc}$.
\end{tablenotes}
\end{threeparttable}
\end{table*}

\begin{table*}[!t]
\caption{Three age groups in the SHRP2 crash dataset\label{tab:agegroups}} 
\centering
\begin{tabular}{ccc}
\hline
Group & Age & Sample size \\
\hline
Young & $<$25 & 24 \\
Middle-aged & 25-64 & 44 \\
Senior & $>$64 & 15 \\
\hline
\end{tabular}
\end{table*}

\begin{table*}[!t]
\caption{Weighted two-sample KS tests among three age groups in the SHRP2 crash dataset\label{tab:kstestsagegroups}} 
\centering
\begin{threeparttable}
\begin{tabular}{ccccccc}
\hline
& \multicolumn{6}{c}{p-value}\\
\cmidrule(rl){2-7}
Comparison$^a$ & $v_c$ & $a_1$ & $a_2$ & $\tau_s$ & $\tau_1$ & $\tau_2$ \\
\hline
Young V.S. Middle-aged & 0.81 & 0.73 & 0.03$^b$ & 0.15 & 0.57 & 0.02$^b$\\
Young V.S. Senior &  0.73 & 0.56 & 0.92 & 0.81 & 0.81 & 0.08$^b$\\
Middle-aged V.S. Senior & 0.96 & 0.96 & 0.44 & 0.49 & 0.49 & 0.96\\
\hline
\end{tabular}
\begin{tablenotes}
\RaggedRight
\item $^a$ The sample sizes are as follows: 24 for the young group, 44 for the middle-aged group, and 15 for the senior group.
\item $^b$ Significant under the significance level of 0.10.
\end{tablenotes}
\end{threeparttable}
\end{table*}

\bibliographystyle{IEEEtran}
\bibliography{Supplement}